\documentclass[twoside]{article}
\usepackage[accepted]{aistats2025}

\usepackage[round]{natbib}

\usepackage[utf8]{inputenc} 
\usepackage[T1]{fontenc}    
\usepackage{hyperref}       
\usepackage{url}            
\usepackage{booktabs}       
\usepackage{amsfonts}       
\usepackage{nicefrac}       
\usepackage{microtype}      
\usepackage{xcolor}         

\usepackage{mathtools}
\usepackage{float}
\usepackage{amsmath,amsfonts,amssymb,amsthm}
\usepackage{thmtools,thm-restate}
\usepackage{bm}
\usepackage{enumitem,array}
\usepackage{algorithm}
\usepackage{algpseudocode}
\usepackage{bbm}
\usepackage{tabulary,multirow}
\usepackage{dsfont}
\usepackage{thmtools,thm-restate}
\usepackage[capitalize,noabbrev]{cleveref}

\usepackage{multicol}
\usepackage{authblk}
\usepackage{upgreek}
\usepackage{tabularx}
\usepackage{lipsum}
\usepackage{enumitem}
\usepackage{authblk}

\usepackage[font=small,labelfont=bf]{caption}

\usepackage{wrapfig}
\usepackage{lipsum}

\usepackage[normalem]{ulem}

\theoremstyle{plain}
\newtheorem{theorem}{Theorem}[section]

\newtheorem{lemma}[theorem]{Lemma}
\newtheorem{corollary}[theorem]{Corollary}
\theoremstyle{definition}
\newtheorem{definition}[theorem]{Definition}
\newtheorem{assumption}[theorem]{Assumption}
\newtheorem{remark}[theorem]{Remark}

\usepackage[textsize=tiny]{todonotes}

\usepackage[acronym]{glossaries} 

\usepackage[frozencache,cachedir=.]{minted}
\definecolor{LightGray}{gray}{0.9}

\hypersetup{%
    colorlinks=true,
    linkcolor=mydarkblue,
    citecolor=mydarkblue,
    filecolor=mydarkblue,
    urlcolor=mydarkblue,
}
\definecolor{mydarkblue}{rgb}{0,0.08,0.45}

\makeglossaries
\newacronym{DNN}{DNN}{Deep Neural Networks}
\newacronym{NTK}{NTK}{Neural Tangent Kernel}
\newacronym{RKHS}{RKHS}{Reproducing kernel Hilbert space}
\newacronym{MLE}{MLE}{Maximum Likelihood Estimation}
\newacronym{SOTA}{SOTA}{State-of-the-art}
\newacronym{MSE}{MSE}{Mean Square Error}
\newacronym{OFUL}{OFUL}{Optimism in the Face of Uncertainty}
\newacronym{UCB}{UCB}{Upper Confidence Bound}
\newacronym{LinUCB}{LinUCB}{Linear Upper Confidence Bound}
\newacronym{SAVE}{SAVE}{Suplin + Adaptive Variance-aware Exploration}
\newacronym{NeuralUCB}{NeuralUCB}{Neural Upper Confidence Bound}
\newacronym{NeuralTS}{NeuralTS}{Neural Thompson Sampling}
\newacronym{NeuralLinUCB}{Neural-LinUCB}{Neural Linear Upper Confidence Bound}
\newacronym{ours}{Neural-$\sigma^2$-LinUCB}{Neural Variance-Aware Linear Upper Confidence Bound}

\usepackage{amssymb}
\usepackage{pifont}
\newcommand{\cmark}{\ding{51}}%
\newcommand{\xmark}{\ding{55}}%

\begin{document}

\runningtitle{Variance-Aware Linear UCB with Deep Representation for Neural Contextual Bandits}
\twocolumn[
\aistatstitle{Variance-Aware Linear UCB with Deep Representation for Neural Contextual Bandits}
\aistatsauthor{  Ha Manh Bui \And Enrique Mallada \And Anqi Liu }
\aistatsaddress{ Johns Hopkins University, Baltimore, MD, U.S.A. }]

\begin{abstract}
By leveraging the representation power of deep neural networks, neural upper confidence bound (UCB) algorithms have shown success in contextual bandits. To further balance the exploration and exploitation, we propose Neural-$\sigma^2$-LinearUCB, a variance-aware algorithm that utilizes $\sigma^2_t$, i.e., an upper bound of the reward noise variance at round $t$, to enhance the uncertainty quantification quality of the UCB, resulting in a regret performance improvement. We provide an oracle version for our algorithm characterized by an oracle variance upper bound $\sigma^2_t$ and a practical version with a novel estimation for this variance bound. Theoretically, we provide rigorous regret analysis for both versions and prove that our oracle algorithm achieves a better regret guarantee than other neural-UCB algorithms in the neural contextual bandits setting. Empirically, our practical method enjoys a similar computational efficiency, while outperforming state-of-the-art techniques by having a better calibration and lower regret across multiple standard settings, including on the synthetic, UCI, MNIST, and CIFAR-10 datasets.
\end{abstract}
\section{Introduction}
The stochastic multi-armed contextual bandits is a sequential decision-making problem that is related to various real-world applications, e.g., healthcare, finance, recommendation, etc. Specifically, this setting considers the interaction between an agent and an environment. In each round, the agent receives a context from the environment and then decides based on a finite arm set. After each decision, the agent receives a reward and its goal is to maximize the cumulative reward over rounds~\citep{sutton1998RL}.

\begin{table}[t!]
    \centering
    \hspace*{-0.15in}
    \scalebox{0.8}{
    \begin{tabular}{cccc}\\
    \toprule  
    Method &$\begin{matrix}
    \text{High \acrshort{UCB}}\\
    \text{quality with $\sigma_t^2$}
    \end{matrix}$ & $\begin{matrix}
    \text{Neural-regret}\\
    \text{analysis}
    \end{matrix}$ & $\begin{matrix}
    \text{Empirical} \\
    \text{efficiency} 
    \end{matrix}$\\\midrule
    NeuralUCB & \xmark & \cmark  & \xmark  \\
    Neural-LinUCB & \xmark & \cmark & \cmark \\
    Variance-aware-\acrshort{UCB} & \cmark & \xmark & \xmark \\
    Ours & \cmark  & \cmark & \cmark \\
    \bottomrule
    \end{tabular}}
    \caption{Contribution comparison between methods in utilizing variance bound $\sigma_t^2$ to improve \acrshort{UCB} uncertainty quality, neural-regret analysis, and empirical efficiency.}
    \vspace{-0.1in}
    \label{tab:teaser}
\end{table}

\begin{figure*}[t!]
    \centering
     \setlength{\tabcolsep}{0pt}
    \begin{tabular}{ccc}
    \includegraphics[width=0.33\linewidth]{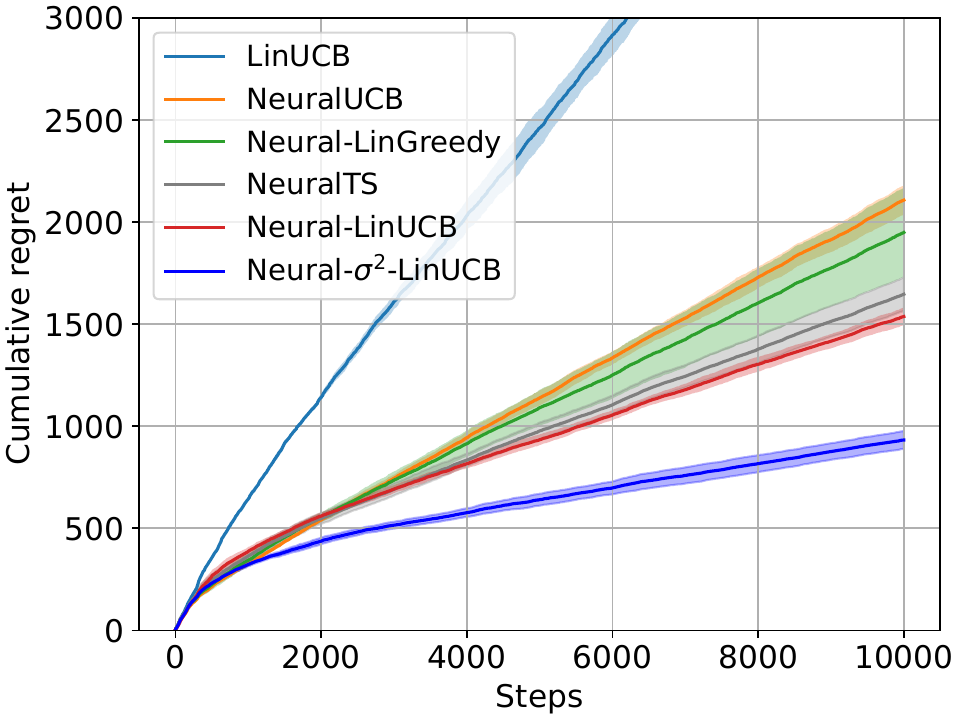}&
    \includegraphics[width=0.33\linewidth]{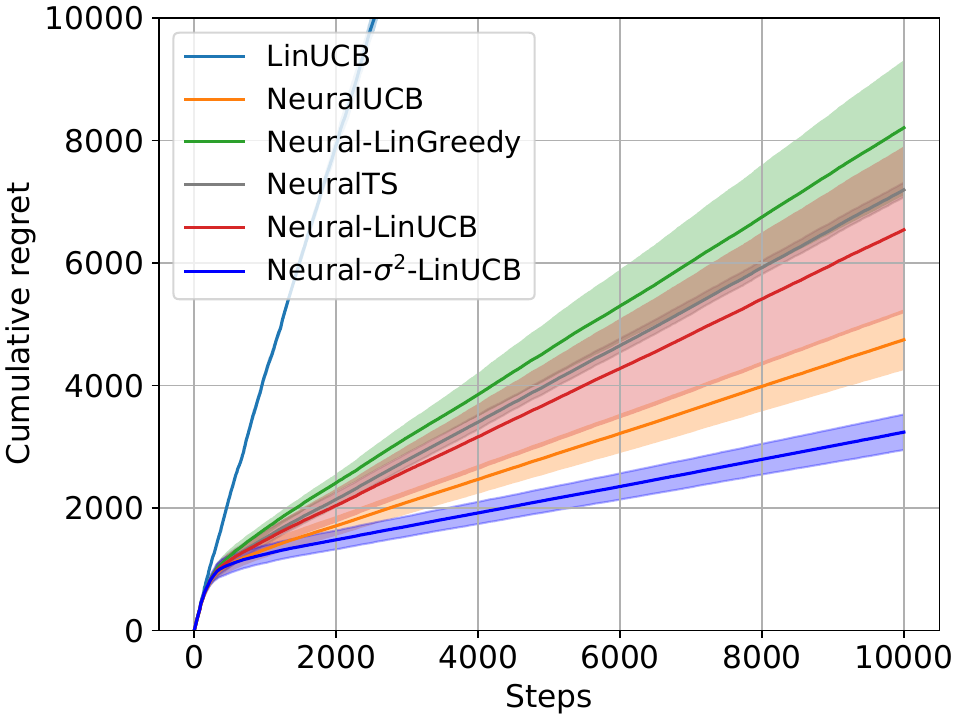}&
    \includegraphics[width=0.33\linewidth]{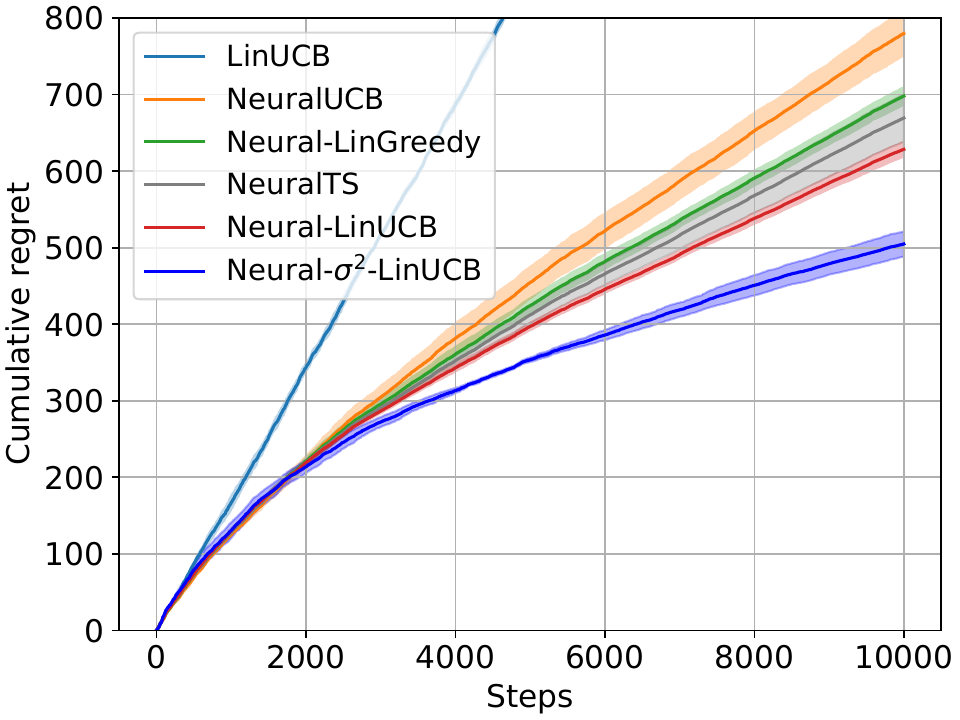}\\
        \scriptsize (a) $h_1(\mathbf{x}) = 10(\mathbf{x}^\top \bm{\theta})^2$ & \scriptsize (b) $h_2(\mathbf{x}) = \mathbf{x}^\top \bm{\theta}^\top \bm{\theta} \mathbf{x}$ & \scriptsize (c) $h_3(\mathbf{x}) = \cos(3\mathbf{x}^\top \bm{\theta})$
    \end{tabular}
    \caption{Cumulative regret results on the synthetic data across 10 runs with different seeds. More baselines for comparison and a zoom-in figure are in Fig.~\ref{fig:full_demo}. \textit{A short demo is available at \href{https://colab.research.google.com/drive/1WmuMocSfIxtfzSrdSZvVRgYNaM6ePgNN?usp=sharing}{this Google Colab link}.}}
    \label{fig:demo}
\end{figure*}

To balance the exploration and exploitation, several algorithms for this setting have been proposed~\citep{lattimore2017bandit,bubeck2012Regret}. Among these methods, based on the principle of \acrfull{OFUL} and the power of \acrfull{DNN}, \acrfull{NeuralUCB}~\citep{zhou2020neuralUCB} and \acrfull{NeuralLinUCB}~\citep{xu2022neural} have become the most practical and are the \acrfull{SOTA} techniques. Specifically, \acrshort{NeuralUCB} is a natural extension of \acrfull{LinUCB}~\citep{liICWWWlinearUCB,chu2011contextual}, which uses a \acrshort{DNN}-based random feature mapping to approximate the underlying reward function. Yet, it is computationally inefficient since the \acrfull{UCB} is performed over the entire \acrshort{DNN} parameter space. \acrshort{NeuralLinUCB} improves the efficiency by learning a mapping that transforms the raw context input into feature vectors using a \acrshort{DNN}, and then performing a \acrshort{UCB} exploration over the linear output layer of the network. These methods achieve $\tilde{\mathcal{O}}(Rd\sqrt{T})$ regret upper bound, where $R$ is the upper bound of the absolute value of the reward noise, $d$ is the feature context dimension, and $T$ is the learning time horizon. This is equivalent to the result of \acrshort{LinUCB} in the linear contextual bandits setting~\citep{yasin2011improved}.

The predictive uncertainty of the \acrshort{UCB}, especially when derived from modern \acrshort{DNN}, however, can be inaccurate and impose a bottleneck on the regret performance~\citep{kuleshov2018accurate}. To tackle this challenge, the idea of improving \acrshort{UCB} uncertainty estimation quality to enhance regret performance has shown promising results~\citep{kuleshov2014algorithms, auer2002finitetime}. 
Notably,~\citet{calibrated2019malik,deshpande2024online} have shown that calibrated neural-\acrshort{UCB} algorithms can result in a lower cumulative regret. Yet, they require a post-hoc re-calibration step on additional hold-out data for every round, leading to inefficiency in practice. Theoretically, in the linear contextual bandits setting, recent works have shown that variance-aware-\acrshort{UCB} algorithms~\citep{zhou2021nearly,zhao2023variance}, i.e., using the reward noise variance to improve uncertainty estimation quality of \acrshort{UCB}, can further achieve a tighter regret bound than \acrshort{LinUCB}. However, even with this non-neural-network approach, estimating the true variance is non-trivial, and such algorithms are often not practically feasible. As a result, there are usually no experimental results shown in the previous literature for this variance-aware-\acrshort{UCB} domain. 

Therefore, towards a variance-aware neural-\acrshort{UCB} algorithm that is both rigorous and practical, we propose \acrfull{ours}. Since estimating the true variance with \acrshort{DNN} is challenging, \acrshort{ours} leverages $\sigma^2_t$, i.e., the upper bound of the reward noise variance at round $t$, to enhance the uncertainty quantification quality of the \acrshort{UCB}, resulting in a regret performance improvement. We propose two versions, including an oracle version that uses a given variance upper bound $\sigma^2_t$ and a practical version that estimates this variance bound. We formally provide regret guarantees for both versions and prove our oracle version achieves a tighter regret guarantee with \acrshort{DNN} than other neural-\acrshort{UCB} bandits. Succinctly, for each round, our practical version calculates the upper bound of the reward noise variance by using the reward range and the estimated reward mean with \acrshort{DNN}. Then, we use this variance-bound information to optimize the linear reward model w.r.t. encoded \acrshort{DNN} context features by using a weighted ridge regression minimizer. 

The key ideas of this approach are: (1) \acrshort{UCB} is performed over the feature representation from the last \acrshort{DNN} layer. Therefore, it enjoys computational efficiency of \acrshort{NeuralLinUCB}; (2) When $\sigma_t^2$ is large, our \acrshort{UCB} will be more uncertain, and vice versa. This intuitively helps improve uncertainty estimation quality of \acrshort{UCB}, resulting in a better regret guarantee.

Our theoretical and practical contributions are summarized in Tab.~\ref{tab:teaser} and are as follows:
\begin{itemize} [itemsep=1pt,topsep=0pt,parsep=0pt,leftmargin=*]
    \item We propose \acrshort{ours}, a variance-aware algorithm that utilizes $\sigma^2_t$ to enhance the exploration-exploitation quality of \acrshort{UCB}. We provide an oracle and a practical version. The oracle algorithm assumes knowledge on $\sigma_t^2$. The practical algorithm estimates $\sigma_t^2$ from the reward range and the reward mean estimator with \acrshort{DNN}.
    \item We prove the regret of our practical version is at most \small$\tilde{\mathcal{O}} \left(R\sqrt{dT} + d\sqrt{\sum_{t=1}^T \sigma_t^2 + \epsilon}\right)$\normalsize, where $\epsilon$ is the estimation error of $\sigma_t^2$. Notably, our oracle version achieves \small$\tilde{\mathcal{O}}\left(R\sqrt{dT}+d\sqrt{\sum_{t=1}^T\sigma_t^2}\right)$\normalsize regret bound. Since our setting considers 
    $\sigma_t\leq R$, this is strictly better than \small$\tilde{\mathcal{O}}\left(Rd\sqrt{T}\right)$\normalsize of \acrshort{NeuralLinUCB}. 
    \item We empirically show our proposed method enjoys a similar computational efficiency while outperforming \acrshort{SOTA} techniques by having a better calibration and lower regret across multiple contextual bandits settings, including on the synthetic, UCI, MNIST, and CIFAR-10 datasets (e.g., Fig.~\ref{fig:demo}). 
\end{itemize}
\section{Background}
\textbf{Notation}. We denote $[k]$ is a set $\{1,\cdots,k\}$, $k\in \mathbb{N}$. For a semi-definite matrix $\mathbf{A}\in \mathbb{R}^{d\times d}$ and a vector $\mathbf{x}\in \mathbb{R}^d$, let $\left\|\mathbf{x}\right\|_{\mathbf{A}}=\sqrt{\mathbf{x}^\top \mathbf{A} \mathbf{x}}$ be the Mahalanobis norm. For a complexity $\mathcal{O}(T)$, let us use $\tilde{\mathcal{O}}(T)$ to hide the constant and logarithmic dependence of $T$. We also use $\mathcal{N}(\cdot)$ to denote the Gaussian distribution and $\mathbb{U}(\cdot)$ for the Uniform distribution.

\subsection{Problem setting}
In the stochastic $K$-armed contextual bandits~\citep{lattimore2017bandit}, at each round $t \in [T]$, the learning agent observes a context consisting of $K$  feature vectors $\{\mathbf{x}_{t,a}\in \mathbb{R}^d\mid a\in [K]\}$ from the environment, then selects an arm $a_t \in [K]$ based on this context, and receives a corresponding reward $r_{t,a_t}$. The agent aims to maximize its expected total reward over these $T$ rounds, i.e., minimizing the pseudo-regret
\begin{align}
    &\textstyle \text{Regret}(T) = \mathbb{E}\left[\sum_{t=1}^T \left(r_{t,a_t^*} - r_{t,a_t}\right) \right],
\end{align}
where $a_t^* = \arg\max_{a\in [K]} \mathbb{E}\left[r_{t,a}\right]$. Following~\citet{zhou2020neuralUCB,xu2022neural}, for any round $t$, we assume the reward generation, defined as follows
\begin{align}\label{eq:reward_gen}
    r_{t,a_t} = h(\mathbf{x}_{t,a_t}) + \xi_t,
\end{align}
where $h$ is an unknown function s.t. $0\leq h(\mathbf{x}) \leq 1, \forall \mathbf{x}$. In terms of the reward noise, following~\citet{zhou2021nearly,yasin2011improved,zhao2023variance}, we assume $\xi_t$ is a random noise variable that satisfies the following conditions
\begin{align}\label{eq:noise_gen} 
    &p(|\xi_t| \leq R) = 1,\quad
    \mathbb{E}[\xi_t\mid \mathbf{x}_{1:t, a_{1:t}}, \xi_{1:t-1}] = 0,\nonumber\\
    &\mathbb{E}[\xi^2_t\mid \mathbf{x}_{1:t, a_{1:t}}, \xi_{1:t-1}] 
        \leq \sigma^2_{t} \leq R^2.
\end{align}

\subsection{Neural Linear Upper Confidence Bound}
To relax the strong linear-reward assumption, we consider a setting that the unknown function $h$ can be non-linear. Our work builds on \acrshort{NeuralLinUCB}~\citep{xu2022neural}, which seeks to extend \acrshort{LinUCB} by leveraging the approximating power of \acrshort{DNN}. In particular, for a neural network
\small
\begin{align}
    f(\mathbf{x};\bm{\theta}^*,\mathbf{w}) = \sqrt{m}{\bm{\theta}^*}^\top g_L (\mathbf{W}_L g_{L-1}(\mathbf{W}_{L-1} \cdots g_1 (\mathbf{W}_1 \mathbf{x}))),
\end{align}
\normalsize
where $\mathbf{x} \in \mathbb{R}^d$ is the input data, $\bm{\theta}^* \in \mathbb{R}^d$ is the weight vector of the output layer, $\mathbf{w}=(Vec({\mathbf{W}_1})^\top, \cdots, Vec({\mathbf{W}_L})^\top)^\top$, $\mathbf{W}_l \in \mathbb{R}^{m_l \times m_{l-1}}$ is the weight matrix of the $l$-th layer, $l\in [L]$, and $g_l=g$ is the ReLU activation function, i.e., $g(x) = \max\{0,x\}$ for $x \in \mathbb{R}$. By further assuming that  $m_1=\cdots = m_{L-1}=m$, $m_0=m_L=d$, one can readily show that the dimension $p$ of vector $\mathbf{w}$ satisfies $p=(L-2)m^2 + 2md$ and the output of the $L$-th hidden layer of neural network $f$ becomes
\small
\begin{align}\label{def:NN}
    \bm{\phi}(\mathbf{x};\mathbf{w}) = \sqrt{m} g(\mathbf{W}_L g(\mathbf{W}_{L-1} \cdots g(\mathbf{W}_1 \mathbf{x}))).
\end{align}
\normalsize
Then, at round $t$, the agent model chooses the action that maximizing the \acrshort{UCB} as follows
\scriptsize
\begin{align}\label{eq:UCB}
    a_t = \arg\max_{k\in [K]}\{\left \langle \bm{\phi}(\mathbf{x}_{t,k};\mathbf{w}_{t-1}), \bm{\theta}_{t-1} \right \rangle + \alpha_t \left\|\bm{\phi}(\mathbf{x}_{t,k};\mathbf{w}_{t-1})\right\|_{\mathbf{A}_{t-1}^{-1}}\},
\end{align}
\normalsize
where the output layer weights $\bm{\theta}_{t-1}$ is updated by using the same ridge regression as in linear contextual bandits~\citep{yasin2011improved}, i.e.,  we consider $\bm{\theta}_t = \mathbf{A}_t^{-1} \mathbf{b}_t$, with 
\small
\begin{equation}\label{eq:a_t_b_t}
    \setlength{\jot}{0.01pt}
    \begin{aligned}
    &\mathbf{A}_t = \lambda \mathbf{I} + \sum_{i=1}^t \bm{\phi}(\mathbf{x}_{i,a_i};\mathbf{w}_{i-1}) \bm{\phi}(\mathbf{x}_{i,a_i};\mathbf{w}_{i-1})^\top,\\
    &\mathbf{b}_t = \sum_{i=1}^t r_{i,a_i} \bm{\phi}(\mathbf{x}_{i,a_i};\mathbf{w}_{i-1}).
    \end{aligned}
\end{equation}
\normalsize
Finally, the \acrshort{DNN} model weights $\mathbf{w}$ are optimized every $H$ time steps, i.e., at times $t=qH$, with $q = 1,2,\cdots$, following the Empirical risk minimization algorithm with the \acrfull{MSE} loss function
\begin{align}\label{eq:MSE}
&\textstyle \mathcal{L}_q(\mathbf{w}) = \sum_{i=1}^{qH} \left(\bm{\theta}_t^\top \bm{\phi}(\mathbf{x}_{i,a_i}; \mathbf{w}) - r_{i,a_i} \right)^2.
\end{align}

By using \acrshort{NTK}~\citep{jacot2018NTK}, \acrshort{NeuralLinUCB} is proven to achieve $\tilde{\mathcal{O}}(Rd\sqrt{T}) + \tilde{\mathcal{O}}\left(m^{-1/6}T\sqrt{(\mathbf{r}-\mathbf{\tilde{r}})^\top \mathbf{H}^{-1} (\mathbf{r}-\mathbf{\tilde{r}})}\right)$ regret~\citep{xu2022neural}, where the first term resembles the regret bound of \acrshort{LinUCB}~\citep{yasin2011improved}. Meanwhile, the second term depends on the estimation error of the neural network $f$ for the reward-generating function $\mathbf{r}$, its estimation $\mathbf{\tilde{r}}$, and the \acrshort{NTK} matrix $\mathbf{H}$ (we relegate to Sec.~\ref{sec:theory} for the precise definition of $\mathbf{r}$, $\mathbf{\tilde{r}}$, and $\mathbf{H}$). Following the assumption that $\mathbf{r}^\top \mathbf{H}^{-1} \mathbf{r}$ can be upper bounded by a constant (can be bounded by the \acrshort{RKHS} norm of $\mathbf{r}$ if it belongs to the \acrshort{RKHS} induced by $\mathbf{H}$), i.e., $\left\|\mathbf{r}-\mathbf{\tilde{r}}\right\|_{\mathbf{H}^{-1}} = \mathcal{O}(1)$~\citep{zhou2020neuralUCB}, and by a selection of $m\geq T^3$~\citep{xu2022neural}, then the final regret of \acrshort{NeuralLinUCB} becomes $\tilde{\mathcal{O}}(Rd\sqrt{T})$.

\section{Neural Variance-Aware Linear Upper Confidence Bound Algorithm}
\begin{figure}[ht!]
    \centering
    \includegraphics[width=1.0\linewidth]{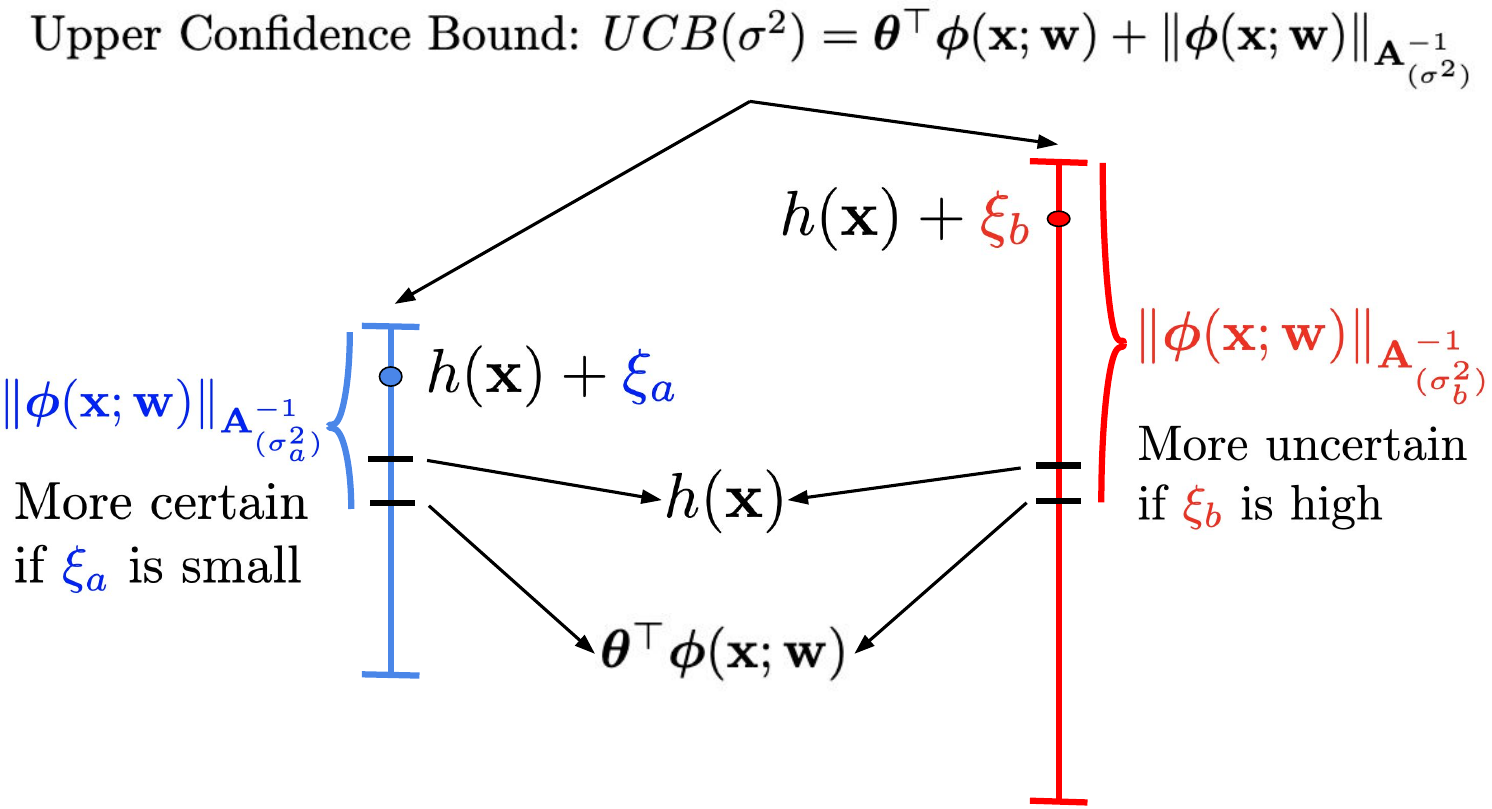}
    \caption{Our \acrshort{ours} can be more uncertain (i.e., more exploration) if the reward noise $Var(\xi_t)$ is high, and more certain (i.e., more exploitation) if $Var(\xi_t)$ is small.}
    \label{fig:framework}
\end{figure}
\subsection{Oracle algorithm}\label{sec:oracle}
To improve the \acrshort{UCB} quality and regret guarantee in the non-linear contextual bandits, we propose the oracle \acrshort{ours}. The main idea of our method is using the high-quality feature representation $\bm{\phi}(\mathbf{x}_{t,a_t};\mathbf{w})$ to estimate the mean reward $\left \langle \bm{\theta}, \bm{\phi}(\mathbf{x}_{t,a_t};\mathbf{w}) \right \rangle$ and the upper bound of the reward noise variance $\sigma_t^2$ per round $t$. Then, based on the \acrshort{OFUL}, we make use of this variance upper bound information to optimize the linear model $\bm{\theta}$ w.r.t. encoded \acrshort{DNN} context feature $\bm{\phi}(\mathbf{x}_{t,a_t};\mathbf{w})$ by minimizing to the weighted ridge regression objective function as follows
\small
\begin{align}\label{eq:weight_L2}
    \bm{\theta}_t = \arg\min_{\bm{\theta} \in \mathbb{R}^d} \lambda ||\bm{\theta}||_2^2 + \sum_{i=1}^t \frac{\left[ \left \langle \bm{\theta}, \bm{\phi}(\mathbf{x}_{i,a_i};\mathbf{w}_{i-1}) \right \rangle - r_{i,a_i} \right]^2}{\Bar{\sigma}_i^2},
\end{align}
\normalsize
where $\Bar{\sigma}_i = \max\{R/\sqrt{d}, \sigma_i\}$. Therefore, by computing the optimality conditions of Eq.~\ref{eq:weight_L2}, it follows that $\bm{\theta}_t = \mathbf{A}_t^{-1} \mathbf{b}_t$, where $\mathbf{A}_t$, which depends on the historical context-arm pairs, and the bias term $\mathbf{b}_t$ are given by
\small
\begin{equation}\label{eq:a_t_b_t_ours}
    \setlength{\jot}{0.01pt}
    \begin{aligned}
    &\mathbf{A}_t = \lambda \mathbf{I} + \sum_{i=1}^t \frac{\bm{\phi}(\mathbf{x}_{i,a_i};\mathbf{w}_{i-1}) \bm{\phi}(\mathbf{x}_{i,a_i};\mathbf{w}_{i-1})^\top}{\Bar{\sigma}_i^2},\\ 
    &\mathbf{b}_t = \sum_{i=1}^t \frac{r_{i,a_i} \bm{\phi}(\mathbf{x}_{i,a_i};\mathbf{w}_{i-1})}{\Bar{\sigma}_i^2}.
    \end{aligned}
\end{equation}
\normalsize
The pseudo-code for \acrshort{ours} is presented in Alg.~\ref{alg:ours}. From the solution of $\mathbf{A}_t$ and $b_t$ above, we can see that our feature matrix $\mathbf{A}_t$ is weighted by the proxy $\Bar{\sigma}_i^2$ of the reward variance upper bound $\sigma_i^2$. 

\begin{remark}\label{rem:algorithmic}
(Algorithmic comparison between \acrshort{ours} and \acrshort{NeuralLinUCB}).
Consider the confidence set $\mathcal{E}_t = \{\bm{\theta}\in \mathbb{R}^d:||\bm{\theta} - \bm{\theta}_t||_{\mathbf{A}_t}^2\}$, which is an ellipsoid centred at $\bm{\theta}_t$ and with principle axis being the eigenvectors of $\mathbf{A}_t$ with
corresponding lengths being the reciprocal of the eigenvalues. Compare our $\mathbf{A}_t$ solution in Eq.~\ref{eq:a_t_b_t_ours} versus the solution in Eq.~\ref{eq:a_t_b_t}, we can see that when $t$ grows, the matrix $\mathbf{A}_t$ in Eq.~\ref{eq:a_t_b_t} has increasing eigenvalues, which means the volume of the ellipse is also frequently shrinking. Meanwhile, our $\mathbf{A}_t$ solution in Eq.~\ref{eq:a_t_b_t_ours} is more flexible by depending on the variance bound $\sigma_t^2$. This means that the volume of the ellipse will shrink not too fast if $\sigma_t^2$ is high, and not too slow if $\sigma_t^2$ is small, suggesting an exploration and exploitation improvement of the \acrshort{UCB}.
\end{remark}

At a high level, our oracle algorithm can be seen as a combination of Weighted~\acrshort{OFUL} and \acrshort{NeuralLinUCB}. Yet, its challenges include: (1) It is unclear whether this can bring out a tighter regret bound than \acrshort{NeuralLinUCB}; (2) It assumes we are given $\sigma_t^2$ at round $t$ while $\sigma_t^2$ is often unavailable and is an unknown quantity in practice. Hence, we address the challenge (1) in Thm.~\ref{theo:regret_upper_bound} in Sec.~\ref{sec:theory}. Regarding challenge (2), we next propose a novel practical version to estimate $\sigma_t^2$.

\subsection{Practical algorithm}\label{sec:practical}
Since estimating the uncertainty of the true variance $Var(\xi_t)$ can be unreliable, especially when derived from \acrshort{DNN}~\citep{kuleshov2018accurate,calibrated2019malik}, we instead estimate the variance bound $\sigma_t^2$. As illustrated in Fig.~\ref{fig:framework}, our Alg.~\ref{alg:ours} intuitively means when the reward noise $Var(\xi_t)$ is high, $\sigma_t^2$ will be high, yielding a high $\mathbf{A}_t^{-1}$, i.e., more uncertainty for \acrshort{UCB}, and vice versa. This suggests a better \acrshort{UCB} uncertainty quantification, resulting in a better regret performance.

Recall $\sigma_t^2$ in Eq.~\ref{eq:noise_gen} is the upper bound of the reward noise variance and is bounded by the magnitude $R^2$, i.e., $\mathbb{E}[\xi^2_t\mid \mathbf{x}_{1:t, a_{1:t}},\xi_{1:t-1}] \leq \sigma^2_{t} \leq R^2$. Therefore, to estimate $\sigma_t^2$ to satisfy this condition, firstly, by the definition of the reward function in Eq.~\ref{eq:reward_gen} and the reward noise in Eq.~\ref{eq:noise_gen}, we can trivially derive to obtain the form of the mean and the variance of the reward by the theorem as follows:
\begin{theorem}\label{thm:reward_distribution}
    The reward r.v. $r_{t,a_t}$ in Eq.~\ref{eq:reward_gen} has the true mean $\mathbb{E}[r_{t,a_t}] = h(\mathbf{x}_{t,a_t})$ and variance $Var(r_{t,a_t}) = \mathbb{E}[\xi^2_t\mid \mathbf{x}_{1:t, a_{1:t}}, \xi_{1:t-1}]$. The proof is in Apd.~\ref{proof:thm:reward_distribution}.
\end{theorem}
By the mean and variance formulation in Thm.~\ref{thm:reward_distribution}, we can calculate the upper bound of the reward noise variance $\sigma_t^2$ at round $t$ by the following theorem:
\begin{theorem}\label{thm:variance_bound}
    If the reward r.v. $r_{t,a_t}$ is restricted to $[a,b]$ and we know the mean $h(\mathbf{x}_{t,a_t})$, then the variance is bounded by 
    \scriptsize
    \begin{align*}
        \mathbb{E}[\xi^2_t\mid \mathbf{x}_{1:t, a_{1:t}}, \xi_{1:t-1}] \leq \sigma_t^2 := (b- h(\mathbf{x}_{t,a_t}))(h(\mathbf{x}_{t,a_t})-a) \leq R^2.
    \end{align*}
    \normalsize
    The proof is in Apd.~\ref{proof:thm:variance_bound}.
\end{theorem}
From Thm.~\ref{thm:variance_bound}, we can see that given a reward range $[a,b]$, at round $t$, we can achieve a tighter upper bound of the reward noise variance than $R^2$. Hence, based on the estimation of the mean, i.e., $\bm{\theta}_{t-1}^\top \bm{\phi}(\mathbf{x}_{t,a_t})$, we can obtain the variance bound by calculating
\small
\begin{align}\label{eq:est_sigma}
    \hat{\sigma}^2_t = (b - \bm{\theta}_{t-1}^\top \bm{\phi}(\mathbf{x}_{t,a_t};\mathbf{w}_{t-1}))(\bm{\theta}_{t-1}^\top \bm{\phi}(\mathbf{x}_{t,a_t};\mathbf{w}_{t-1})-a).
\end{align}
\normalsize
The efficient estimation $\hat{\sigma}^2_t$ in Eq.~\ref{eq:est_sigma} is based on the estimation of the reward mean $\bm{\theta}_{t-1}^\top \bm{\phi}(\mathbf{x}_{t,a_t})$. So, as this estimation quality improves, our estimation quality for $\sigma^2_t$ will improve correspondingly. We visualize the quality of our estimation $\hat{\sigma}^2_t$ for $\sigma^2_t$ in Fig.~\ref{fig:calib}~(b). Furthermore, as discussed in Rem.~\ref{rem:algorithmic}, $\sigma^2_t$ intuitively can improve the uncertainty quantification quality of \acrshort{UCB}, we therefore also visualize the calibration performance of our \acrshort{UCB} with $\hat{\sigma}_t^2$ in Fig.~\ref{fig:calib}~(a). 

Eq.~\ref{eq:est_sigma} requires knowing the reward range $[a,b]$, which can be plausible in practice. For instance, in the real-world datasets in the experiment section, we may already know the range of $[a, b]$ when defining the reward function. Another real-world example is in a personalized recommendation website system~\citep{liICWWWlinearUCB}, the decision to optimize is articles to display to users; context is user data (e.g., browsing history); actions are available news articles; and reward is user engagement (click or no click). Hence, we can set the reward as $1$ if a user clicks and $0$ otherwise, i.e., $[a,b]=[0,1]$. It is also worth noticing that our Alg.~\ref{alg:ours} can be extended to handle cases where the reward range $[a,b]$ is dynamic, i.e., $[a,b]$ changes across rounds $t\in [T]$. We additionally show a result of this extension in Fig.~\ref{fig:ab_dynamic}~(b).

\textbf{Alg.~\ref{alg:ours} with \acrfull{MLE}}. Eq.~\ref{eq:MSE} back-propagate the neural network to update parameter $\mathbf{w}$ by using the \acrshort{MSE}. This objective function, however, only tries to improve the mean of the estimation from \acrshort{DNN}. To enhance the predictive uncertainty quality of the neural network~\citep{chua2018deepRL,tran2020methods}, we further propose Eq.~\ref{eq:MLE} to update parameter $\mathbf{w}$ via \acrshort{MLE}. Given our current model, the predictive variance of the expected payoff $\bm{\theta}_{t}^\top \bm{\phi}(\mathbf{x}_{i,a_i}; \mathbf{w})$ is evaluated as \small$\bm{\phi}(\mathbf{x}_{i,a_i}; \mathbf{w})^\top \mathbf{A}_t^{-1} \bm{\phi}(\mathbf{x}_{i,a_i}; \mathbf{w})$\normalsize. Therefore, we can formalize \acrshort{MLE} with the normal distribution via the following loss function
\small
\begin{align}\label{eq:MLE}
    \mathcal{L}_q(\mathbf{w}) &= \sum_{i=1}^{qH} \left[\frac{1}{2} \log\left(2\pi \cdot \bm{\phi}(\mathbf{x}_{i,a_i}; \mathbf{w})^\top \mathbf{A}_t^{-1} \bm{\phi}(\mathbf{x}_{i,a_i}; \mathbf{w}) \right)\right.\nonumber\\ 
    &\quad \left.+ \frac{\left[r_{i,a_i} - \bm{\theta}_t^\top \bm{\phi}(\mathbf{x}_{i,a_i}; \mathbf{w}) \right]^2}{2 \cdot \bm{\phi}(\mathbf{x}_{i,a_i}; \mathbf{w})^\top \mathbf{A}_t^{-1} \bm{\phi}(\mathbf{x}_{i,a_i}; \mathbf{w})}\right].
\end{align}
\normalsize

\begin{figure}[t!]
\vspace{-0.1in}
\begin{algorithm}[H]
    \caption{\small \acrshort{ours} (code is in Apd.~\ref{apd:code})}\label{alg:ours}
    \small
    \begin{algorithmic}[1]
        \State {\bfseries Input:} $\lambda, T, H, K, d, L, n, m, \{\alpha_t\}_{t\in [T]}$
        \State $\mathbf{A}_0 \leftarrow \lambda \mathbf{I}$, $\mathbf{b}_0 \leftarrow \mathbf{0}$, $\bm{\theta}_0 \sim \mathcal{N}(0,1/d)$ $\mathbf{w}_0\sim \mathbb{P}(\mathbf{w})$, $q\leftarrow 0$
        \For{$t=1\rightarrow T$}
            \State Observe $K$ feature $\mathbf{x}_{t,1},\cdots, \mathbf{x}_{t,K} \in \mathbb{R}^d$
            \For{$k=1\rightarrow K$}
                \scriptsize
                \State $p_{t,k} = \bm{\theta}_{t-1}^\top \bm{\phi}(\mathbf{x}_{t,k};\mathbf{w}_{t-1}) + \alpha_t \left\|\bm{\phi}(\mathbf{x}_{t,k};\mathbf{w}_{t-1})\right\|_{\mathbf{A}_{t-1}^{-1}}$ \hfill{\color{gray}$\triangleright$Compute \acrshort{UCB} in Eq.~\ref{eq:UCB}}
                \small
            \EndFor
            \State Chose $a_t = \arg\max_{a} p_{t,a}$ \hfill{\color{gray}$\triangleright$Select action by Eq.~\ref{eq:UCB}}
            \State Observe the corresponding reward $r_t$
            \State Receive $\sigma_{t}^2$ in the oracle algorithm, or estimate $\sigma_{t}^2$ by Eq.~\ref{eq:est_sigma} in the practical algorithm
            \State $\Bar{\sigma}_t^2 = \max(\sigma_t^2, R^2/d)$
            \State $\mathbf{A}_{t} \leftarrow \mathbf{A}_{t-1} + \frac{\bm{\phi}(\mathbf{x}_{t,a_t};\mathbf{w}_{t-1}) \bm{\phi}(\mathbf{x}_{t,a_t};\mathbf{w}_{t-1})^\top}{\Bar{\sigma}_{t}^2}$ 
            \State $\mathbf{b}_{t} \leftarrow \mathbf{b}_{t-1} + \frac{\bm{\phi}(\mathbf{x}_{t,a_t};\mathbf{w}_{t-1})r_t}{\Bar{\sigma}_{t}^2}$\\
            \State $\bm{\theta}_t \leftarrow \mathbf{A}_{t}^{-1}\mathbf{b}_{t}$\hfill{\color{gray}$\triangleright$Update linear model by Eq.~\ref{eq:a_t_b_t_ours}}
            \If{$\mod(t,H)=0$}
                \State Initialize $\mathbf{w}_q^{(0)} = \mathbf{w}_t$
                \For{$s=1\rightarrow n$}
                    \State $\mathbf{w}_q^{(s)} = \mathbf{w}_q^{(s-1)} - \eta_q \nabla_{\mathbf{w}}\mathcal{L}_q\left(\mathbf{w}_q^{(s-1)}\right)$
                \EndFor
                \State $\mathbf{w}_t\leftarrow \mathbf{w}_q^{(n)}$ \hfill{\color{gray}$\triangleright$Update neural model}
                \State $q\leftarrow q + 1$ \hfill{\color{gray}$\triangleright$Update epoch scheduler}
            \Else
                \State $\mathbf{w}_t\leftarrow \mathbf{w}_{t-1}$
            \EndIf
        \EndFor
    \end{algorithmic}
    \normalsize
\end{algorithm}
\end{figure}
\section{Theoretical analysis}\label{sec:theory}
To analyze the regret for \acrshort{ours}, for convenience in analysis for the context $\mathbf{x}$~\citep{zou2019anImproved},  we first apply the following transformation inspired by existing work~\citep{zhu2019aConvergence,zhou2020neuralUCB} to ensure arm contexts are of unit length. In particular, without loss of generality:
\begin{remark}\label{rem:context_norm}(Arm context normalization).
     With unprocessed context $\tilde{\mathbf{x}}_{i,k}$, we formulate the corresponding normalized arm context $\mathbf{x}_{i,k}$ by $\mathbf{x}_{i,k}=\left[\frac{\tilde{\mathbf{x}}_{i,k}}{2\cdot \left\|\tilde{\mathbf{x}}_{i,k}\right\|_2}, \frac{1}{2}, \frac{\tilde{\mathbf{x}}_{i,k}}{2\cdot \left\|\tilde{\mathbf{x}}_{i,k}\right\|_2}, \frac{1}{2}\right]$ to achieve $\left\|\mathbf{x}_{i,k}\right\|_2=1$, for all $i\geq 1$ and $k\in [K]$. Then, for any context $\left\|\mathbf{x}_{i,k}\right\|_2=1$, we could replace $\mathbf{x}_{i,k}$ by $\mathbf{x}'_{i,k}=[\mathbf{x}_{i,k}^\top, \mathbf{x}_{i,k}^\top]^\top/\sqrt{2}$ to verify its entries satisfy $[\mathbf{x}_{i,k}]_j=[\mathbf{x}_{i,k}]_{j+d/2}$.
\end{remark}

Then, we follow two main assumptions from~\citet{zhou2020neuralUCB,xu2022neural} for the results in this section to hold. Specifically, the first assumption is about the stability condition on the spectral norm of the neural network gradient~\citep{zhaoran2014optimal,balakrishnan52017Statistical,xu2017speeding}:
\begin{assumption}\label{assm:grad_norm}
    For a specific weights parameters $\mathbf{w}_0$, $\exists \ell_{Lip}>0$ s.t. $\left \| \frac{\partial \bm{\phi}}{\partial \mathbf{w}}(\mathbf{x}, \mathbf{w}_0) - \frac{\partial \bm{\phi}}{\partial \mathbf{w}}(\mathbf{x}', \mathbf{w}_0) \right \|_2 \leq \ell_{Lip}\left \| \mathbf{x} - \mathbf{x}' \right \|_2$, $\forall \mathbf{x},\mathbf{x}' \in \{\mathbf{x}_{i,k}\}_{i\in [T],k\in [K]}$.
\end{assumption}
As discussed in~\citet{xu2022neural}, Asm.~\ref{assm:grad_norm} is widely made in nonconvex optimization. Furthermore, it is also worth noticing that this assumption is only required on the $TK$ training data points and a specific weight parameter $\mathbf{w}_0$. Therefore, the conditions in Asm.~\ref{assm:grad_norm} will hold if the raw feature data lie in a certain subspace of $\mathbb{R}^d$. To describe our last assumption, it is necessary to describe the \acrshort{NTK} matrix $\mathbf{H}$.
\begin{definition}\label{def:NTK}~\citep{jacot2018NTK}
    Define $\mathbf{H} \in \mathbb{R}^{TK\times TK}$ be the \acrfull{NTK} matrix, based on all features vectors $\{\mathbf{x}_{t,k}\}_{t\in [T], k\in [K]}$, renumbered as $\{x_i\}_{i\in [TK]}$. Then for all $i,j\in [TK]$, each entry $\mathbf{H}_{ij} := \frac{1}{2}\left(\tilde{\mathbf{\Sigma}}^{(L)}(\mathbf{x}_i, \mathbf{x}_j) + \mathbf{\Sigma}^{(L)}(\mathbf{x}_i, \mathbf{x}_j)\right)$, where the covariance between two data point $\mathbf{x}, \mathbf{y}\in \mathbb{R}^d$ is given as follows:
    \small$\Tilde{\mathbf{\Sigma}}^{(0)}(\mathbf{x}, \mathbf{y}) = \mathbf{\Sigma}^{(0)}(\mathbf{x}, \mathbf{y}) = \mathbf{x}^\top \mathbf{y}$, $\mathbf{\Lambda}^{(l)}(\mathbf{x}, \mathbf{y}) = \begin{bmatrix}
        \mathbf{\Sigma}^{l-1}(\mathbf{x}, \mathbf{x}) & \mathbf{\Sigma}^{l-1}(\mathbf{x}, \mathbf{y}) \\
        \mathbf{\Sigma}^{l-1}(\mathbf{y}, \mathbf{x}) & \mathbf{\Sigma}^{l-1}(\mathbf{y}, \mathbf{y})\\
    \end{bmatrix}$, $\mathbf{\Sigma}^{(l)}(\mathbf{x}, \mathbf{y}) = 2\mathbb{E}_{(u,v)\sim \mathcal{N}\left(\mathbf{0}, \mathbf{\Lambda}^{(l-1)}(\mathbf{x}, \mathbf{y})\right)}[g(u)g(v)]$, $\Tilde{\mathbf{\Sigma}}^{(l)}(\mathbf{x}, \mathbf{y}) = 2\Tilde{\mathbf{\Sigma}}^{(l-1)}(\mathbf{x}, \mathbf{y})\mathbb{E}_{u,v}[g'(u)g'(v)] + \mathbf{\Sigma}^{(l)}(\mathbf{x}, \mathbf{y})$\normalsize,
    with $(u,v)\sim \mathcal{N}\left(\mathbf{0}, \mathbf{\Lambda}^{(l-1)}(\mathbf{x}, \mathbf{y})\right)$, and $g'(\cdot)$ being the derivative of the activation function $g$. 
\end{definition}
The last assumption essentially requires the \acrshort{NTK} matrix $\mathbf{H}$ to be non-singular~\citep{du2019gradient,arora2019onExact,cao2019genbounds}:
\begin{assumption}\label{assm:NTK}
     The \acrshort{NTK} matrix $\mathbf{H}$ is positive definite, i.e., $\lambda_{\min}(\mathbf{H})\geq \lambda_0$ for some constants $\lambda_0>0$.
\end{assumption}
Asm.~\ref{assm:NTK} could be mild since we can derive from Rem.~\ref{rem:context_norm} with two ReLU layers~\citep{zou2019anImproved, xu2022neural}. We use Asm.~\ref{assm:NTK} to characterize the properties of \acrshort{DNN} to represent the feature vectors. Following these assumptions, we next provide the regret bound for our oracle \acrshort{ours} algorithm:
\begin{theorem}\label{theo:regret_upper_bound}
    Suppose Asm~\ref{assm:grad_norm}, and~\ref{assm:NTK} hold and further assume that $\left\|\theta^*\right\|_2 \leq M$, $\left\|\mathbf{x}_t\right\|_2 \leq G$, and $\lambda\geq \max\{1,G^2\}$ for some $M,G>0$. For any $\delta \in (0,1)$, let us choose $\alpha_t$ as
    \small
    \begin{align*}
    \alpha_t &= 8\sqrt{d\log \left(1+t d(\log HK)/(\bar{\sigma}_{t}^2d\lambda)\right)\log(4t^2/\delta)}\\ 
    &\quad + 4R/\bar{\sigma}_t \log(4t^2/\delta) + \lambda^{1/2}M,
    \end{align*}
    \normalsize
    the step size $\eta_q \leq C_0\left(d^2mnT^{5.5}L^6\log(TK/\delta)\right)^{-1}$, and the neural network width satisfies $m=poly(L,d,1/\delta,H,\log(TK/\delta))$, then,
    with probability at least $1-\delta$ over the randomness of the neural network initialization, the regret of the oracle algorithm satisfies
    \small
    \begin{align*}
        &\text{Regret}(T) \leq C_1 \alpha_T \sqrt{\left(TR^2+d\sum_{t=1}^T \sigma_t^2\right) \log(1+TG^2/(\lambda R^2))}\\ 
        &\quad + \frac{C_2 \ell_{Lip}L^3d^{5/2}T\sqrt{\log m \log(\frac{1}{\delta})\log(\frac{TK}{\delta})}\left\|\mathbf{r}-\mathbf{\tilde{r}}\right\|_{\mathbf{H}^-1}}{m^{1/6}},
    \end{align*}
    \normalsize
    where constants $\{C_i\}_{i\in [2]}$ are independent of the problem, $\mathbf{r} = \left(r(\mathbf{x}_1), r(\mathbf{x}_2), \cdots, r(\mathbf{x}_{TK})\right)^\top \in \mathbb{R}^{TK}$, and $\mathbf{\tilde{r}} = \left(f(\mathbf{x}_1;\theta_0, \mathbf{w}_0), \cdots, f(\mathbf{x}_{TK};\theta_{T-1}, \mathbf{w}_{T-1})\right)^\top \in \mathbb{R}^{TK}$.
\end{theorem}
The proof for Thm.~\ref{theo:regret_upper_bound} adapts the techniques of the Bernstein inequality for vector-valued martingales over the linear output \acrshort{DNN} last layer from~\citet{zhou2021nearly} and the \acrshort{NTK} for the raw context-feature \acrshort{DNN} mapping from~\citet{xu2022neural}, details are in Apd.~\ref{proof:theo:regret_upper_bound}. From Thm~\ref{theo:regret_upper_bound}, we obtain the following conclusion:
\begin{corollary}\label{corol:regret_upper_bound}
    Under the conditions in Thm.~\ref{theo:regret_upper_bound}, then, with probability at least $1-\delta$, the regret of the oracle algorithm is bounded by
    \small
    \begin{align*}
        \text{Regret}(T) &\leq \tilde{\mathcal{O}}\left(R\sqrt{dT}+d\sqrt{\sum_{t=1}^T\sigma_t^2}\right)\\
        &\quad+ \tilde{\mathcal{O}}\left(m^{-1/6}T\sqrt{(\mathbf{r}-\mathbf{\tilde{r}})^\top \mathbf{H}^{-1} (\mathbf{r}-\mathbf{\tilde{r}})}\right).
    \end{align*}
    \normalsize
\end{corollary}

\begin{remark}(Regret comparison between \acrshort{ours} and previous methods).
    Our second regret term resembles the second regret term bound of \acrshort{NeuralLinUCB}, which we can assume to have a constant bound for $\left\|\mathbf{r}-\mathbf{\tilde{r}}\right\|_{\mathbf{H}^{-1}} = \mathcal{O}(1)$ and the selection of $m\geq T^3$~\citep{zhou2020neuralUCB}. So the whole bound depends mainly on the first term. Regarding the first term in our regret upper bound, since $\sigma_t \leq R$ by the condition in  Eq.~\ref{eq:noise_gen}, it can be seen that the first term of the regret of our oracle \acrshort{ours}, i.e., $\tilde{\mathcal{O}}\left(R\sqrt{dT}+d\sqrt{\sum_{t=1}^T\sigma_t^2}\right)$ is strictly better than $\tilde{\mathcal{O}}\left(Rd\sqrt{T}\right)$ of \acrshort{NeuralLinUCB}.
\end{remark}

Finally, we conclude the regret bound for our practical \acrshort{ours} algorithm:
\begin{theorem}\label{theo:regret_upper_bound_alg}
    Under the conditions in Thm.~\ref{theo:regret_upper_bound}, then, with probability at least $1-\delta$, the regret of the practical algorithm in Alg.~\ref{alg:ours} is bounded by
    \small
    \begin{align*}
        \text{Regret}(T) &\leq \tilde{\mathcal{O}} \left(R\sqrt{dT} + d\sqrt{\sum_{t=1}^T \sigma_t^2 + \epsilon}\right)\\
        &\quad + \tilde{\mathcal{O}}\left(m^{-1/6}T\sqrt{(\mathbf{r}-\mathbf{\tilde{r}})^\top \mathbf{H}^{-1} (\mathbf{r}-\mathbf{\tilde{r}})}\right),
    \end{align*}
    \normalsize
    where the estimation error is bounded by $\epsilon = d^3\left(T^2n^9L^{11}\log^6(m)\right)^{-1}$. The proof is in Apd.~\ref{proof:thm:regret_upper_bound_alg}.
\end{theorem}

\begin{remark}\label{rem:est_error}(Regret comparison between oracle and practical version).
    If $\epsilon$ is small enough, then the regret bound in Thm.~\ref{theo:regret_upper_bound_alg} becomes close to the oracle \acrshort{ours} in Thm.~\ref{theo:regret_upper_bound}, i.e., $\tilde{\mathcal{O}}\left(R\sqrt{dT}+d\sqrt{\sum_{t=1}^T\sigma_t^2}\right)$. This is practically possible when the time horizon $T$ increases and we can design a neural network with deep layers $L$, hidden width size $m$, and enough training iteration $n$.
\end{remark}
\section{Experiments}\label{sec:exp}
We empirically compare our practical \acrshort{ours} algorithm with five main baselines in the main paper, including \acrshort{LinUCB}~\citep{yasin2011improved}, \acrshort{NeuralUCB}~\citep{zhou2020neuralUCB}, \acrshort{NeuralTS}~\citep{zhang2021neural}, Neural-LinGreedy~\citep{xu2022neural}, and \acrshort{NeuralLinUCB}~\citep{xu2022neural}. More baseline comparisons and experimental details are in Apd.~\ref{apd:exo_details}. 
\begin{figure*}[t!]
    \centering
     \setlength{\tabcolsep}{0pt}
    \begin{tabular}{cccc}
    \includegraphics[width=0.25\linewidth]{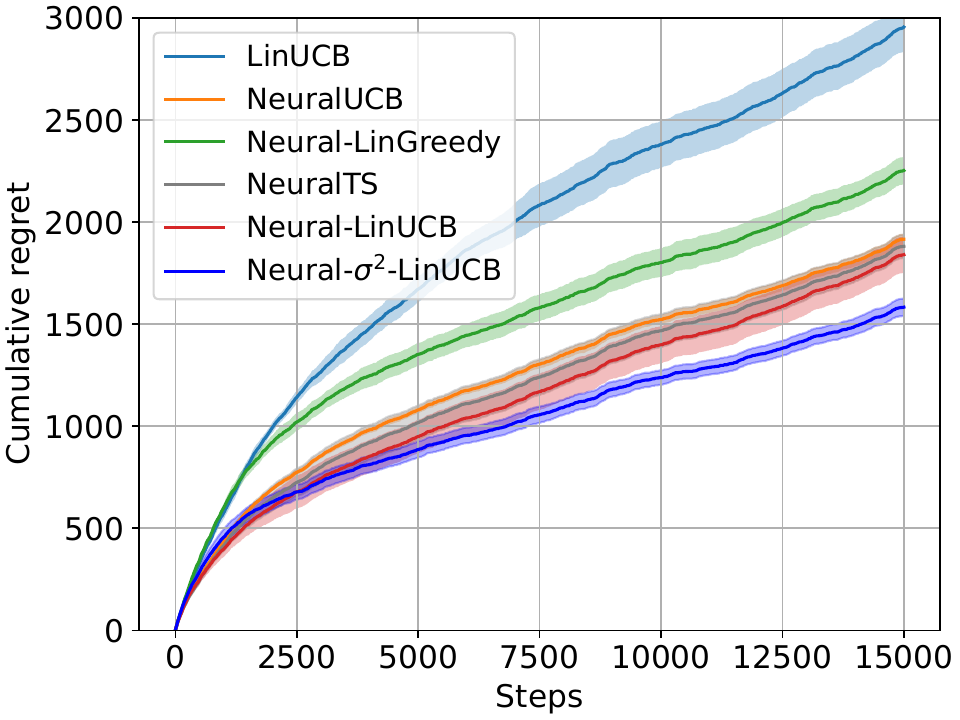}&
    \includegraphics[width=0.25\linewidth]{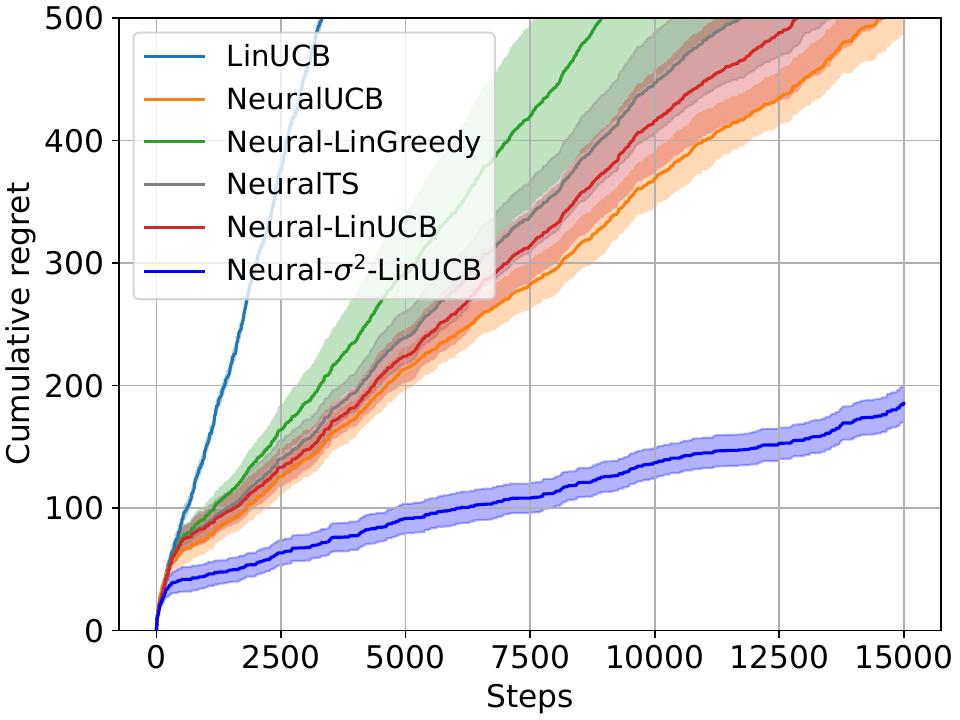} &
    \includegraphics[width=0.25\linewidth]{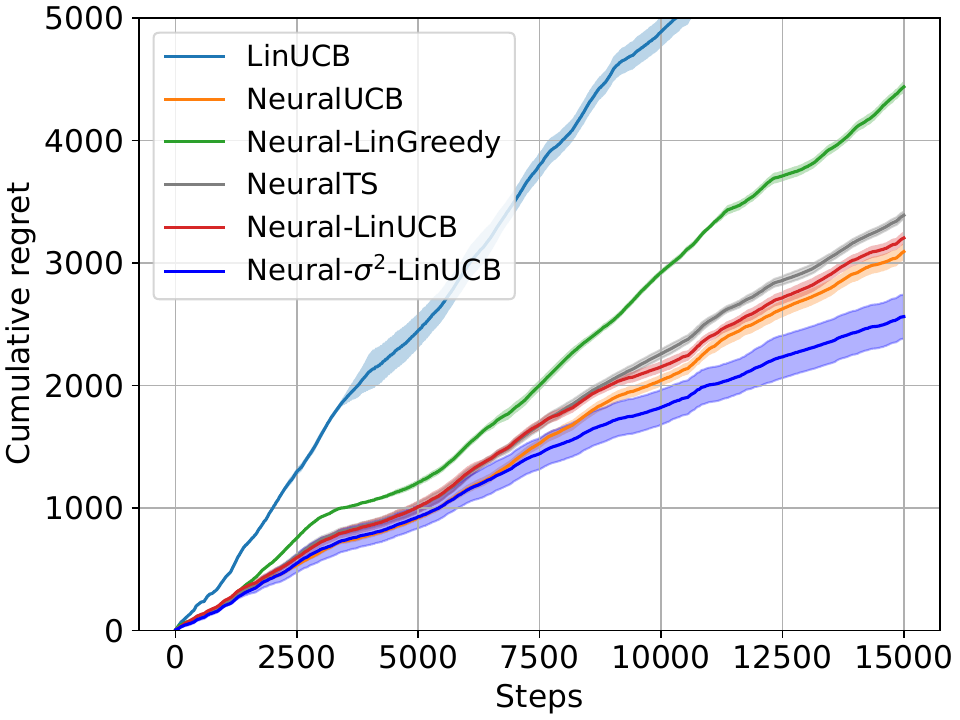}&
    \includegraphics[width=0.25\linewidth]{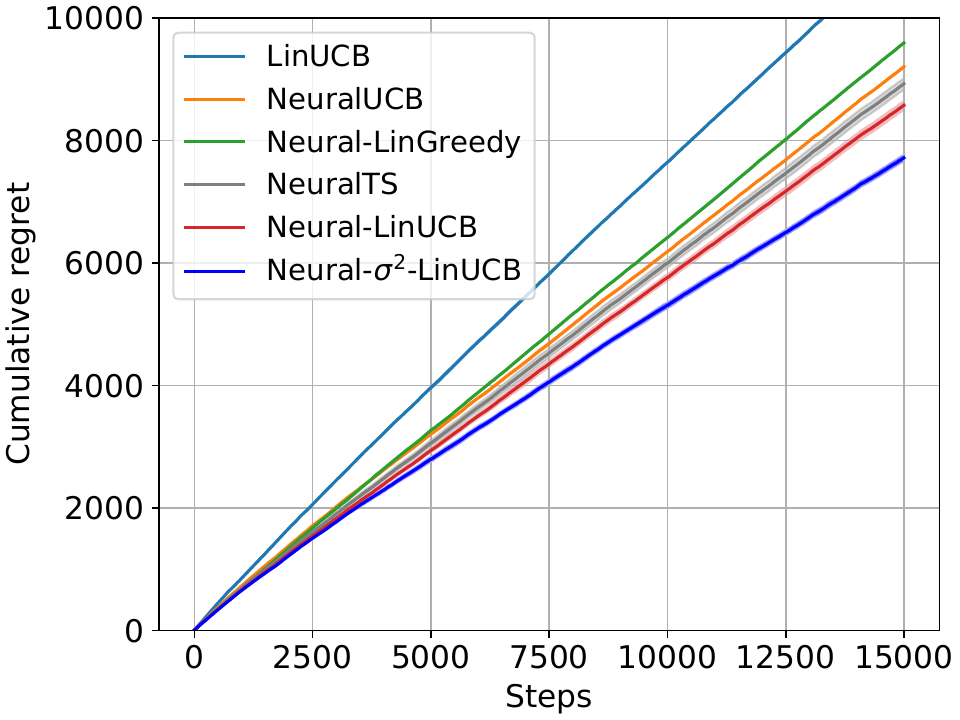}\\
         \scriptsize (a) MNIST & \scriptsize (b) UCI-shuttle & \scriptsize (c) UCI-covertype & \scriptsize (d) CIFAR-10
    \end{tabular}
    \caption{Cumulative regret results on the real-world data across 10 runs with different seeds.}
    \label{fig:real_world}
\end{figure*}

\subsection{Synthetic datasets}
We follow~\citet{zhou2020neuralUCB} by setting the context dimension $d=20$, arms number $K=4$, and time horizon $T=10000$. We sample the context uniformly at random from the unit ball, i.e., \small$\mathbf{x}_{t,a}=\left(\mathbf{x}_{t,a}^{(1)}, \mathbf{x}_{t,a}^{(2)}, \cdots, \mathbf{x}_{t,a}^{(d)}\right)$, $\mathbf{x}_{t,a}^{(i)}\sim \mathcal{N}(0,1)/\left\|\mathbf{x}_{t,a}\right\|_2$, $\forall i \in [d]$\normalsize. Then, we define the reward function $h$ with 3 settings: \small$h_1(\mathbf{x}_{t,a}) = 10(\mathbf{x}_{t,a}^\top \bm{\theta})^2$, $h_2(\mathbf{x}_{t,a}) = \mathbf{x}_{t,a}^\top \bm{\theta}^\top \bm{\theta} \mathbf{x}_{t,a}$, and $h_3(\mathbf{x}_{t,a}) = \cos(3\mathbf{x}_{t,a}^\top \bm{\theta})$\normalsize, where $\bm{\theta}$ is also randomly generated uniformly over the unit ball. For each $h_i(\cdot)$, the reward is generated by $r_{t,a}^{(i)}=h_i(\mathbf{x}_{t,a}) + \xi_t$. We consider a randomly changing variance by setting at each time $t$, \small$\xi_t \sim \mathcal{N}(0, \mathbb{E}[\xi^2_t\mid \mathbf{x}_{1:t, a_{1:t}}, \xi_{1:t-1}])$\normalsize, where \small$\mathbb{E}[\xi^2_t\mid \mathbf{x}_{1:t, a_{1:t}}, \xi_{1:t-1}] \sim \mathbb{U}(0,1)$\normalsize. 

For each algorithm, we run 5 traces with different random seeds per run, and then we summarize their cumulative regret results in Fig.~\ref{fig:demo}. Firstly, we observe that \acrshort{ours} is consistently better than \acrshort{SOTA} baselines by having a significantly low cumulative regret as $t\in [T]$ grows. For instance, in the first setting with $h_1(\mathbf{x}_{t,a})$, at the final round $t=T$, our regret is below $1000$, which is better than \acrshort{NeuralLinUCB} by about $600$, and remarkably better than \acrshort{NeuralUCB} by about $1200$. Secondly, neural-bandit algorithms significantly outperform the non-neural algorithm \acrshort{LinUCB} in all settings (more details are in Fig.~\ref{fig:full_demo}). This continues to confirm the hypothesis that non-linear models can address the limitation of the linear-reward assumption~\citep{riquelme2018deep,zhou2020neuralUCB}.

\subsection{Real-world datasets}
To validate our model's effectiveness in the real world, we deploy on the MNIST~\citep{lecun2010MNIST}, UCI-shuttle~(statlog), UCI-covertype~\citep{dua2017UCI}, and CIFAR-10 dataset~\citep{cifar10}. Following~\citet{beygelzimer2009offset}, we convert these dataset to $K$-armed contextual bandits by transforming each labeled data $(\mathbf{x}\in \mathbb{R}^d, \mathbf{y}\in \mathbb{R}^K)$ into $K$ context vector \small$\mathbf{x}^{(1)}=(\mathbf{x},\mathbf{0}, \cdots, \mathbf{0}), \cdots, \mathbf{x}^{(K)}=(\mathbf{0},\mathbf{0}, \cdots, \mathbf{x})\in \mathbb{R}^{dK}$\normalsize. We define the reward function by $1$ if the agent selects the exact arm $i\in [K]$ s.t. $\mathbf{y}_i=1$, and $0$ otherwise.

We compare methods over $T=15000$ rounds across $5$ runs in Fig.~\ref{fig:real_world}. In low-dimensional data like UCI-shuttle, our model behaves similarly to the synthetic data with a significantly lower regret. In high-dimensional data like MNIST, UCI-covertype, and CIFAR-10, although all models find it hard to estimate the underlying reward function, \acrshort{ours} still consistently outperforms other methods. Furthermore, our results are stable across different running seeds with small variance intervals on 10 runs. In Fig.~\ref{fig:model_size} in Apd.~\ref{apd:exp_regret}, we also show a case when the model capacity (i.e., $L$ and $m$) increases, we can further achieve a lower cumulative regret on CIFAR-10. To this end, we can see that our method not only has a lower regret than others in the synthetic data but also in the real-world dataset, confirming the tighter regret bound of Thm.~\ref{theo:regret_upper_bound} and~\ref{theo:regret_upper_bound_alg}.

\subsection{Uncertainty estimation evaluations} 
\begin{figure}[ht!]
    \hspace*{-0.2in}
    \centering
     \setlength{\tabcolsep}{0pt}
    \begin{tabular}{cc}
    \includegraphics[width=0.55\linewidth]{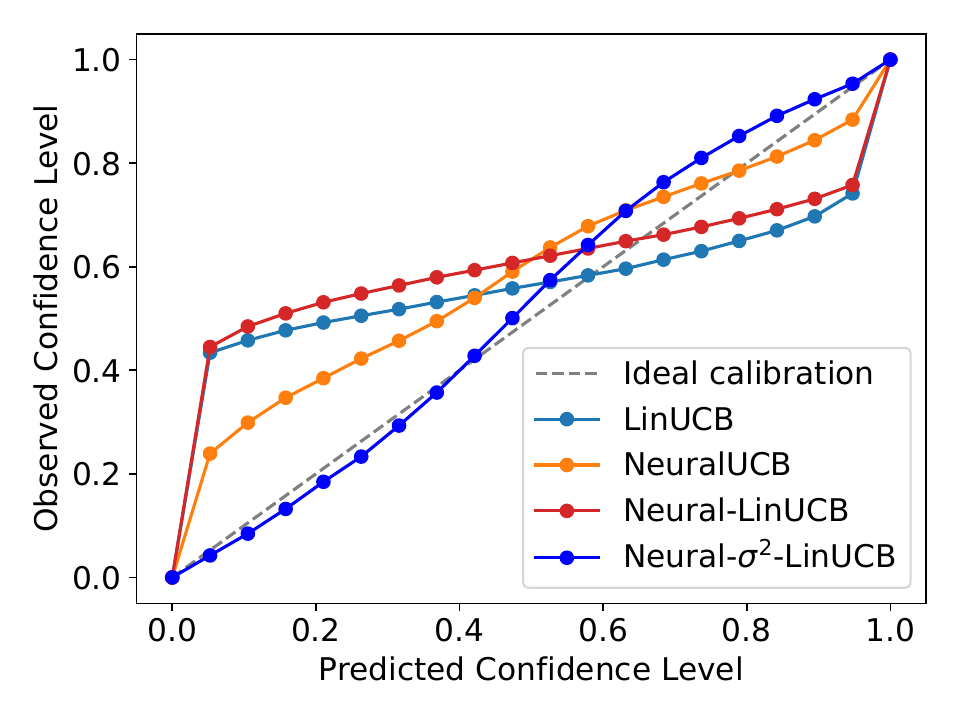}&
    \includegraphics[width=0.55\linewidth]{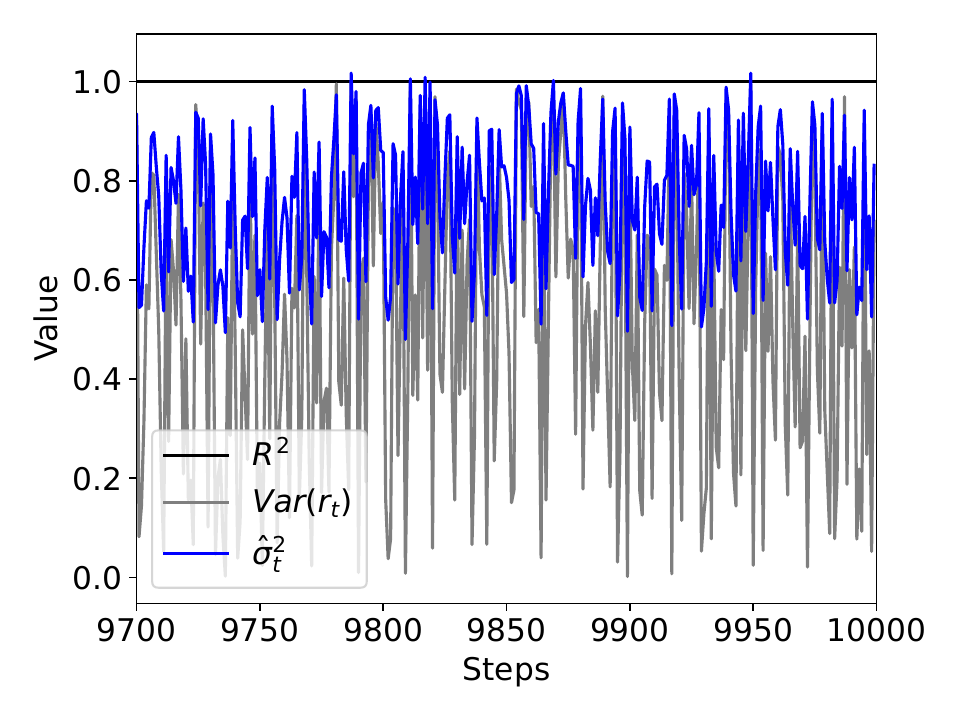}\\
        \scriptsize (a) & \scriptsize (b)
    \end{tabular}
    \caption{(a) Visualization of calibration error in Eq.~\ref{eq:calib} with reliability diagram on $h_1(\mathbf{x}_{t,a})$ dataset. (b) Reward variance $Var(r_t)$, our estimation for the variance upper bound $\sigma_t^2$, and the upper bound $R^2$ comparison.}
    \label{fig:calib}
\end{figure}

To better understand the uncertainty quality of \acrshort{UCB}, Fig.~\ref{fig:calib}~(a) compare the calibration performance across \acrshort{UCB} methods. Intuitively, calibration means a $p$ confidence interval contains the reward $p$ of the time~\citep{gneiting2007probabilistic} (evaluation metrics details are in Apd.~\ref{apd:exp_uncertainty}). We can see that by leveraging a high-quality estimation for $\sigma_t^2$ in Eq.~\ref{eq:est_sigma}, our \acrshort{UCB} is more well-calibrated by less over-confidence and under-confidence than other methods. More quantitative results are in Tab.~\ref{tab:quatitative} in Apd.~\ref{apd:exp_uncertainty}. We also evaluate calibration by different checkpoints across time steps on a hold-old validation set in Fig.~\ref{fig:calib_details_0},~\ref{fig:calib_details_1},~\ref{fig:ablation_study2},~\ref{fig:calib_details_3} in Apd.~\ref{apd:exp_uncertainty}. Overall, we also observe that our method is more calibrated than other algorithms. These results are consistent with the observation that a calibrated model can further improve the cumulative regret~\citep{calibrated2019malik,deshpande2024online}.

Regarding the estimation quality for $\sigma_t^2$ of Eq.~\ref{eq:est_sigma}, we visualize our estimated $\hat{\sigma}_t^2$, the true $Var(r_t)$, and the magnitude $R^2$ at the last $300$ steps in Fig.~\ref{fig:calib}~(b) (details are in Fig.~\ref{fig:ablation_study2} in Apd.~\ref{apd:exp_uncertainty}). We can see that since we set \small$\mathbb{E}[\xi^2_t\mid \mathbf{x}_{1:t, a_{1:t}}, \xi_{1:t-1}] \sim \mathbb{U}(0,1)$\normalsize, so $R^2=1$ and in almost all steps, our estimated $\hat{\sigma}^2_t$ has a higher value than $Var(r_t)$ and lower value than $R^2$, showing a high-quality estimation in our Eq.~\ref{eq:est_sigma}.

\subsection{Computational efficiency evaluations}
\begin{table}[ht!]
    \centering
    \scalebox{0.8}{
    \begin{tabular}{ccc}
        \toprule  
         \textbf{Methods} & \textbf{Arm selection ($\downarrow$)} & \textbf{\acrshort{DNN} update ($\downarrow$)}\\
         \midrule
         NeuralUCB & 6.42 $\pm$ 0.33 & 2.28 $\pm$ 0.21\\
         NeuralTS & 7.27 $\pm$ 0.40 & 2.62 $\pm$ 0.29\\
         Neural-LinUCB & \textbf{0.43 $\pm$ 0.02} & \textbf{1.86 $\pm$ 0.17}\\
         \acrshort{ours} & \textbf{0.43 $\pm$ 0.02} & \textbf{1.86 $\pm$ 0.17}\\
         \bottomrule
    \end{tabular}}
    \caption{Computational cost comparison on MNIST for running $100$ rounds: runtime (seconds) on RTX-A5000.}
    \label{tab:latency}
\end{table}
We compare the latency of neural-bandit algorithms in Tab.~\ref{tab:latency} with $H=10$ and $t\in [100]$. Overall, similar to \acrshort{NeuralLinUCB}, we can see that our method is more efficient than \acrshort{NeuralUCB} and \acrshort{NeuralTS} in both the arm selection (Lines L-5-8) and \acrshort{DNN} update step (L-17-21) in Alg.~\ref{alg:ours}. Especially, since they require to perform \acrshort{UCB}/sampling on entire \acrshort{DNN} parameters, our method is much faster than by around $6$~seconds in the arm selection step by the \acrshort{UCB} is performed over the linear mode with the feature from the last \acrshort{DNN} layer. A detailed comparison is in Fig.~\ref{fig:latency} in Apd.~\ref{apd:exp_latency}. Given better regret performances, \acrshort{ours} makes a significant contribution by achieving a balance of computational efficiency, high uncertainty quality, and accurate reward estimation in real-world domains.

\subsection{Ablation study for \acrshort{ours}}
To take a closer look at our Alg.~\ref{alg:ours}, we compare 4 settings, including: (1) using \acrshort{MSE} in Eq.~\ref{eq:MSE} with the true variance $Var(r_t)$ from the generating process of the synthetic data (Oracle\_Neural\_MSE); (2) using \acrshort{MLE} in Eq.~\ref{eq:MLE} with the estimated $Var(r_t)$ from $\bm{\phi}(\mathbf{x}_{t,a_t}; \mathbf{w})^\top \mathbf{A}_t^{-1} \bm{\phi}(\mathbf{x}_{t,a_t}; \mathbf{w})$ (Neural\_MLE\_Var); (3) using the estimated $\sigma_t^2$ in Eq.~\ref{eq:est_sigma} with \acrshort{MSE} in Eq.~\ref{eq:MSE} (Neural\_MSE, i.e., \acrshort{ours}); (4) using the estimated $\sigma_t^2$ in Eq.~\ref{eq:est_sigma} with \acrshort{MLE} in Eq.~\ref{eq:MLE} (Neural\_MLE). More results are in Apd.~\ref{apd:exp_regret}.

\begin{figure}[ht!]
    \centering
     \setlength{\tabcolsep}{0pt}
    \begin{tabular}{cc}
    \includegraphics[width=0.5\linewidth]{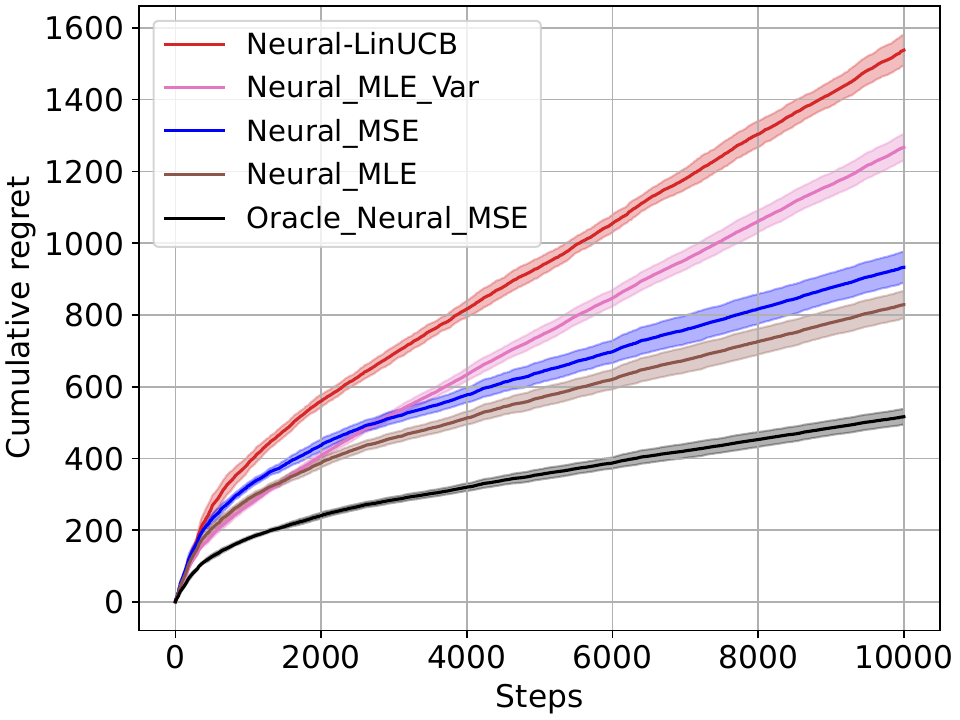}&
    \includegraphics[width=0.5\linewidth]{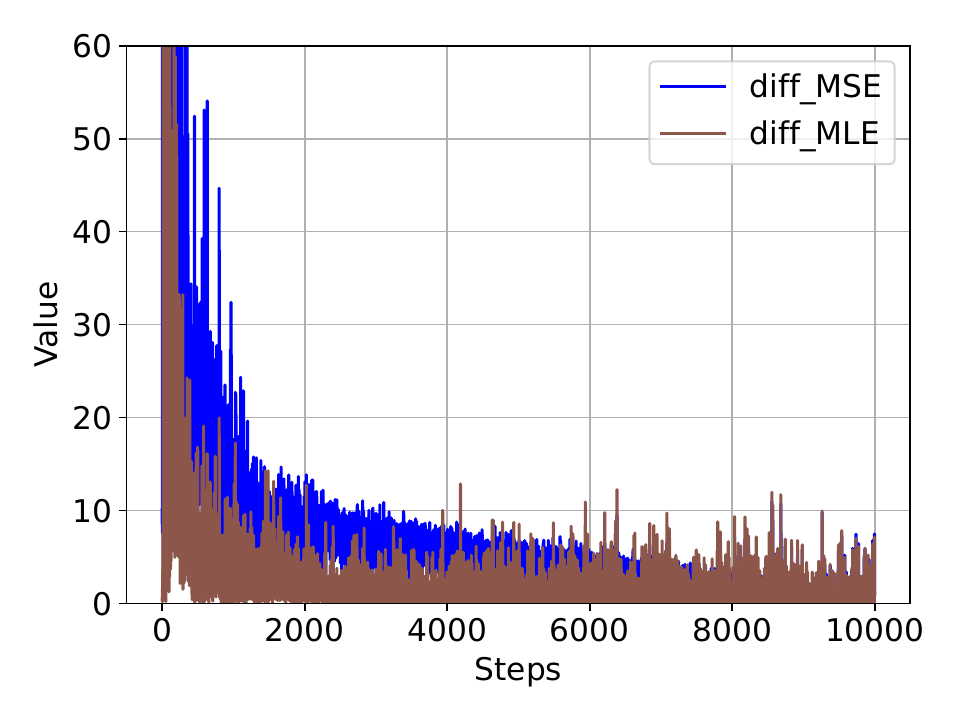}\\
        \scriptsize (a) & \scriptsize (b)
    \end{tabular}
    \caption{(a) Comparison in cumulative regret between different settings. (b) Variance estimation error between \acrshort{MLE} and \acrshort{MSE} settings of our \acrshort{ours}.}
    \label{fig:ablation_study}
\end{figure}

Fig.~\ref{fig:ablation_study}~(a) shows oracle \acrshort{ours}, i.e., Oracle\_Neural\_MSE has the lowest regret on the synthetic data $h_1(\mathbf{x}_{t,a})$. After that is our practical \acrshort{ours} versions, including Neural\_MSE and Neural\_MLE with the estimated $\sigma^2_t$ in Eq.~\ref{eq:est_sigma}. Notably, all of them are significantly better than \acrshort{NeuralLinUCB}. 

It is also worth noticing that using \acrshort{MLE} in Eq.~\ref{eq:MLE} brings out a slightly better performance than \acrshort{MSE} in Eq.~\ref{eq:MSE} by a lower regret of Neural\_MLE than Neural\_MSE in Fig.~\ref{fig:ablation_study}~(a), and a lower variance estimation error in Fig.~\ref{fig:ablation_study}~(b), where the y-axis is the difference between the estimation and the true variance at time $t$, i.e., \small$\left|\bm{\phi}(\mathbf{x}_{i,a_i}; \mathbf{w})^\top \mathbf{A}_t^{-1} \bm{\phi}(\mathbf{x}_{i,a_i}; \mathbf{w})-Var(r_t)\right|$\normalsize. This confirms the effectiveness of using \acrshort{MLE} to improve uncertainty estimation quality of \acrshort{DNN}~\citep{chua2018deepRL,bui2024density}. That said, estimating the true variance $Var(r_t)$ with \acrshort{MLE} is still difficult by having a high estimation error. As a result, the regret performance of Neural\_MLE\_Var is still worse than estimating the upper bound $\sigma_t^2$ (i.e., Neural\_MLE).

\begin{figure}[ht!]
    \centering
     \setlength{\tabcolsep}{0pt}
    \begin{tabular}{cc}
    \includegraphics[width=0.5\linewidth]{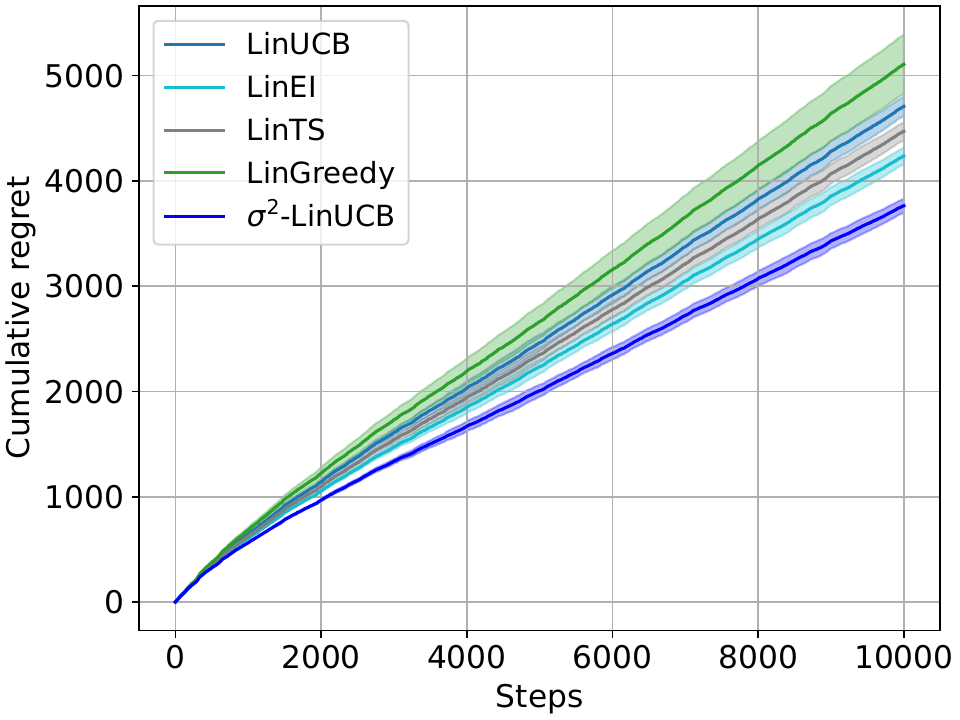}&
    \includegraphics[width=0.5\linewidth]{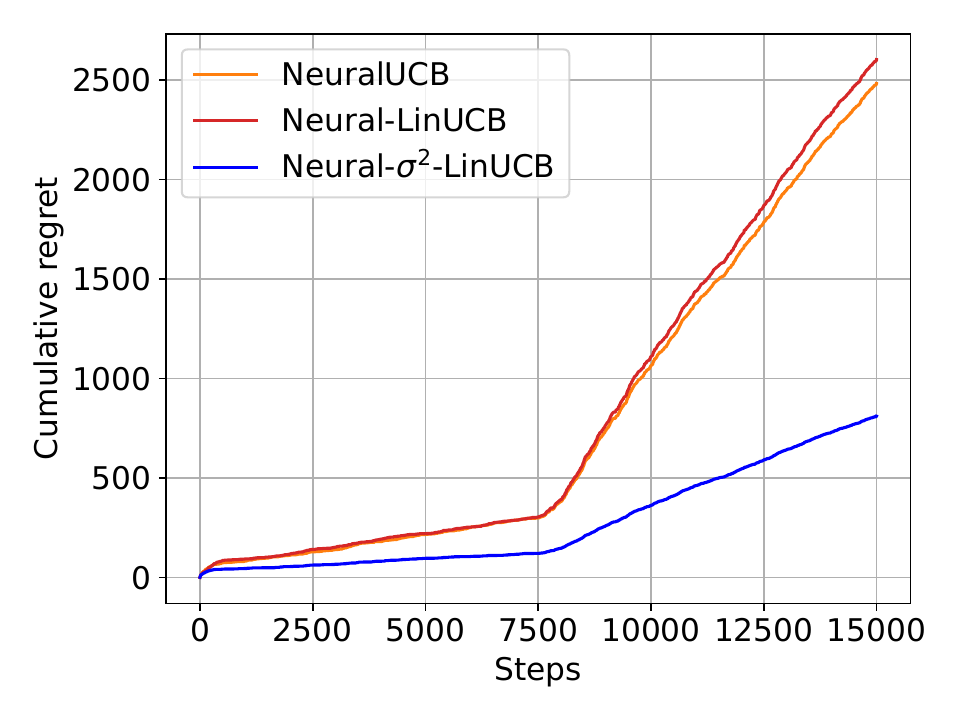}\\
        \scriptsize (a) & \scriptsize (b)
    \end{tabular}
    \caption{(a) Cumulative regret results on $h_1(\mathbf{x}) = 10(\mathbf{x}^\top \bm{\theta})^2$ without \acrshort{DNN} setting. (b) Cumulative regret results on UCI-shuttle with dynamic reward range.}
    \label{fig:ab_dynamic}
\end{figure}

We provide an additional result in the linear contextual bandits setting (i.e., without the \acrshort{DNN} version). Fig.~\ref{fig:ab_dynamic}~(a) shows that our results are consistent between linear and neural contextual bandits settings by having a lower regret than other linear baselines.

Finally, we test our model on UCI-shuttle with a dynamic reward range. Specifically, when $t\in [0,T/2]$, we define the reward function by $1$ if the agent selects the exact arm $i\in [K]$ s.t. $\mathbf{y}_i=1$, and $0$ otherwise. And, when $t\in [T/2,T]$, we define the reward function by $3$ if the agent selects the exact arm $i\in [K]$ s.t. $\mathbf{y}_i=1$, and $1$ otherwise. Therefore, the reward range will be $[a,b] = [0,1]$ if $t\in [0,T/2]$, and $[a,b] = [1,3]$ if $t\in [T/2,T]$. Fig.~\ref{fig:ab_dynamic}~(b) shows that our method achieves a lower cumulative regret than \acrshort{NeuralUCB} and \acrshort{NeuralLinUCB} in this dynamic reward range setting, confirming our empirical results.
\section{Related work}
We can categorize methods in the non-linear contextual bandits into three main approaches. First is \textbf{using non-parametric modeling}, including perception~\citep{sham2008efficient}, random forest~\citep{feraud2016random}, Gaussian processes~\citep{srinivas2010GP,krause2011Contextual}, and kernel space~\citep{valko2013finite,bubeck2011xbandit}. The second approach is \textbf{reducing to supervised-learning problems}, which optimizes objective function based on fully-labeled data with context-reward pair~\citep{langford2007greedy,foster2020Beyond,agarwal2014Taming}. 

Our method is relevant to the last approach, which is \textbf{considering generalized linear bandits} by decomposing the reward function to a linear and a non-linear link function~\citep{filippi2010parametric, li2017Provably, jun2017Scalable}, e.g., the mixture of linear experts~\citep{beygelzimer2009offset}, or using the non-linear link function by \acrshort{DNN}~\citep{riquelme2018deep,kveton2020Randomized,zhou2020neuralUCB}. That said, our method is sampling-free and more computationally efficient than the sampling-based methods, e.g., the mixture of experts and \acrfull{NeuralTS}-based model~\citep{zhang2021neural,xu22LMC}. Compared to other sampling-free approaches, e.g., \acrshort{NeuralUCB}~\citep{zhou2020neuralUCB,ban2022eenet}, our algorithm has a lower regret and also is more efficient by \acrshort{NeuralUCB} has a $\Tilde{\mathcal{O}}(R\Tilde{d}\sqrt{T})$ regret bound, where $\Tilde{d}$ is the dimensions of \acrshort{NTK} matrix which can potentially scale with $\mathcal{O}(TK)$. Compared with the most relevant method, i.e., \acrshort{NeuralLinUCB}~\citep{xu2022neural}, our method enjoys a similar computational complexity, while having a better regret bound.

\textbf{Variance-aware-\acrshort{UCB} algorithms.} In the standard bandits setting, leveraging reward uncertainty of the agent model to enhance \acrshort{UCB} has shown promising results. In particular, previous work~\citep{kuleshov2014algorithms,Audibert2009Exploration} have shown UCB1-Tuned and UCB1-Normal, models using the estimated variance can obtain a lower regret than the standard \acrshort{UCB} method~\citep{auer2002finitetime}. Regarding the contextual bandits setting, several theoretical studies~\citep{zhao2023variance,ye2023Corruption,Zhou2024Computationally} have shown that variance-dependent regret bounds for linear contextual bandits are lower than the sublinear regret of \acrshort{LinUCB}. However, all of them only consider the linear contextual bandits setting while \acrshort{ours} considers the non-linear case. Furthermore, these algorithms often require a known variance of the noise and its upper bound~\citep{zhou2021nearly,Zhou2024Computationally}. Additionally, they are inefficient in terms of computation or even computationally intractable in practice~\citep{zhang2021improved,kim2022improved,zhao2023variance}. As a result, none of these works provide empirical evidence with experimental results.

In particular, we can compare with \acrfull{SAVE}~\citep{zhao2023variance}, which can be seen as a variance-aware version of SupLinUCB~\citep{chu2011contextual}. Beyond the differences between linear and neural contextual bandits, our setting is also different because our \acrshort{ours} and our other baselines in Sec.~\ref{sec:exp} are not given knowledge of the time horizon $T$, i.e., the knowledge of the optimal regret. Meanwhile, \acrshort{SAVE} is a theoretical approach that is not practical since their algorithm requires prior knowledge of $T$ (i.e., $\alpha$ and $K$ in~\citet{zhao2023variance}) to balance exploration and exploitation.

In summary, compared to previous works, we provide both theoretical evidence for variance-aware neural bandits and empirical evidence with the practical algorithm in real-world contextual settings. We believe we are one of the first works that analyze regret bound for variance-aware-\acrshort{UCB} with \acrshort{DNN}. Importantly, we also introduce a practical algorithm with extensive empirical results for variance-aware-\acrshort{UCB}, opening an empirical direction for this approach.
\section{Conclusion}
Neural-\acrshort{UCB} bandits have shown success in practice and theoretically achieve $\tilde{\mathcal{O}}(Rd\sqrt{T})$ regret bound. To enhance the \acrshort{UCB} quality and regret guarantee, we introduce \acrshort{ours}, a variance-aware algorithm that leverages $\sigma_t^2$, i.e., an upper bound of the reward noise variance at round $t$. We propose an oracle algorithm with an oracle $\sigma_t^2$ and a practical version with a novel estimation for $\sigma_t^2$. We analyze regret bounds for both oracle and practical versions. Notably, our oracle algorithm achieves a tighter bound with \small$\tilde{\mathcal{O}}\left(R\sqrt{dT}+d\sqrt{\sum_{t=1}^T\sigma_t^2}\right)$\normalsize regret. Given a reward range, our practical algorithm estimates $\sigma_t^2$ using the estimated reward mean with \acrshort{DNN}. Experimentally, our practical method enjoys a similar computational efficiency while outperforming \acrshort{SOTA} techniques by having a lower calibration error and lower cumulative regret across different settings on benchmark datasets. With these promising results, we hope that our work will open a door for the direction of understanding uncertainty estimation to enhance neural-bandit algorithms in both theoretical and practical aspects.

\section*{Acknowledgments}
We would like to thank the anonymous reviewers for their helpful comments. H.M.Bui~is supported by the Discovery Award of Johns Hopkins University and the Challenge grant from the JHU Institute of Assured Autonomy. A.Liu~is partially supported by the Amazon Research Award, the Discovery Award of the Johns Hopkins University, and a seed grant from the JHU Institute of Assured Autonomy.

\bibliographystyle{unsrtnat}
\bibliography{refs}
\printglossary

\section*{Checklist}
 \begin{enumerate}

 \item For all models and algorithms presented, check if you include:
 \begin{enumerate}
   \item A clear description of the mathematical setting, assumptions, algorithm, and/or model. [Yes]
   \item An analysis of the properties and complexity (time, space, sample size) of any algorithm. [Yes]
   \item (Optional) Anonymized source code, with specification of all dependencies, including external libraries. [Yes]
 \end{enumerate}

 \item For any theoretical claim, check if you include:
 \begin{enumerate}
   \item Statements of the full set of assumptions of all theoretical results. [Yes]
   \item Complete proofs of all theoretical results. [Yes]
   \item Clear explanations of any assumptions. [Yes]     
 \end{enumerate}

 \item For all figures and tables that present empirical results, check if you include:
 \begin{enumerate}
   \item The code, data, and instructions needed to reproduce the main experimental results (either in the supplemental material or as a URL). [Yes]
   \item All the training details (e.g., data splits, hyperparameters, how they were chosen). [Yes]
         \item A clear definition of the specific measure or statistics and error bars (e.g., with respect to the random seed after running experiments multiple times). [Yes]
         \item A description of the computing infrastructure used. (e.g., type of GPUs, internal cluster, or cloud provider). [Yes]
 \end{enumerate}

 \item If you are using existing assets (e.g., code, data, models) or curating/releasing new assets, check if you include:
 \begin{enumerate}
   \item Citations of the creator If your work uses existing assets. [Yes]
   \item The license information of the assets, if applicable. [Not Applicable]
   \item New assets either in the supplemental material or as a URL, if applicable. [Not Applicable]
   \item Information about consent from data providers/curators. [Not Applicable]
   \item Discussion of sensible content if applicable, e.g., personally identifiable information or offensive content. [Not Applicable]
 \end{enumerate}

 \item If you used crowdsourcing or conducted research with human subjects, check if you include:
 \begin{enumerate}
   \item The full text of instructions given to participants and screenshots. [Not Applicable]
   \item Descriptions of potential participant risks, with links to Institutional Review Board (IRB) approvals if applicable. [Not Applicable]
   \item The estimated hourly wage paid to participants and the total amount spent on participant compensation. [Not Applicable]
 \end{enumerate}

 \end{enumerate}

\appendix
\onecolumn
\noindent\rule{\textwidth}{3pt}
\section*{\Large\centering{Variance-Aware Linear UCB with Deep Representation for Neural Contextual Bandits\\(Supplementary Material)}}
\noindent\rule{\textwidth}{1pt}
\textbf{Broader impacts}. Contextual bandits involve several artificial intelligence applications, e.g., personal healthcare, finance, recommendation systems, etc. There has been growing interest in using \acrshort{DNN} to improve bandits algorithms. Our \acrshort{ours} improves the quality of such models by having a lower calibration error and a lower regret. This could particularly benefit the aforementioned high-stake applications.

\textbf{Limitations}: 
\begin{enumerate}[leftmargin=13pt,topsep=0pt,itemsep=0mm]
    \item \textbf{The gap between oracle and practical algorithm.} Although showing a better regret guarantee in theory and experiments than other related methods, there is still a gap between our oracle and practical algorithm as we need to estimate the upper bound variance with a known reward range $[a,b]$ and the magnitude $R$.
    \item \textbf{The gap between theory and experiments}. Similar to the literature on neural bandits, although $\alpha_t$ has a specific form in Theorem~\ref{theo:regret_upper_bound}, in experiments, we set it to be constant for a fair comparison with other baselines. Specifically, the exploration rate $\alpha_t$ in~\citet{xu2022neural} ($\gamma_t$ in~\citet{zhou2020neuralUCB}) is also set to be a constant instead of the true value in the theorems. We also compare with the true value $\alpha_t$ in the theory in Figure~\ref{fig:true_alpha}.
\end{enumerate}
\textbf{Remediation.} Given the aforementioned limitations, we encourage people who extend our work to proactively confront the model design and parameters to desired behaviors in real-world use cases.

\textbf{Future work.} We plan to tackle \acrshort{ours} limitation, reduce assumptions in theory, and add more estimation techniques for $\sigma_t^2$ to enhance the quality of the practical algorithm.

\textbf{Reproducibility.} The source code to reproduce our results is available at \href{https://github.com/Angie-Lab-JHU/neuralVarLinUCB}{https://github.com/Angie-Lab-JHU/neuralVarLinUCB}. We provide all proofs in Appendix~\ref{apd:proof}, experimental settings, and detailed results in Appendix~\ref{apd:exo_details}.
\section{Proofs}\label{apd:proof}
\subsection{Proof of Theorem~\ref{thm:reward_distribution}}\label{proof:thm:reward_distribution}
\begin{proof}
    By the definition of the noise random variable, i.e., $\mathbb{E}[\xi_t\mid \mathbf{x}_{1:t, a_{1:t}}, \xi_{1:t-1}] = 0$ in Equation~\ref{eq:noise_gen}, we have 
    \begin{align}
        Var(\xi_t\mid \mathbf{x}_{1:t, a_{1:t}}, \xi_{1:t-1}) &= \mathbb{E}[\xi^2_t\mid \mathbf{x}_{1:t, a_{1:t}}, \xi_{1:t-1}] - \left\{\mathbb{E}[\xi_t\mid \mathbf{x}_{1:t, a_{1:t}}, \xi_{1:t-1}]\right\}^2\\
        &= \mathbb{E}[\xi^2_t\mid \mathbf{x}_{1:t, a_{1:t}}, \xi_{1:t-1}].
    \end{align}
    Since $\mathbb{E}[\xi_t\mid \mathbf{x}_{1:t, a_{1:t}}, \xi_{1:t-1}] = 0$ by Definition in Equation~\ref{eq:noise_gen}, applying the Law of total variance, we get
    \begin{align}
        Var(\xi_t) &= \mathbb{E}\left[Var(\xi_t\mid \mathbf{x}_{1:t, a_{1:t}}, \xi_{1:t-1})\right] + Var(\mathbb{E}\left[\xi_t\mid \mathbf{x}_{1:t, a_{1:t}}, \xi_{1:t-1}\right])\\
        &= \mathbb{E}\left[\mathbb{E}[\xi^2_t\mid \mathbf{x}_{1:t, a_{1:t}}, \xi_{1:t-1}]\right] + Var(0) = \mathbb{E}[\xi^2_t\mid \mathbf{x}_{1:t, a_{1:t}}, \xi_{1:t-1}].
    \end{align}
    On the other hand, by the Law of Expectation, we have
    \begin{align}
        \mathbb{E}[\xi_t] = \mathbb{E}\left[\mathbb{E}[\xi_t\mid \mathbf{x}_{1:t, a_{1:t}}, \xi_{1:t-1}]\right] = \mathbb{E}[0] = 0.
    \end{align}
    Using definition in Equation~\ref{eq:reward_gen}, i.e., $r_{t,a_t} = h(\mathbf{x}_{t,a_t}) + \xi_t$, since $\mathbb{E}[h(\mathbf{x}_{t,a_t})] = h(\mathbf{x}_{t,a_t})$ and $Var(h(\mathbf{x}_{t,a_t})) = 0$, by the linearity of expectation, we obtain
    \begin{align}
        \mathbb{E}[r_{t,a_t}] = \mathbb{E}[h(\mathbf{x}_{t,a_t})] + \mathbb{E}[\xi_t] = h(\mathbf{x}_{t,a_t}),
    \end{align}
    and by the sum of the variance of independent variables, yielding
    \begin{align}
        Var(r_{t,a_t}) &= Var(h(\mathbf{x}_{t,a_t})) + Var(\xi_t) + 2\cdot Cov(h(\mathbf{x}_{t,a_t}), \xi_t)\\
        &= \mathbb{E}[\xi^2_t\mid \mathbf{x}_{1:t, a_{1:t}}, \xi_{1:t-1}]
    \end{align}
    of Theorem~\ref{thm:reward_distribution}. 
\end{proof}

\subsection{Proof of Theorem~\ref{thm:variance_bound}}\label{proof:thm:variance_bound}
\begin{proof}
    Let us first consider a random variable $X$ is restricted in the interval $[0,1]$. Note that for all $x\in [0,1]$, we always have $x^2 \leq x$, yielding $\mathbb{E}[X^2] \leq \mathbb{E}[X]$. Therefore, we have
    \begin{align}
        Var(X) = \mathbb{E}[X^2] - \mathbb{E}[X]^2 \leq \mathbb{E}[X] - \mathbb{E}[X]^2 = \mathbb{E}[X] (1-\mathbb{E}[X]).
    \end{align}
    To generalize to intervals $[a,b]$ with $b>a$, consider $r_{t,a_t}$ in Theorem~\ref{thm:reward_distribution} restricted to $[a,b]$. Let us define 
    \begin{align}
        X=\frac{r_{t,a_t}-a}{b-a},
    \end{align}
    which is restricted in $[0,1]$. Equivalently, $r_{t,a_t} = (b-a)X+a$, thus, we get
    \begin{align}
        Var(r_{t,a_t}) = \mathbb{E}[\xi^2_t\mid \mathbf{x}_{1:t, a_{1:t}}, \xi_{1:t-1}] = (b-a)^2 \cdot Var(X) \leq (b-a)^2 \cdot \mathbb{E}[X] (1-\mathbb{E}[X]),
    \end{align}
    where the inequality is based on the first result. Hence, by substituting
    \begin{align}
        \mathbb{E}[X] = \frac{h(\mathbf{x}_{t,a_t})-a}{b-a},
    \end{align}
    we obtain the first bound
    \begin{align}
        Var(r_{t,a_t}) = \mathbb{E}[\xi^2_t\mid \mathbf{x}_{1:t, a_{1:t}}, \xi_{1:t-1}] &\leq (b-a)^2 \cdot \frac{h(\mathbf{x}_{t,a_t})-a}{b-a}\left(1-\frac{h(\mathbf{x}_{t,a_t})-a}{b-a}\right)\\ 
        &= (b-a)^2 \cdot \frac{h(\mathbf{x}_{t,a_t})-a}{b-a} \cdot \frac{b-h(\mathbf{x}_{t,a_t})}{b-a}\\
        &= (h(\mathbf{x}_{t,a_t})-a)(b-h(\mathbf{x}_{t,a_t}))
    \end{align}
    of Theorem~\ref{thm:variance_bound}, To show the second bound, let us consider the function
    \begin{align}
        \mathcal{F}(h(\mathbf{x}_{t,a_t})) = (h(\mathbf{x}_{t,a_t})-a)(b-h(\mathbf{x}_{t,a_t})) = -h(\mathbf{x}_{t,a_t})^2 + (a+b)\cdot h(\mathbf{x}_{t,a_t}) -ab.
    \end{align}
    Since $\mathcal{F}(h(\mathbf{x}_{t,a_t}))$ is a quadratic function, we know that $\mathcal{F}(h(\mathbf{x}_{t,a_t}))$ is maximized at $h(\mathbf{x}_{t,a_t}) = \frac{a+b}{2}$, yielding
    \begin{align}
        \mathcal{F}(h(\mathbf{x}_{t,a_t}))  \leq \frac{(b-a)^2}{4}.
    \end{align}
    On the other hand, by the definition in Equation~\ref{eq:noise_gen}, we have $-R \leq \xi_t \leq R$, by the reward definition in Equation~\ref{eq:reward_gen}, we obtain the second bound
    \begin{align}
        \mathcal{F}(h(\mathbf{x}_{t,a_t}))  \leq \frac{(b-a)^2}{4} = \frac{\left[\left(h(\mathbf{x}_{t,a_t})+R\right)-\left(h(\mathbf{x}_{t,a_t})-R\right)\right]^2}{4} = R^2
    \end{align}
    of Theorem~\ref{thm:variance_bound}.
\end{proof}

\subsection{Proof of Theorem~\ref{theo:regret_upper_bound} and  Corollary~\ref{corol:regret_upper_bound}}
This proof is based on the following provable Lemma:
\begin{lemma}\label{lemma:feature_COV_bound}(The elliptical potential lemma~\citep{yasin2011improved}).
    Let $\{\mathbf{x}_t\}_{t=1}^\infty$ be a sequence in $\mathbb{R}^d$ and $\lambda>0$. Suppose $||\mathbf{x}_t||_2 \leq G$ and $\lambda\geq \max\{1,G^2\}$ for some $G>0$. Let $\mathbf{A}_t = \lambda \mathbf{I} + \sum_{s=1}^t \mathbf{x}_t \mathbf{x}_t^\top$. Then, we have
    \begin{align*}
        \det(\mathbf{A}_t) \leq (\lambda + t G^2/d)^d, \text{ and, } \sum_{t=1}^T ||\mathbf{x}_t||_{\mathbf{A}_{t=1}^{-1}}^2 \leq 2 \log \frac{\det(\mathbf{A}_T)}{\det(\lambda \mathbf{I})} \leq 2d \log(1+TG^2/(\lambda d)).
    \end{align*}
\end{lemma} 

\begin{lemma}\label{lemma:appr_reward}(Approximate reward by a linear function around initial point~\citep{xu2022neural}).
    Suppose Assumption~\ref{assm:NTK} holds, for matrix $\nabla_\mathbf{w}\bm{\phi}(\mathbf{x};\mathbf{w})\in \mathbb{R}^{d\times p}$, there exists $\mathbf{w}^* \in \mathbb{R}^p$ s.t. $||\mathbf{w}^*-\mathbf{w}^{(0)}||_2 \leq \left(1/\sqrt{m}\right)\sqrt{(\mathbf{r}-\Tilde{\mathbf{r}})^\top \mathbf{H}^{-1}(\mathbf{r}-\Tilde{\mathbf{r}})}$ and
    \begin{align*}
        r(\mathbf{x}_{t,k}) = \bm{\theta}^{*\top}\bm{\bm{\phi}}(\mathbf{x}_{t,k};\mathbf{w}_{t-1}) + \bm{\theta}_0^\top \nabla_\mathbf{w^{(0)}}\bm{\phi}(\mathbf{x}_{t,k};\mathbf{w}^{(0)})(\mathbf{w}^*-\mathbf{w}^{(0)}),
    \end{align*}
    for all $k\in [K]$ and $t\in [T]$.
\end{lemma}

\begin{lemma}\label{lemma:NN_bounds}(Upper bounds of the neural network's output and its gradient~\citep{xu2022neural}).
     Suppose Assumptions~\ref{assm:NTK} holds, then for any round index $t\in [T]$,  suppose it is in the $q$-th epoch, i.e., $t=(q-1)/H+i$ for some $i\in [H]$. If the step size $\eta_q$ satisfies 
     \begin{align*}
         \eta \leq \frac{C_0}{d^2mnT^{5.5}L^6\log(TK/\delta)},
     \end{align*}
     and the width of the neural network satisfies
     \begin{align*}
         m \geq \max\{L\log(TK/\delta),dL^2\log(m/\delta),\delta^{-6}H^{18}L^{16}\log^3(TK)\},
     \end{align*}
     then, with probability at least $1-\delta$ we have
     \begin{align*}
         &||\mathbf{w}_t - \mathbf{w}^{(0)}||_2 \leq \frac{\delta^{3/2}}{m^{1/2}Tn^{9/2}L^6\log^3(m)},\\
         &||\nabla_{\mathbf{w}^{(0)}} \bm{\phi}(\mathbf{x}_{t,a_k};\mathbf{w}^{(0)})||_{F} \leq C_1\sqrt{dLm},\\
         &||\bm{\phi}(\mathbf{x};\mathbf{w}_t)||_2 \leq \sqrt{d\log(n)\log(TK/\delta)},
     \end{align*}
     for all $t\in [T]$, $k\in [K]$.
\end{lemma}

\begin{lemma}\label{lemma:confidence_ellipsoid}(The confidence bound of our estimation).
    Suppose Assumption~\ref{assm:NTK} holds and assume $||\bm{\theta}^*||_2\leq M$, for some positive constant $M>0$, then for any $\delta \in (0,1)$, with probability at least $1-\delta$, the distance between the estimate weights vector $\bm{\theta}_t$ and $\bm{\theta}^*$ can be bounded as follows
    \begin{align*}
        &\left\|\bm{\theta}_t - \bm{\theta}^* - \mathbf{A}_t^{-1} \sum_{s=1}^t \frac{\phi(\mathbf{x}_{s,a_s};\mathbf{w}_{s-1})}{\Bar{\sigma}_s}\bm{\theta}_0^\top \nabla_\mathbf{w^{(0)}}\frac{\bm{\phi}(\mathbf{x}_{s,a_s};\mathbf{w}^{(0)})}{\Bar{\sigma}_s}(\mathbf{w}^*-\mathbf{w}^{(0)})\right\|_{\mathbf{A}_t}\\ 
        &\leq 8\sqrt{d\log \left(1+t d(\log HK)/(\Bar{\sigma}_{t}^2d\lambda)\right)\log(4t^2/\delta)} + 4R/\Bar{\sigma}_{t} \log(4t^2/\delta) + \lambda^{1/2}M,
    \end{align*}
    for any $t\in [T]$. The proof is in Appendix~\ref{proof:lemma:confidence_ellipsoid}.
\end{lemma}

\begin{lemma}\label{lemma:neighborhood}(Small neighborhood of the initialization point~\citep{cao2019genbounds}).
    Let $\mathbf{w}, \mathbf{w}'$ be in the neighborhood of $\mathbf{w}_0$, i.e., $\mathbf{w}, \mathbf{w}' \in \mathbb{B}(\mathbf{w}_0, \omega)$ for some $\omega>0$. Consider the neural network defined in Equation~\ref{def:NN}, if the width $m$ and the radius $\omega$ of the neighborhood satisfy
    \begin{align*}
        &m\geq C_0 \max\{dL^2\log(m/\delta),\omega^{-4/3}L^{8/3}\log(TK)\log(m/(\omega\delta))\}, \text{ and, } \omega\leq C_1L^{-5}(\log m)^{-3/2},
    \end{align*}
    then for all $\mathbf{x} \in \{\mathbf{x}_{t,k}\}_{t\in [T],k\in [K]}$, with probability at least $1-\delta$ it holds that
    \begin{align*}
        |\phi_j(\mathbf{x};\mathbf{w}) - \hat{\phi}_j\mathbf{x};\mathbf{w})| \leq C_2 \omega^{4/3}L^3d^{-1/2}\sqrt{m\log m},
    \end{align*}
    where $\hat{\phi}_j(\mathbf{x};\mathbf{w})$ is the linearization of $\phi_j(\mathbf{x};\mathbf{w})$ at $\mathbf{w}'$ defined as follow:
    \begin{align*}
        \hat{\phi}_j(\mathbf{x};\mathbf{w}) = \phi_j(\mathbf{x};\mathbf{w}') + \left \langle \nabla_{\mathbf{w}}\phi_j(\mathbf{x};\mathbf{w}'), \mathbf{w} - \mathbf{w}'\right \rangle.
    \end{align*}
\end{lemma}
\begin{lemma}\label{lemma:M_bound}(Extra term of confidence bound~\citep{xu2022neural}).
    Assume that $\mathbf{A}_t=\lambda \mathbf{I}+\sum_{s=1}^t \bm{\phi}_s \bm{\phi}_s^\top$, where $\bm{\phi}_t \in \mathbb{R}^d$ and $||\bm{\phi}_t||\leq G$ for all $t\geq 1$ and some constants $\lambda,G>0$. Let $\{\zeta_t\}_{t=1.\cdots}$ be a real-value sequence s.t. $|\zeta_t|\leq U$ for some constant $U>0$. Then we have
    \begin{align*}
        \left\| \mathbf{A}_t^{-1} \sum_{s=1}^t \bm{\phi}_s \zeta_s \right\|_2 \leq 2Ud, \quad \forall t = 1,2,\cdots.
    \end{align*}
\end{lemma}

\subsubsection{Proof of Theorem~\ref{theo:regret_upper_bound}}
\begin{proof}\label{proof:theo:regret_upper_bound}
    For a time horizon $T$, without loss of generality, assume $T=QH$ for epoch number $Q$, episode length $H$ to backpropagte $f$, then we have the regret as follows
    \begin{align}
        \text{Regret}(T) = \mathbb{E}\left[\sum_{t=1}^T \left(r_{t,a_t^*} - r_{t,a_t}\right) \right] = \mathbb{E}\left[\sum_{q=1}^Q \sum_{i=1}^H \left(r_{qH_{i+1},a_{qH_{i+1}}^*} - r_{qH_{i+1},a_{qH_{i+1}}}\right) \right],
    \end{align}
    i.e., we rewrite the time index $t = qH+1$ as the $i$-th iteration in the $q$-th epoch. By Lemma~\ref{lemma:appr_reward}, there exists vector $\mathbf{w}^*\in \mathbb{R}^p$ s.t. we can write the expectation of the reward generating function as a linear function
    \begin{align}
        h(\mathbf{x}_{t,a_t^*}) - h(\mathbf{x}_{t,a_t})&= \bm{\theta}_0^\top \left[\nabla_{\mathbf{w}^{(0)}} \bm{\phi}(\mathbf{x}_{t,a_t^*};\mathbf{w}^{(0)}) - \nabla_{\mathbf{w}^{(0)}} \bm{\phi}(\mathbf{x}_{t,a_t};\mathbf{w}^{(0)})\right] (\mathbf{w}^* - \mathbf{w}^{(0)})\nonumber\\
        &\quad+ \bm{\theta}^{*\top}\left[\bm{\phi}(\mathbf{x}_{t,a_t^*};\mathbf{w}_{t-1}) - \bm{\phi}(\mathbf{x}_{t,a_t};\mathbf{w}_{t-1})\right]\\
        &= \bm{\theta}_0^\top \left[\nabla_{\mathbf{w}^{(0)}} \bm{\phi}(\mathbf{x}_{t,a_t^*};\mathbf{w}^{(0)}) - \nabla_{\mathbf{w}^{(0)}} \bm{\phi}(\mathbf{x}_{t,a_t};\mathbf{w}^{(0)})\right] (\mathbf{w}^* - \mathbf{w}^{(0)})\nonumber\\
        &\quad+ \bm{\theta}_{t-1}^\top\left[\bm{\phi}(\mathbf{x}_{t,a_t^*};\mathbf{w}_{t-1}) - \bm{\phi}(\mathbf{x}_{t,a_t};\mathbf{w}_{t-1})\right] - (\bm{\theta}_{t-1}-\bm{\theta}^*)^\top \left[\bm{\phi}(\mathbf{x}_{t,a_t^*};\mathbf{w}_{t-1}) - \bm{\phi}(\mathbf{x}_{t,a_t};\mathbf{w}_{t-1})\right].
    \end{align}
    For the first term, we can bound as follows
    \begin{align}
        &\bm{\theta}_0^\top \left[\nabla_{\mathbf{w}^{(0)}} \bm{\phi}(\mathbf{x}_{t,a_t^*};\mathbf{w}^{(0)}) - \nabla_{\mathbf{w}^{(0)}} \bm{\phi}(\mathbf{x}_{t,a_t};\mathbf{w}^{(0)})\right] (\mathbf{w}^* - \mathbf{w}^{(0)})\nonumber\\
        &\leq ||\bm{\theta}_0||_2 \left\|\nabla_{\mathbf{w}^{(0)}} \bm{\phi}(\mathbf{x}_{t,a_t^*};\mathbf{w}^{(0)}) - \nabla_{\mathbf{w}^{(0)}} \bm{\phi}(\mathbf{x}_{t,a_t};\mathbf{w}^{(0)})\right\|_2 ||\mathbf{w}^* - \mathbf{w}^{(0)}||_2\\
        &\leq \ell_{Lip} ||\bm{\theta}_0||_2 ||\mathbf{x}_{t,a_t^*} - \mathbf{x}_{t,a_t}||_2 ||\mathbf{w}^* - \mathbf{w}^{(0)}||_2 \text{ (by Assumption~\ref{assm:grad_norm}).}
    \end{align}
    For the second term, we can bound as follows, by the UCB algorithm, we have
    \begin{align}
        \bm{\theta}_{t-1}^\top \bm{\phi}(\mathbf{x}_{t,a_t^*};\mathbf{w}_{t-1}) + \alpha_t ||\bm{\phi}(\mathbf{x}_{t,a_t^*};\mathbf{w}_{t-1})||_{\mathbf{A}_{t-1}^{-1}} \leq \bm{\theta}_{t-1}^\top \bm{\phi}(\mathbf{x}_{t,a_t};\mathbf{w}_{t-1}) + \alpha_t ||\bm{\phi}(\mathbf{x}_{t,a_t};\mathbf{w}_{t-1})||_{\mathbf{A}_{t-1}^{-1}},
    \end{align}
    so, we get
    \begin{align}
        \bm{\theta}_{t-1}^\top\left[\bm{\phi}(\mathbf{x}_{t,a_t^*};\mathbf{w}_{t-1}) - \bm{\phi}(\mathbf{x}_{t,a_t};\mathbf{w}_{t-1})\right] \leq \alpha_t ||\bm{\phi}(\mathbf{x}_{t,a_t};\mathbf{w}_{t-1})||_{\mathbf{A}_{t-1}^{-1}} - \alpha_t ||\bm{\phi}(\mathbf{x}_{t,a_t^*};\mathbf{w}_{t-1})||_{\mathbf{A}_{t-1}^{-1}}.
    \end{align}
    For the last term, we need to prove that the estimate of weights parameter $\bm{\theta}_{t-1}$ lies in a confidence ball centered at $\bm{\theta}^*$. For the ease of notation, we define
    \begin{align}
        \mathbf{M}_t = \mathbf{A}_t^{-1} \sum_{s=1}^t \frac{\bm{\phi}(\mathbf{x}_{s,a_s};\mathbf{w}_{s-1})}{\Bar{\sigma}_s}\bm{\theta}_0^\top \nabla_{\mathbf{w}^{(0)}} \frac{\bm{\phi}(\mathbf{x}_{s,a_s};\mathbf{w}^{(0)}}{\Bar{\sigma}_s})(\mathbf{w}^* - \mathbf{w}^{(0)}).
    \end{align}
    Then the last term can be bounded as follows
    \begin{align}
        &- (\bm{\theta}_{t-1}-\bm{\theta}^*)^\top \left[\bm{\phi}(\mathbf{x}_{t,a_t^*};\mathbf{w}_{t-1}) - \bm{\phi}(\mathbf{x}_{t,a_t};\mathbf{w}_{t-1})\right]\nonumber\\
        &= -(\bm{\theta}_{t-1} - \bm{\theta}^* - \mathbf{M}_{t-1})^\top \bm{\phi}(\mathbf{x}_{t,a_t^*};\mathbf{w}_{t-1}) + (\bm{\theta}_{t-1} - \bm{\theta}^* - \mathbf{M}_{t-1})^\top \bm{\phi}(\mathbf{x}_{t,a_t};\mathbf{w}_{t-1})\nonumber\\
        &\quad- \mathbf{M}_{t-1}^\top \left[\bm{\phi}(\mathbf{x}_{t,a_t^*};\mathbf{w}_{t-1}) - \bm{\phi}(\mathbf{x}_{t,a_t};\mathbf{w}_{t-1}) \right]\\
        &\leq ||\bm{\theta}_{t-1} - \bm{\theta}^* - \mathbf{M}_{t-1}||_{\mathbf{A}_{t-1}} ||\bm{\phi}(\mathbf{x}_{t,a_t^*};\mathbf{w}_{t-1})||_{\mathbf{A}_{t-1}^{-1}} + ||\bm{\theta}_{t-1} - \bm{\theta}^* - \mathbf{M}_{t-1}||_{\mathbf{A}_{t-1}} ||\bm{\phi}(\mathbf{x}_{t,a_t};\mathbf{w}_{t-1})||_{\mathbf{A}_{t-1}^{-1}}\nonumber\\
        &\quad +\left\|\mathbf{M}_{t-1}^\top \left[\bm{\phi}(\mathbf{x}_{t,a_t^*};\mathbf{w}_{t-1} - \mathbf{x}_{t,a_t};\mathbf{w}_{t-1})\right]\right\|\\
        &\leq \alpha_t ||\bm{\phi}(\mathbf{x}_{t,a_t^*};\mathbf{w}_{t-1})||_{\mathbf{A}_{t-1}^{-1}} + \alpha_t ||\bm{\phi}(\mathbf{x}_{t,a_t};\mathbf{w}_{t-1})||_{\mathbf{A}_{t-1}^{-1}} + ||\mathbf{M}_{t-1}||_2 ||\bm{\phi}(\mathbf{x}_{t,a_t^*};\mathbf{w}_{t-1})-\bm{\phi}(\mathbf{x}_{t,a_t};\mathbf{w}_{t-1})||_2\nonumber\\ 
        &\quad \text{ (by Lemma~\ref{lemma:confidence_ellipsoid} and the choice of $\alpha_t$)}.
    \end{align}
    Hence, we get
    \begin{align}\label{proof:theorem:eq1}
        h(\mathbf{x}_{t,a_t^*}) - h(\mathbf{x}_{t,a_t}) &\leq \ell_{Lip} ||\bm{\theta}_0||_2 ||\mathbf{x}_{t,a_t^*} - \mathbf{x}_{t,a_t}||_2 ||\mathbf{w}^* - \mathbf{w}^{(0)}||_2 + 2\alpha_t ||\bm{\phi}(\mathbf{x}_{t,a_t};\mathbf{w}_{t-1})||_{\mathbf{A}_{t-1}^{-1}}\nonumber\\
        &\quad +||\mathbf{M}_{t-1}||_2 ||\bm{\phi}(\mathbf{x}_{t,a_t^*};\mathbf{w}_{t-1})-\bm{\phi}(\mathbf{x}_{t,a_t};\mathbf{w}_{t-1})||_2.
    \end{align}
    Recall the linearization of $\phi_j$ in Lemma~\ref{lemma:neighborhood}, we have
    \begin{align}
        \hat{\phi}(\mathbf{x},\mathbf{w}_{t-1}) = \phi(\mathbf{x},\mathbf{w}_0) + \nabla_{\mathbf{w}_0} \bm{\phi}(\mathbf{x};\mathbf{w}_0)(\mathbf{w}_{t-1}-\mathbf{w}_0).
    \end{align}
    Note that by the initialization, we have $\phi(\mathbf{x};\mathbf{w}_0)=\mathbf{0}$ for any $\mathbf{x} \in \mathbb{R}^d$. Thus, it holds that
    \begin{align}
        \phi(\mathbf{x}_{t,a_t^*};\mathbf{w}_{t-1}) - \phi(\mathbf{x}_{t,a_t};\mathbf{w}_{t-1}) &= \phi(\mathbf{x}_{t,a_t^*};\mathbf{w}_{t-1}) + \phi(\mathbf{x}_{t,a_t^*};\mathbf{w}_0) + \phi(\mathbf{x}_{t,a_t};\mathbf{w}_0) -  \phi(\mathbf{x}_{t,a_t};\mathbf{w}_{t-1})\\
        &= \phi(\mathbf{x}_{t,a_t^*};\mathbf{w}_{t-1}) - \hat{\phi}(\mathbf{x}_{t,a_t^*};\mathbf{w}_{t-1}) + \nabla_{\mathbf{w}_0} \bm{\phi}(\mathbf{x}_{t,a_t^*};\mathbf{w}_0)(\mathbf{w}_{t-1}-\mathbf{w}_0)\nonumber\\
        &\quad + \phi(\mathbf{x}_{t,a_t};\mathbf{w}_{t-1}) - \hat{\phi}(\mathbf{x}_{t,a_t};\mathbf{w}_{t-1}) - \nabla_{\mathbf{w}_0} \bm{\phi}(\mathbf{x}_{t,a_t};\mathbf{w}_0)(\mathbf{w}_{t-1}-\mathbf{w}_0),
    \end{align}
    yielding
    \begin{align}
        &||\phi(\mathbf{x}_{t,a_t^*};\mathbf{w}_{t-1}) - \phi(\mathbf{x}_{t,a_t};\mathbf{w}_{t-1})||_2\nonumber\\
        &\leq ||\phi(\mathbf{x}_{t,a_t^*};\mathbf{w}_{t-1}) - \hat{\phi}(\mathbf{x}_{t,a_t^*};\mathbf{w}_{t-1})||_2 + ||\phi(\mathbf{x}_{t,a_t};\mathbf{w}_{t-1}) - \hat{\phi}(\mathbf{x}_{t,a_t};\mathbf{w}_{t-1})||_2\nonumber\\
        &\quad + ||\left(\nabla_{\mathbf{w}_0} \bm{\phi}(\mathbf{x}_{t,a_t^*};\mathbf{w}_0) - \nabla_{\mathbf{w}_0} \bm{\phi}(\mathbf{x}_{t,a_t};\mathbf{w}_0)\right)(\mathbf{w}_{t-1}-\mathbf{w}_0)||_2\\
        &\leq C_0 \omega^{4/3}L^3d^{1/2}\sqrt{m\log m} + \ell_{\text{Lip}}||\mathbf{x}_{t,a_t^*} - \mathbf{x}_{t,a_t}||_2||\mathbf{w}_{t-1}-\mathbf{w}^{(0)}||_2 \text{ (by Lemma~\ref{lemma:neighborhood} and Assumption~\ref{assm:grad_norm})}.
    \end{align}
    Plugging into Equation~\ref{proof:theorem:eq1}, we get
    \begin{align}\label{proof:theorem:eq2}
        h(\mathbf{x}_{t,a_t^*}) - h(\mathbf{x}_{t,a_t}) &\leq \ell_{Lip} ||\bm{\theta}_0||_2 ||\mathbf{x}_{t,a_t^*} - \mathbf{x}_{t,a_t}||_2 ||\mathbf{w}^* - \mathbf{w}^{(0)}||_2 + 2\alpha_t ||\bm{\phi}(\mathbf{x}_{t,a_t};\mathbf{w}_{t-1})||_{\mathbf{A}_{t-1}^{-1}}\nonumber\\
        &\quad +||\mathbf{M}_{t-1}||_2 ||\bm{\phi}(\mathbf{x}_{t,a_t^*};\mathbf{w}_{t-1})-\bm{\phi}(\mathbf{x}_{t,a_t};\mathbf{w}_{t-1})||_2\\
        &\leq \ell_{Lip} ||\bm{\theta}_0||_2 ||\mathbf{x}_{t,a_t^*} - \mathbf{x}_{t,a_t}||_2 ||\mathbf{w}^* - \mathbf{w}^{(0)}||_2 + 2\alpha_t ||\bm{\phi}(\mathbf{x}_{t,a_t};\mathbf{w}_{t-1})||_{\mathbf{A}_{t-1}^{-1}}\nonumber\\
        &\quad + ||\mathbf{M}_{t-1}||_2 \left(C_0 \omega^{4/3}L^3d^{1/2}\sqrt{m\log m} + \ell_{\text{Lip}}||\mathbf{x}_{t,a_t^*} - \mathbf{x}_{t,a_t}||_2||\mathbf{w}_{t-1}-\mathbf{w}^{(0)}||_2\right).
    \end{align}
    By Remark~\ref{rem:context_norm}, we have $||\mathbf{x}_{t,a_t^*} - \mathbf{x}_{t,a_t}||_2 \leq 2$. By Lemma~\ref{lemma:appr_reward} and Lemma~\ref{lemma:NN_bounds}, we have
    \begin{align}
        ||\mathbf{w}^* - \mathbf{w}^{(0)}||_2 \leq \sqrt{1/m (\mathbf{r}-\tilde{\mathbf{r}})^\top \mathbf{H}^{-1}(\mathbf{r}-\tilde{\mathbf{r}})}\quad \text{and}\quad ||\mathbf{w}_t - \mathbf{w}^{(0)}||_2 \leq \frac{\delta^{3/2}}{m^{1/2}Tn^{9/2}L^6\log^3(m)}.
    \end{align}
    Additionally, since the entries of $\bm{\theta}_0$ are i.i.d. generated from $\mathcal{N}(0,1/d)$, we have $||\bm{\theta}_0||_2 \leq 2(2+\sqrt{d^{-1}\log(1/\delta)})$ with probability at least $1-\delta$ for any $\delta>0$. By Lemma~\ref{lemma:NN_bounds}, we have $||\nabla_{\mathbf{w}^{(0)}}\bm{\phi}(\mathbf{x}_{t,a_t};\mathbf{w}^{(0)})||_F \leq C_1\sqrt{dm}$. Therefore, 
    \begin{align}
        |\bm{\theta}_0^\top \nabla_{\mathbf{w}^{(0)}}\bm{\phi}(\mathbf{x}_{s,a_s};\mathbf{w}^{(0)})(\mathbf{w}^* - \mathbf{w}^{(0)})| \leq C_2d\sqrt{\log(1/\delta)(\mathbf{r}-\tilde{\mathbf{r}})^\top \mathbf{H}(\mathbf{r}-\tilde{\mathbf{r}})}.
    \end{align}
    Then, by the definition of $\mathbf{M}_t$ and Lemma~\ref{lemma:M_bound}, we have
    \begin{align}
        ||\mathbf{M}_{t-1}||_2 \leq C_3d^2\sqrt{\log(1/\delta)(\mathbf{r}-\tilde{\mathbf{r}})^\top \mathbf{H}^{-1} (\mathbf{r}-\tilde{\mathbf{r}})}.
    \end{align}
    Continue plugging into Equation~\ref{proof:theorem:eq2}, we get
    \begin{align}
        h(\mathbf{x}_{t,a_t^*}) - h(\mathbf{x}_{t,a_t})
        &\leq C_4\ell_{Lip} m^{-1/2}\sqrt{\log(1/\delta)(\mathbf{r}-\tilde{\mathbf{r}})^\top \mathbf{H}^{-1} (\mathbf{r}-\tilde{\mathbf{r}})} + 2\alpha_t ||\bm{\phi}(\mathbf{x}_{t,a_t};\mathbf{w}_{t-1})||_{\mathbf{A}_{t-1}^{-1}}\nonumber\\
        &\quad + \left(C_0 \omega^{4/3}L^3d^{1/2}\sqrt{m\log m} + \frac{2\ell_{\text{Lip}}\delta^{3/2}}{m^{1/2}Tn^{9/2}L^6\log^3(m)}\right) C_3d^2\sqrt{\log(1/\delta)(\mathbf{r}-\tilde{\mathbf{r}})^\top \mathbf{H}^{-1} (\mathbf{r}-\tilde{\mathbf{r}})}.
    \end{align}
    Combining with the fact that $\omega = \mathcal{O}\left(m^{-1/2}||\mathbf{r} - \tilde{\mathbf{r}}||_{\mathbf{H}^{-1}}\right)$ by Lemma~\ref{lemma:appr_reward}, using Cauchy’s inequality, we obtain
    \begin{align}\label{eq:regret_proof:theo:regret_upper_bound}
        \text{Regret}(T) &\leq \sqrt{\sum_{t=1}^T 4\alpha_t^2 \Bar{\sigma}_t^2 ||\bm{\phi}(\mathbf{x}_{t,a_t};\mathbf{w}_{t-1})/\Bar{\sigma}_t||_{\mathbf{A}_{t-1}^{-1}}^2} + C_4 \ell_{\text{Lip}}m^{-1/2}T\sqrt{\log(1/\delta)}||\mathbf{r}-\tilde{\mathbf{r}}||_{\mathbf{H}^{-1}}\nonumber\\
        &\quad + \left(\frac{C_0 TL^3d^{1/2}\sqrt{\log m}||\mathbf{r}-\tilde{\mathbf{r}}||_{\mathbf{H}^{-1}}^{4/3}}{m^{1/6}} + \frac{2\ell_{\text{Lip}}\delta^{3/2}}{m^{1/2}n^{9/2}L^6\log^3(m)}\right)C_3d^2\sqrt{\log(1/\delta)}||\mathbf{r}-\tilde{\mathbf{r}}||_{\mathbf{H}^{-1}}.
    \end{align}
    Let $\sigma_{T_{\min}} = \min_{t\in [T]}\sigma_t$, since $\bar{\sigma}_t= \max\{R/\sqrt{d},\sigma_t\}$, yielding $\bar{\sigma}_{T_{\min}} \geq R/\sqrt{d}$, therefore, we have
    \begin{align}
        \alpha_t &= 8\sqrt{d\log \left(1+t d(\log HK)/(\Bar{\sigma}_{t}^2d\lambda)\right)\log(4t^2/\delta)} + 4R/\Bar{\sigma}_{t} \log(4t^2/\delta) + \lambda^{1/2}M\\
        &\leq  8\sqrt{d\log \left(1+T d(\log HK)/(\lambda R^2)\right)\log(4T^2/\delta)} + 4R/\sigma_{T_{\min}} \log(4T^2/\delta) + \lambda^{1/2}M.
    \end{align}
    Since $\Bar{\sigma}_t^2=\max\{R^2/d,\sigma_t^2\}\leq R^2/d + \sigma_t^2$, $\sqrt{|x|+|y|}\leq \sqrt{|x|}+\sqrt{|y|}$, using the upper bound of $\alpha_t$ in Lemma~\ref{lemma:confidence_ellipsoid} and Lemma~\ref{lemma:feature_COV_bound}, we finally obtain
    \begin{align}
        \text{Regret}(T) &\leq C_5\alpha_T\sqrt{\left(TR^2+d\sum_{t=1}^T \sigma_t^2\right) \log(1+TG^2/(\lambda R^2))}\nonumber\\
        &\quad + C_6\ell_{Lip}L^3 d^{5/2}m^{-1/6}T\sqrt{\log m \log(1/\delta) \log(TK/\delta)} ||\mathbf{r} - \tilde{\mathbf{r}}||_{\mathbf{H}^{-1}}
    \end{align}
    of Theorem~\ref{theo:regret_upper_bound}.
\end{proof}
\subsubsection{Proof of Corollary~\ref{corol:regret_upper_bound}}
\begin{proof}
    It directly follows the result in Theorem~\ref{theo:regret_upper_bound}  by using the Big $\mathcal{O}$ notation.
\end{proof}
\subsection{Proof of Lemma~\ref{lemma:confidence_ellipsoid}}\label{proof:lemma:confidence_ellipsoid}
This proof is based on the following provable Lemma of~\citet{zhou2021nearly}:
\begin{lemma}\label{lemma:bernstein}(Bernstein inequality for vector-valued martingales~\citep{zhou2021nearly}).
    Let $\{\mathcal{G}_t\}_{t=1}^\infty$ be a filtration, $\{x_t, \xi_t\}_{t\geq 1}$ a stochastic process so that $x_t \in \mathbb{R}^d$ is $\mathcal{G}_t$-measurable and $\xi_t \in \mathbb{R}$ is $\mathcal{G}_{t+1}$-measurable. Fix $R,G,\sigma,\lambda>0$, $\bm{\theta}^*\in \mathbb{R}^d$. For $t\geq 1$ let $r_t=\langle\bm{\theta}^*, x_t\rangle+\xi_t$ and suppose that $\xi_t,x_t$ also satisfy
    \begin{align*}
        |\xi_t| \leq R,\quad \mathbb{E}[\xi_t|\mathcal{G}_t] = 0,\quad \mathbb{E}[\xi_t^2|\mathcal{G}_t] \leq \sigma^2,\quad ||x_t||_2 \leq G. 
    \end{align*}
    Then, for any $0\leq \delta\leq 1$, with prob. at least $1-\delta$, we have
    \begin{align*}
        \forall t>0,\quad \left\|\sum_{i=1}^t x_i\xi_i\right\|_{V_t^{-1}}\leq \beta_t,\quad ||\bm{\theta}_t - \bm{\theta}^*||_{V_t} \leq \beta_t + \sqrt{\lambda}||\bm{\theta}^*||_2,
    \end{align*}
    where for $t\geq 1$, $\bm{\theta}_t = V_t^{-1}b_t$, $V_t=\lambda \mathbf{I} + \sum_{i=1}^t x_i x_i^\top$, $b_t=\sum_{i=1}^t r_i x_i$ and
    \begin{align*}
        \beta_t = 8\sigma\sqrt{d\log(1+tG^2/(d\lambda))\log(4t^2/\delta)} + 4R\log(4t^2/\delta).
    \end{align*}
\end{lemma}
Now, we provide our proof of Lemma~\ref{lemma:confidence_ellipsoid} as follows:
\begin{proof}
    Let $\bm{\Phi}_t = [\bm{\phi}(\mathbf{x}_{1,a_1}; \mathbf{w}_0), \cdots, \bm{\phi}(\mathbf{x}_{1,a_t}; \mathbf{w}_{t-1})]\in \mathbb{R}^{d\times t}$ be the collection of feature vectors of the chosen arms up to time $t$ and $\mathbf{r}_t = (r_{1,a_1}, \cdots, r_{t,a_t})^\top$ be the concatenation of all received rewards. According to Algorithm~\ref{alg:ours}, we have $\mathbf{A}_t = \lambda \mathbf{I} + \frac{\bm{\Phi}_t \bm{\Phi}_t^\top}{\Bar{\sigma}^2_t}$ and thus
    \begin{align}
        \bm{\theta}_t = \mathbf{A}_t^{-1} \mathbf{b}_t = \left(\lambda \mathbf{I} + \frac{\bm{\Phi}_t \bm{\Phi}_t^\top}{\Bar{\sigma}^2_t}\right)^{-1} \frac{\bm{\Phi}_t \mathbf{r}_t}{\Bar{\sigma}^2_t}.
    \end{align}
    By Lemma~\ref{lemma:appr_reward}, we can rewrite the reward as
    \begin{align}
        r_{t,a_t} = \left \langle \bm{\theta}^*, \bm{\phi}(\mathbf{x}_{t,a_t};\mathbf{w}_{t-1})  \right \rangle + \bm{\theta}_0^\top \nabla_\mathbf{w^{(0)}}\bm{\phi}(\mathbf{x}_{t,a_t};\mathbf{w}^{(0)})(\mathbf{w}^*-\mathbf{w}^{(0)}).
    \end{align}
    Therefore, it holds that
    \begin{align}
        \bm{\theta}_t &= \mathbf{A}_t^{-1}\frac{\bm{\Phi}_t \mathbf{\Phi}_t^\top}{\Bar{\sigma}_t^2}\bm{\theta}^* + \mathbf{A}_t^{-1} \sum_{s=1}^t \frac{\bm{\phi}(\mathbf{x}_{s,a_s};\mathbf{w}_{s-1})}{\Bar{\sigma}_s}\left(\bm{\theta}_0^\top \nabla_\mathbf{w^{(0)}}\frac{\bm{\phi}(\mathbf{x}_{s,a_s};\mathbf{w}^{(0)})}{\Bar{\sigma}_s}(\mathbf{w}^*-\mathbf{w}^{(0)}) + \frac{\xi_s}{\Bar{\sigma}_s}\right)\\
        &= \bm{\theta}^* - \lambda \mathbf{A}_t^{-1}\bm{\theta}^* + \mathbf{A}_t^{-1} \sum_{s=1}^t \frac{\bm{\phi}(\mathbf{x}_{s,a_s};\mathbf{w}_{s-1})}{\Bar{\sigma}_s}\left(\bm{\theta}_0^\top \nabla_\mathbf{w^{(0)}}\frac{\bm{\phi}(\mathbf{x}_{s,a_s};\mathbf{w}^{(0)})}{\Bar{\sigma}_s}(\mathbf{w}^*-\mathbf{w}^{(0)}) + \frac{\xi_s}{\Bar{\sigma}_s}\right).
    \end{align}
    Then for any $\delta \in (0,1)$, by triangle inequality, we have
    \begin{align}
        \left\|\bm{\theta}_t - \bm{\theta}^* - \mathbf{A}_t^{-1} \frac{\bm{\Phi}_t}{\Bar{\sigma}_t} \bm{\Theta}_t \nabla_{\mathbf{w}^{(0)}} \frac{\bm{\Phi}_t}{\Bar{\sigma}_t}(\mathbf{w}^*-\mathbf{w}^{(0)})\right\|_{\mathbf{A}_t} \leq \lambda||\bm{\theta}^*||_{\bm{A}_t^{-1}} + \left\|\frac{\bm{\Phi}_t}{\Bar{\sigma}_t} \frac{\bm{\xi}_t}{\Bar{\sigma}_t}\right\|_{\mathbf{A}_t^{-1}}.
    \end{align}
    Applying Lemma~\ref{lemma:bernstein} to 
    \begin{align}
        |\xi_t/\Bar{\sigma}_t| \leq R/\Bar{\sigma}_t,\quad \mathbb{E}[(\xi_t/\Bar{\sigma}_t)|\mathcal{G}_t] = 0,\quad \mathbb{E}[(\xi_t/\Bar{\sigma}_t)^2|\mathcal{G}_t] \leq 1,\quad ||\bm{\phi}(\mathbf{x};\mathbf{w})/\Bar{\sigma}_t||_2 \leq G/\Bar{\sigma}_t,
    \end{align}
    and the fact that $||\bm{\phi}(\mathbf{x};\mathbf{w})||_2 \leq C\sqrt{d\log HK}$, we get
    \begin{align}
        \left\|\frac{\bm{\Phi}_t}{\Bar{\sigma}_t} \frac{\bm{\xi}_t}{\Bar{\sigma}_t}\right\|_{\mathbf{A}_t^{-1}} \leq 8\sqrt{d\log \left(1+tC^2 d(\log HK)/(\Bar{\sigma}_{t}^2d\lambda)\right)\log(4t^2/\delta)} + 4R/\Bar{\sigma}_{t} \log(4t^2/\delta).
    \end{align}
    Combining with the fact that $||\bm{\theta}^*||_{\mathbf{A}_t^{-1}}\leq \lambda^{-1/2}||\bm{\theta}^*||_2\leq \lambda^{-1/2}M$ by Lemma~\ref{lemma:appr_reward} and the assumption that $||\bm{\theta}^*||_2\leq M$, we obtain
    \begin{align}
        &\left\|\bm{\theta}_t - \bm{\theta}^* - \mathbf{A}_t^{-1} \sum_{s=1}^t \frac{\bm{\phi}(\mathbf{x}_{s,a_s};\mathbf{w}_{s-1})}{\Bar{\sigma}_s}\bm{\theta}_0^\top \nabla_\mathbf{w^{(0)}}\frac{\bm{\phi}(\mathbf{x}_{s,a_s};\mathbf{w}^{(0)})}{\Bar{\sigma}_s}(\mathbf{w}^*-\mathbf{w}^{(0)})\right\|_{\mathbf{A}_t}\nonumber\\ 
        &\leq 8\sqrt{d\log \left(1+tC^2 d(\log HK)/(\Bar{\sigma}_{t}^2d\lambda)\right)\log(4t^2/\delta)} + 4R/\Bar{\sigma}_{t} \log(4t^2/\delta) + \lambda^{1/2}M
    \end{align}
    of Lemma~\ref{lemma:confidence_ellipsoid}.
\end{proof}

\subsection{Proof of Theorem~\ref{theo:regret_upper_bound_alg}}\label{proof:thm:regret_upper_bound_alg}
\begin{proof}
    Firstly, we can bound the estimation error of the upper bound of the reward noise variance at round $t$ as follows
    \begin{align}
        \left|\sigma^2_t - \hat{\sigma}^2_t\right| &= \left| \left(b - {\bm{\theta}^*}^\top\bm{\phi}(\mathbf{x}_{t,a_t};\mathbf{w}^*)\right)\left({\bm{\theta}^*}^\top\bm{\phi}(\mathbf{x}_{t,a_t};\mathbf{w}^*)-a\right) - \left(b - \bm{\theta}_{t}^\top \bm{\phi}(\mathbf{x}_{t,a_t};\mathbf{w})\right)\left(\bm{\theta}_{t}^\top \bm{\phi}(\mathbf{x}_{t,a_t};\mathbf{w})-a\right)\right|\\
        &= \left| \left(b \cdot {\bm{\theta}^*}^\top\bm{\phi}(\mathbf{x}_{t,a_t};\mathbf{w}^*) - ba - {\bm{\theta}^*}^\top\bm{\phi}(\mathbf{x}_{t,a_t};\mathbf{w}^*){\bm{\theta}^*}^\top\bm{\phi}(\mathbf{x}_{t,a_t};\mathbf{w}^*) +{\bm{\theta}^*}^\top\bm{\phi}(\mathbf{x}_{t,a_t};\mathbf{w}^*)\cdot a\right) \right.\nonumber \\
        & \quad \left.{} - \left(b \cdot \bm{\theta}_{t}^\top \bm{\phi}(\mathbf{x}_{t,a_t};\mathbf{w}) - ba - \bm{\theta}_{t}^\top \bm{\phi}(\mathbf{x}_{t,a_t};\mathbf{w}) \bm{\theta}_{t}^\top \bm{\phi}(\mathbf{x}_{t,a_t};\mathbf{w}) + \bm{\theta}_{t}^\top \bm{\phi}(\mathbf{x}_{t,a_t};\mathbf{w}) \cdot a\right)\vphantom{\frac12} \right|\\
        &= \left| (b+a) \left({\bm{\theta}^*}^\top\bm{\phi}(\mathbf{x}_{t,a_t};\mathbf{w}^*) - \bm{\theta}_{t}^\top \bm{\phi}(\mathbf{x}_{t,a_t};\mathbf{w}) \right) \right.\nonumber\\
        &\quad \left.{}- \left[{\bm{\theta}^*}^\top\bm{\phi}(\mathbf{x}_{t,a_t};\mathbf{w}^*){\bm{\theta}^*}^\top\bm{\phi}(\mathbf{x}_{t,a_t};\mathbf{w}^*) -  \bm{\theta}_{t}^\top \bm{\phi}(\mathbf{x}_{t,a_t};\mathbf{w}) \bm{\theta}_{t}^\top \bm{\phi}(\mathbf{x}_{t,a_t};\mathbf{w})\right]\vphantom{\frac12} \right|\\
        &= \left| (b+a) \left({\bm{\theta}^*}^\top\bm{\phi}(\mathbf{x}_{t,a_t};\mathbf{w}^*) - \bm{\theta}_{t}^\top \bm{\phi}(\mathbf{x}_{t,a_t};\mathbf{w}) \right) \right.\nonumber\\
        &\quad \left.{}- \left[\left({\bm{\theta}^*}^\top\bm{\phi}(\mathbf{x}_{t,a_t};\mathbf{w}^*) - \bm{\theta}_{t}^\top \bm{\phi}(\mathbf{x}_{t,a_t};\mathbf{w}) \right) \left({\bm{\theta}^*}^\top\bm{\phi}(\mathbf{x}_{t,a_t};\mathbf{w}^*) + \bm{\theta}_{t}^\top \bm{\phi}(\mathbf{x}_{t,a_t};\mathbf{w}) \right)\right]\vphantom{\frac12} \right|\\
        &= \left| \left({\bm{\theta}^*}^\top\bm{\phi}(\mathbf{x}_{t,a_t};\mathbf{w}^*) - \bm{\theta}_{t}^\top \bm{\phi}(\mathbf{x}_{t,a_t};\mathbf{w})\right) \left[b+a-\left({\bm{\theta}^*}^\top\bm{\phi}(\mathbf{x}_{t,a_t};\mathbf{w}^*) + \bm{\theta}_{t}^\top \bm{\phi}(\mathbf{x}_{t,a_t};\mathbf{w}\right)\right] \right|\\
        &= \left|{\bm{\theta}^*}^\top\bm{\phi}(\mathbf{x}_{t,a_t};\mathbf{w}^*) - \bm{\theta}_{t}^\top \bm{\phi}(\mathbf{x}_{t,a_t};\mathbf{w})\right| \left|b+a-\left({\bm{\theta}^*}^\top\bm{\phi}(\mathbf{x}_{t,a_t};\mathbf{w}^*) + \bm{\theta}_{t}^\top \bm{\phi}(\mathbf{x}_{t,a_t};\mathbf{w}\right)\right|.
    \end{align}
    By the triangle inequality and Hölder's inequality 
    \begin{align}
        \left|\sigma^2_t - \hat{\sigma}^2_t\right| &\leq \left(\left|{\bm{\theta}^*}^\top\bm{\phi}(\mathbf{x}_{t,a_t};\mathbf{w}^*)\right| + \left|\bm{\theta}_{t}^\top \bm{\phi}(\mathbf{x}_{t,a_t};\mathbf{w})\right|\right) \left(\left|b+a\right| +\left|{\bm{\theta}^*}^\top\bm{\phi}(\mathbf{x}_{t,a_t};\mathbf{w}^*) + \bm{\theta}_{t}^\top \bm{\phi}(\mathbf{x}_{t,a_t};\mathbf{w}\right|\right)\\
        &\leq \left(\left|{\bm{\theta}^*}^\top\bm{\phi}(\mathbf{x}_{t,a_t};\mathbf{w}^*)\right| + \left|\bm{\theta}_{t}^\top \bm{\phi}(\mathbf{x}_{t,a_t};\mathbf{w})\right|\right) \left(\left|b+a\right| +\left|{\bm{\theta}^*}^\top\bm{\phi}(\mathbf{x}_{t,a_t};\mathbf{w}^*)\right| + \left|\bm{\theta}_{t}^\top \bm{\phi}(\mathbf{x}_{t,a_t};\mathbf{w}\right|\right)\\
        &\leq \left(\left|{\bm{\theta}^*}^\top\bm{\phi}(\mathbf{x}_{t,a_t};\mathbf{w}^*)\right| + \left\|\bm{\theta}_{t}\right\|_2 \left\|\bm{\phi}(\mathbf{x}_{t,a_t};\mathbf{w})\right\|\right) \left(\left|b+a\right| +\left|{\bm{\theta}^*}^\top\bm{\phi}(\mathbf{x}_{t,a_t};\mathbf{w}^*)\right| + \left\|\bm{\theta}_{t}\right\|_2 \left\|\bm{\phi}(\mathbf{x}_{t,a_t};\mathbf{w}\right\|\right).
    \end{align}
    Since $0\leq {\bm{\theta}^*}^\top\bm{\phi}(\mathbf{x}_{t,a_t};\mathbf{w}^*) = h(\mathbf{x}_{t,a_t})\leq 1$, we can further bound
    \begin{align}
        \left|\sigma^2_t - \hat{\sigma}^2_t\right| \leq \left(1 + \left\|\bm{\theta}_t\right\|_2 \left\|\bm{\phi}(\mathbf{x}_{t,a_t};\mathbf{w})\right\|\right) \left(\left|b+a\right| + 1 + \left\|\bm{\theta}_t\right\|_2 \left\|\bm{\phi}(\mathbf{x}_{t,a_t};\mathbf{w})\right\|\right).
    \end{align}
    Using the result from proof of Lemma~\ref{lemma:NN_bounds}, we get
    \begin{align}
        ||\bm{\theta}_t||_2 = \left\|\left(\lambda \mathbf{I} + \sum_{i=1}^t \bm{\phi}(\mathbf{x}_{i,a_i};\mathbf{w}_{i-1}) \bm{\phi}(\mathbf{x}_{i,a_i};\mathbf{w}_{i-1})^\top \right)^{-1} \sum_{i=1}^t \bm{\phi}(\mathbf{x}_{i,a_i};\mathbf{w}_{i-1}) \hat{\mathbf{r}} \right\|_2 \leq 2d.
    \end{align}
    On the other hand, by $\bm{\phi}(\mathbf{x}_{t,a_t};\mathbf{w}_0) = \mathbf{0}$, using Lemma~\ref{lemma:neighborhood}, we have
    \begin{align}
        \left\|\bm{\phi}(\mathbf{x}_{t,a_t};\mathbf{w})\right\| &= \left\|\bm{\phi}(\mathbf{x}_{t,a_t};\mathbf{w}) - \bm{\phi}(\mathbf{x}_{t,a_t};\mathbf{w}_0)\right\|\\
        &= \left\|\bm{\phi}(\mathbf{x}_{t,a_t};\mathbf{w}) - \hat{\bm{\phi}}(\mathbf{x}_{t,a_t};\mathbf{w}) +  \left \langle \nabla_{\mathbf{w}^{(0)}} \bm{\phi}(\mathbf{x}_{t,a_k};\mathbf{w}^{(0)}), \left(\mathbf{w}_t - \mathbf{w}^{(0)}\right)\right \rangle\right\|\\
        &\leq \left\|\bm{\phi}(\mathbf{x}_{t,a_t};\mathbf{w}) - \hat{\bm{\phi}}(\mathbf{x}_{t,a_t};\mathbf{w})\right\| +  \left\| \left \langle \nabla_{\mathbf{w}^{(0)}} \bm{\phi}(\mathbf{x}_{t,a_k};\mathbf{w}^{(0)}), \left(\mathbf{w}_t - \mathbf{w}^{(0)}\right)\right \rangle\right\| \text{ (by triangle inequality)}\\
        &\leq \left\|\bm{\phi}(\mathbf{x}_{t,a_t};\mathbf{w}) - \hat{\bm{\phi}}(\mathbf{x}_{t,a_t};\mathbf{w})\right\| + \left\|\nabla_{\mathbf{w}^{(0)}} \bm{\phi}(\mathbf{x}_{t,a_k};\mathbf{w}^{(0)})\right\|_{\mathcal{F}} \left\|\mathbf{w}_t - \mathbf{w}^{(0)}\right\|_2 \text{ (by Hölder's inequality)}\\
        &\leq C_0 \omega^{4/3}L^3d^{-1/2}\sqrt{m\log m} + \frac{\delta^{3/2}\sqrt{d}}{Tn^{9/2}L^{11/2}\log^3(m)} \text{ (by Lemma~\ref{lemma:neighborhood} and~\ref{lemma:NN_bounds})}.
    \end{align}
    Therefore, combining with the fact that $\omega = \mathcal{O}\left(m^{-1/2}\left\|\mathbf{r} - \tilde{\mathbf{r}}\right\|_{\mathbf{H}^{-1}}\right)$ by Lemma~\ref{lemma:appr_reward}, we get
    \begin{align}
        \left\|\bm{\theta}_t\right\|_2 \left\|\bm{\phi}(\mathbf{x}_{t,a_t};\mathbf{w})\right\| &\leq \frac{C_1L^3\sqrt{d\log m}\left\|\mathbf{r} - \tilde{\mathbf{r}}\right\|^{4/3}_{\mathbf{H}^{-1}}}{m^{1/6}} + \frac{2\delta^{3/2}d^{3/2}}{Tn^{9/2}L^{11/2}\log^3(m)}.
    \end{align}
    Using the result that $\mathbf{r}^\top \mathbf{H}^{-1} \mathbf{r}$ can be bounded by the \acrshort{RKHS} norm of $\mathbf{r}$ if it belongs to the \acrshort{RKHS} induced by the \acrshort{NTK}~\citep{xu2022neural, zhou2020neuralUCB, arora2019fineGrained, arora2019onExact, lee2019wide}, we obtain
    \begin{align}\label{eq:est_sigma_bound}
        \left|\sigma^2_t - \hat{\sigma}^2_t\right| 
        &\leq \left(1+\frac{C_1L^3\sqrt{d\log m}\left\|\mathbf{r} - \tilde{\mathbf{r}}\right\|^{4/3}_{\mathbf{H}^{-1}}}{m^{1/6}} + \frac{2\delta^{3/2}d^{3/2}}{Tn^{9/2}L^{11/2}\log^3(m)}\right)\nonumber\\
        &\quad \left(|b+a| + 1 + \frac{C_1L^3\sqrt{d\log m}\left\|\mathbf{r} - \tilde{\mathbf{r}}\right\|^{4/3}_{\mathbf{H}^{-1}}}{m^{1/6}} + \frac{2\delta^{3/2}d^{3/2}}{Tn^{9/2}L^{11/2}\log^3(m)}\right) = \tilde{\mathcal{O}}\left(\frac{d^3}{T^2n^9L^{11}\log^6(m)}\right).
    \end{align}
    Using the regret in Equation~\ref{eq:regret_proof:theo:regret_upper_bound} from the proof of Theorem~\ref{proof:theo:regret_upper_bound}, i.e.,
    \begin{align}
        \text{Regret}(T) &\leq \sqrt{\sum_{t=1}^T 4\alpha_t^2 \Bar{\sigma}_t^2 ||\bm{\phi}(\mathbf{x}_{t,a_t};\mathbf{w}_{t-1})/\Bar{\sigma}_t||_{\mathbf{A}_{t-1}^{-1}}^2} + C_4 \ell_{\text{Lip}}m^{-1/2}T\sqrt{\log(1/\delta)}||\mathbf{r}-\tilde{\mathbf{r}}||_{\mathbf{H}^{-1}}\nonumber\\
        &\quad + \left(\frac{C_0 TL^3d^{1/2}\sqrt{\log m}||\mathbf{r}-\tilde{\mathbf{r}}||_{\mathbf{H}^{-1}}^{4/3}}{m^{1/6}} + \frac{2\ell_{\text{Lip}}\delta^{3/2}}{m^{1/2}n^{9/2}L^6\log^3(m)}\right)C_3d^2\sqrt{\log(1/\delta)}||\mathbf{r}-\tilde{\mathbf{r}}||_{\mathbf{H}^{-1}}.
    \end{align}
    Let $\hat{\sigma}_{T_{\min}} = \min_{t\in [T]}\hat{\sigma}_t$, since $\bar{\sigma}_t= \max\{R/\sqrt{d},\hat{\sigma}_t\}$, yielding $\bar{\sigma}_{T_{\min}} \geq R/\sqrt{d}$, therefore, we have
    \begin{align}
        \alpha_t &= 8\sqrt{d\log \left(1+t d(\log HK)/(\Bar{\sigma}_{t}^2d\lambda)\right)\log(4t^2/\delta)} + 4R/\Bar{\sigma}_{t} \log(4t^2/\delta) + \lambda^{1/2}M\\
        &\leq  8\sqrt{d\log \left(1+T d(\log HK)/(\lambda R^2)\right)\log(4T^2/\delta)} + 4R/\hat{\sigma}_{T_{\min}} \log(4T^2/\delta) + \lambda^{1/2}M = \tilde{\mathcal{O}}\left(\sqrt{d}\right).
    \end{align}
    Since $\Bar{\sigma}_t^2=\max\{R^2/d,\hat{\sigma}_t^2\}\leq R^2/d + \hat{\sigma}_t^2\leq R^2/d + \sigma_t^2 + \tilde{\mathcal{O}}\left(\frac{d^3}{T^2n^9L^{11}\log^6(m)}\right)$ (by Equation~\ref{eq:est_sigma_bound}), $\sqrt{|x|+|y|}\leq \sqrt{|x|}+\sqrt{|y|}$, using the upper bound of $\alpha_t$ in Lemma~\ref{lemma:confidence_ellipsoid} and Lemma~\ref{lemma:feature_COV_bound}, we finally obtain
    \begin{align}
        \text{Regret}(T) &\leq C_5\alpha_T\sqrt{\left[TR^2+d\sum_{t=1}^T \left(\sigma_t^2 + \frac{d^3}{T^2n^9L^{11}\log^6(m)}\right)\right] \log(1+TG^2/(\lambda R^2))}\nonumber\\
        &\quad + C_6\ell_{Lip}L^3 d^{5/2}m^{-1/6}T\sqrt{\log m \log(1/\delta) \log(TK/\delta)} ||\mathbf{r} - \tilde{\mathbf{r}}||_{\mathbf{H}^{-1}}\\            
        &\leq \tilde{\mathcal{O}} \left(R\sqrt{dT} + d\sqrt{\sum_{t=1}^T \sigma_t^2 + \frac{d^3}{T^2n^9L^{11}\log^6(m)}}\right) + \tilde{\mathcal{O}}\left(m^{-1/6}T\sqrt{(\mathbf{r}-\mathbf{\tilde{r}})^\top \mathbf{H}^{-1} (\mathbf{r}-\mathbf{\tilde{r}})}\right)
    \end{align}
    of Theorem~\ref{theo:regret_upper_bound_alg}.
\end{proof}
\section{Experimental Details}\label{apd:exo_details}
\subsection{Demo notebook code for Algorithm~\ref{alg:ours}}\label{apd:code}
\begin{tabular}{cc}
\begin{minipage}{0.1\linewidth}
\end{minipage}&
\begin{minipage}{0.96\linewidth}
\begin{minted}
[
frame=lines,
framesep=2mm,
baselinestretch=1,
bgcolor=LightGray,
fontsize=\fontsize{8.2pt}{8.2pt},
linenos
]
{python}
import torch

class NeuralVarLinearUCB:
 def __init__(self, dim, n_arm=4, lamdba=1, nu_R=1, hidden=100):
  #Initialize model parameters
  self.func = Network(dim, hidden_size=hidden).cuda()
  self.context_list, self.arm_list, self.reward  = [], [], []
  self.theta = np.random.uniform(-1, 1, (self.n_arm, dim))
  self.b = np.zeros((self.n_arm, dim))
  self.A_inv = np.array([np.eye(dim) for _ in range(self.n_arm)])
  self.sigma = self.nu_R/dim

 def select(self, context):
  #Select action by UCB
  features = self.func(context).cpu().detach().numpy()
  ucb = [np.sqrt(np.dot(features[a,:], np.dot(self.A_inv[a], features[a,:].T))) for a in range(self.n_arm)]
  mu = [np.dot(features[a,:], self.theta[a]) for a in range(self.n_arm)]
  arm = np.argmax(mu + ucb)
  return arm, mu[arm]

 def train(self, context, arm_select, reward):
  #Update neural network model parameters
  self.context_list.append(torch.from_numpy(context[arm_select].reshape(1, -1)).float())
  self.arm_list.append(arm_select)
  self.reward.append(reward)
  optimizer = optim.SGD(self.func.parameters(), lr=1e-2, weight_decay=self.lamdba)
  train_set = []
  for idx in range(len(self.context_list)):
   train_set.append((self.context_list[idx], self.arm_list[idx], self.reward[idx]))
  train_loader = DataLoader(train_set, batch_size = 64, shuffle = True)
  for batch_idx, (samples, arms, labels) in enumerate(train_loader):
   optimizer.zero_grad()
   features = self.func(samples.cuda())
   mu = (features * torch.from_numpy(self.theta[arms]).float().cuda()).sum()
   A_inv = torch.from_numpy(self.A_inv[arms]).float().cuda()
   sigma = (features * torch.squeeze(torch.bmm(A_inv, torch.unsqueeze(features, 2)))).sum()
   loss = torch.mean(1/2 * torch.log(2*np.pi*sigma) + (labels-mu)**2/(2*sigma))
   loss.backward()
   optimizer.step()

 def update_model(self, context, arm_select, reward, mu, a_low, b_up):
  #Update linear model parameters
  context = self.func(context)
  self.theta = np.array([np.matmul(self.A_inv[a], self.b[a]) for a in range(self.n_arm)])
  self.sigma = (b_up - mu) * (mu - a_low)		
  self.sigma = max(self.sigma, self.nu_R/dim)
  self.b[arm_select] += (context[arm_select] * reward[arm_select])/self.sigma
  self.A_inv[arm_select] = inv_sherm_morri(context[arm_select,:]/np.sqrt(self.sigma), self.A_inv[arm_select])

if __name__ == "__main__"
 agent = NeuralVarLinearUCB(dim = 20)
 list_regrets = []
 for t in range(T):
  context, rwd, psd_rwd = contexts[t], rewards[t], psd_rewards[t]
  arm_select, mu = agent.select(context)
  regret = np.max(psd_rwd) - psd_rwd[arm_select]
  list_regrets.append(regret)
  agent.update_model(context, arm_select, rwd, mu, a_low, b_up)
  if t%100 == 0:
   agent.train(context, arm_select, rwd[arm_select])
 plot(list_regrets)
\end{minted}
\end{minipage}
\end{tabular}

\subsection{Experimental settings}\label{apd:exp_settings}
\textbf{Dataset and hyper-parameters details},
We deploy the models on five datasets in the main paper. Regarding real-world data, we use the MNIST dataset which contains $70000$, $d=28\times 28$ digit handwriting images with $K=10$ classes~\citep{lecun2010MNIST}; the UCI-shuttle~(statlog) dataset related to physics and chemistry area, containing $58000$ features with $d=9$ numerical attributes and $K=7$ classes~\citep{dua2017UCI}; the UCI-covertype includes $581012$ biology instances with $d=54$ forest cover types attributes and $K=7$ classes; the CIFAR-10 dataset consists $60000$, $d=32\times 32 \times 3$ color images in $K=10$ classes. Regarding the model architecture and hyper-parameters settings, we mainly follow~\citet{xu2022neural,zhou2020neuralUCB}. In particular, we use \acrshort{DNN} with ReLU activation, $L=2$ layers, $m = 100$ dimension for the encoder weights matrices, and the last output feature dimension $m_L = d=20$ for the synthetic and $d=64$ for real-world datasets. We set $\lambda=1$, the exploration rate $\alpha_t=0.02$, and the number of iterations to update the neural network $n=1000$. We also set $H=100$ ($10$ on MNIST, UCI-covertype, and CIFAR-10) rounds starting from round $2000$ ($10000$ on MNIST, UCI-covertype, and CIFAR-10)) to update \acrshort{DNN} weights $\mathbf{w}$ following \acrshort{NeuralLinUCB} setting. 

\textbf{Baseline details.} Regarding the baseline comparison, there also exists GLMUCB~\citep{filippi2010parametric,Zenati2022kernelUCB}, KernelUCB~\citep{valko2013finite}, and BootstrappedNN~\citep{riquelme2018deep}. That said, since~\citet{xu2022neural,zhou2020neuralUCB} has shown \acrshort{NeuralUCB} and \acrshort{NeuralLinUCB} better than them in such setting, while our results have lower regret than \acrshort{NeuralUCB} and \acrshort{NeuralLinUCB}, this implies that our \acrshort{ours} is also better than these aforementioned related baselines in terms of cumulative regret. We additionally show this comparison in Figure~\ref{fig:full_demo}.

\textbf{Source code and computing systems}.
Our source code includes the notebook demo, dataset scripts, setup for the environment, and our provided code (detail in README.md). We run our code on a single GPU: NVIDIA RTX~A5000-24564MiB with 8-CPUs: AMD Ryzen Threadripper 3960X 24-Core with 8GB RAM per each and require 8GB available disk space for storage. 

\subsection{Additional results}\label{apd:add_results}
\subsubsection{Uncertainty estimation evaluations}\label{apd:exp_uncertainty}
We evaluate the uncertainty quality by using calibration and sharpness of models across time horizon $t\in [T]$~\citep{bui2024density,bui2024densifysoftmax}. Regarding calibration, this intuitively means that a $p$ confidence interval
contains the target reward $r$ $p$ of the time. Hence, given a forecast from \acrshort{UCB} at time $t$, let $F_t: \mathbb{R} \rightarrow [0,1]$ to denote the CDF of this forecast at $\mathbf{x}_t$, then the calibration error for this forecast is
\small
\begin{equation}\label{eq:calib}
    cal_1(\{F_t,r_t\}_{t=1}^T) := \sum_{j=1}^m \left(p_j - \frac{\left|\{r_t|F_t(r_t) \leq p_j, t=1,\cdots,T\}\right|}{T}\right)^2,
\end{equation}
\normalsize
for each threshold $p_j$ from the chosen of $m$ confidence level $0\leq p_1<p_2<\cdots<p_m\leq 1$.

Regarding sharpness, this means that the confidence intervals should be as tight as possible, i.e., $Var(F_t)$ of the random variable whose CDF is $F_t$ to be small~\citep{kuleshov2018accurate}. Formally, the sharpness score follows
\small
\begin{equation}\label{eq:sharpness}
    sha(F_1,\cdots,F_T) := \sqrt{\frac{1}{T} \sum_{t=1}^T var(F_t)}.
\end{equation}
\normalsize
We show the quantitative results for calibration in Equation~\ref{eq:calib} and sharpness in Equation~\ref{eq:sharpness} in Table~\ref{tab:quatitative} and qualitatively visualize on Figure~\ref{fig:calib}~(a).

\begin{table}[ht!]
    \centering
    \begin{tabular}{cccc}
        \toprule  
         \textbf{Methods} & \textbf{Cumulative reward ($\uparrow$)} & \textbf{Calibration Error ($\downarrow$)} & \textbf{Sharpness ($\downarrow$)}\\
         \midrule
         LinUCB & 7459.0812 $\pm$ 32.9722 & 0.7425 $\pm$ 0.0301 & 0.2095 $\pm$ 0.0191\\
         NeuralUCB & 10658.3046 $\pm$ 60.5330 & 0.2634 $\pm$ 0.0146 & 1.0733 $\pm$ 0.0110\\
         Neural-LinUCB & 10929.2430 $\pm$ 58.8243 & 0.8991 $\pm$ 0.1840 & 0.2042 $\pm$ 0.0213\\
         \textbf{\acrshort{ours}} & \textbf{11326.6471 $\pm$ 50.1880} & \textbf{0.1492 $\pm$ 0.0659} & \textbf{0.8242 $\pm$ 0.2802}\\
         \bottomrule
    \end{tabular}
    \caption{Cumulative reward and uncertainty quality performance (i.e., calibration~\citep{calibrated2019malik} and sharpness~\citep{gneiting2007probabilistic}) on $h_1(\mathbf{x}) = 10(\mathbf{x}^\top \bm{\theta})^2$ dataset, averaged over $5$ trials. \textbf{Our \acrshort{ours} is more well-calibrated and still having a sharp \acrshort{UCB}, resulting in a higher cumulative rewards than other methods.}}
    \label{tab:quatitative}
\end{table}

To further understand the improvement of uncertainty estimation across the learning time horizon, we additionally evaluate calibration on hold-out validation data over different checkpoints across time steps, $t=\{0, 2000, 5000, 7500\}$. Since we use validation data to evaluate, then the calibration in this setting is as follows: given a forecast from UCB at time $t$, let $F_i^t: \mathbb{R} \rightarrow [0,1]$ to denote the CDF of this forecast at $\mathbf{x}_i$, then the calibration error for this forecast is
\small
\begin{equation}\label{eq:calib2}
    cal_2(\{F^t_i,r_i\}_{i=1}^n) := \sum_{j=1}^m \left(p_j - \frac{\left|\{r_i|F_i^t(r_i) \leq p_j, i=1,\cdots,n\}\right|}{n}\right)^2,
\end{equation}
\normalsize
where $n$ is the number of samples in the validation set.

\begin{figure*}[ht!]
    \centering
    \setlength{\tabcolsep}{0.5pt}
    \hspace*{-0.05in}
    \begin{tabular}{cccc}
    \includegraphics[width=0.25\linewidth]{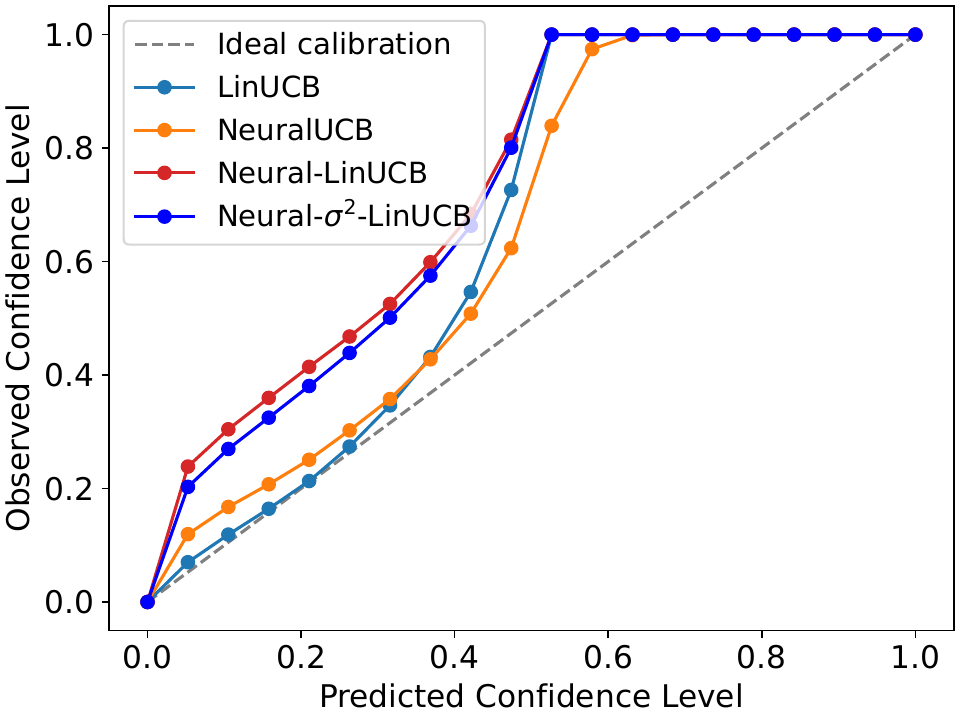}&
    \includegraphics[width=0.25\linewidth]{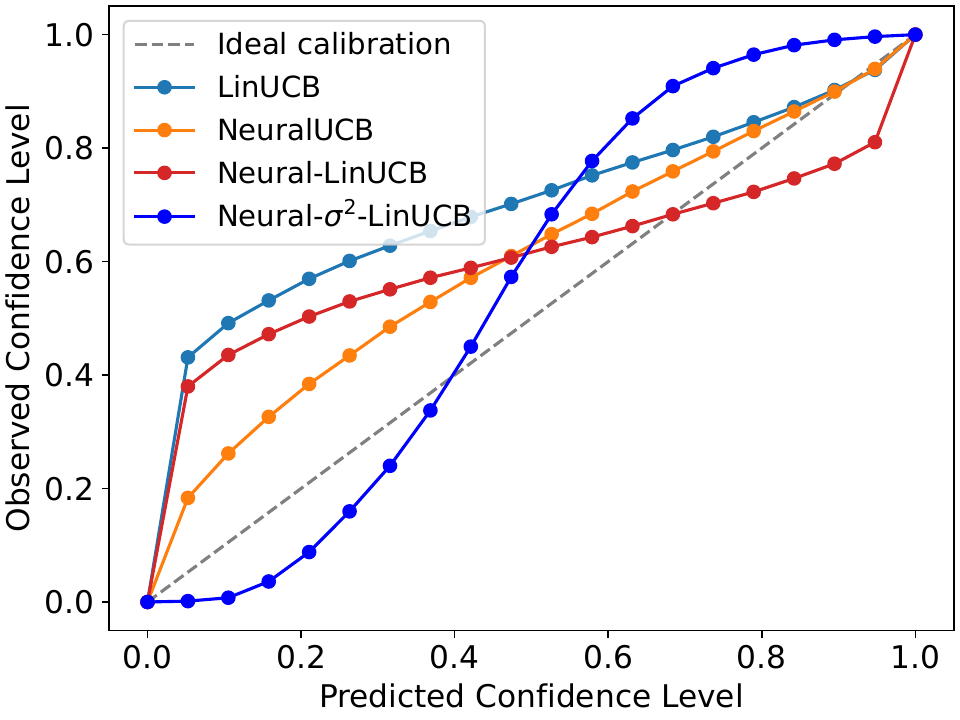}&
    \includegraphics[width=0.25\linewidth]{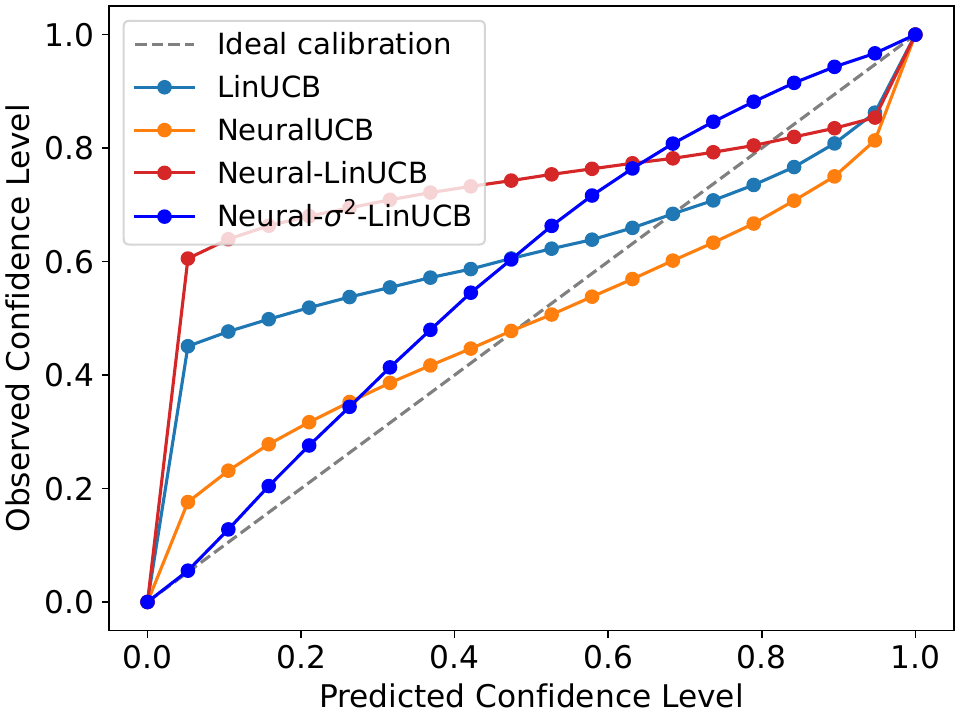}&
    \includegraphics[width=0.25\linewidth]{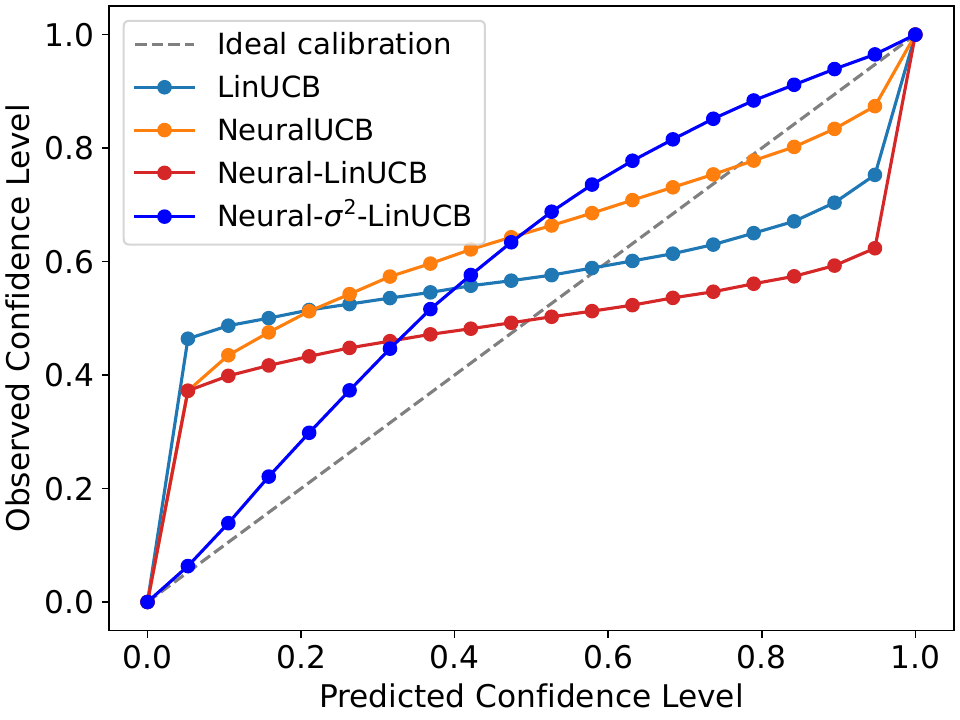}\\
        \scriptsize Round $t=0$ & \scriptsize Round $t=2000$  & \scriptsize Round $t=5000$ & \scriptsize Round $t=7500$
    \end{tabular}
    \caption{Visualization of calibration error in Equation~\ref{eq:calib2} with reliability diagram on $h_1(\mathbf{x}_{t,a})$ dataset (arm: $0$).}
    \label{fig:calib_details_0}
\end{figure*}

\begin{figure*}[ht!]
    \centering
    \setlength{\tabcolsep}{0.5pt}
    \hspace*{-0.05in}
    \begin{tabular}{cccc}
    \includegraphics[width=0.25\linewidth]{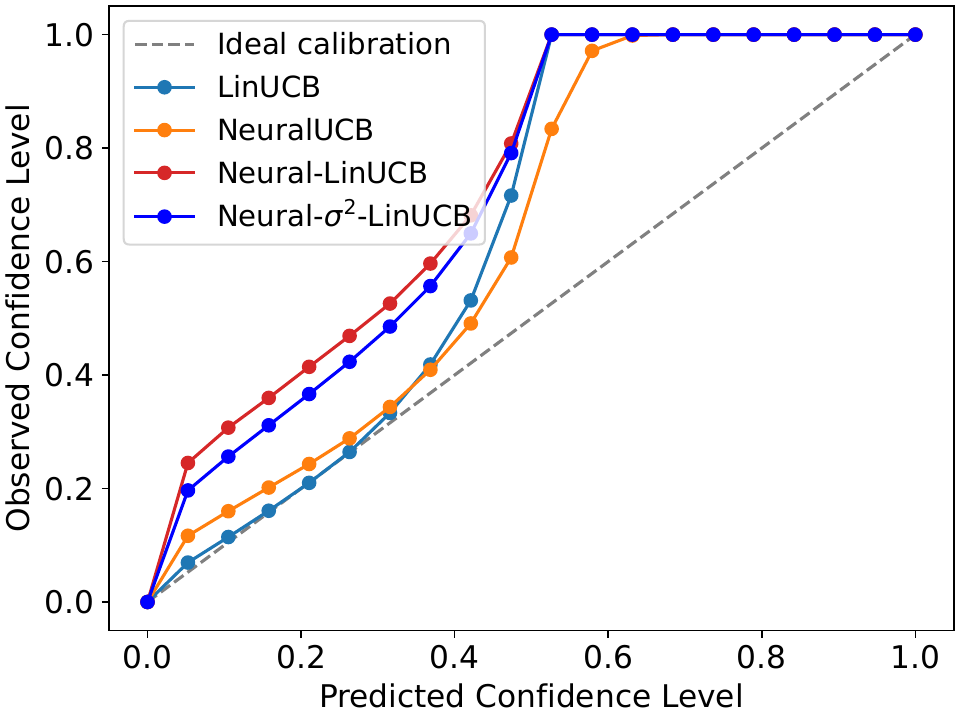}&
    \includegraphics[width=0.25\linewidth]{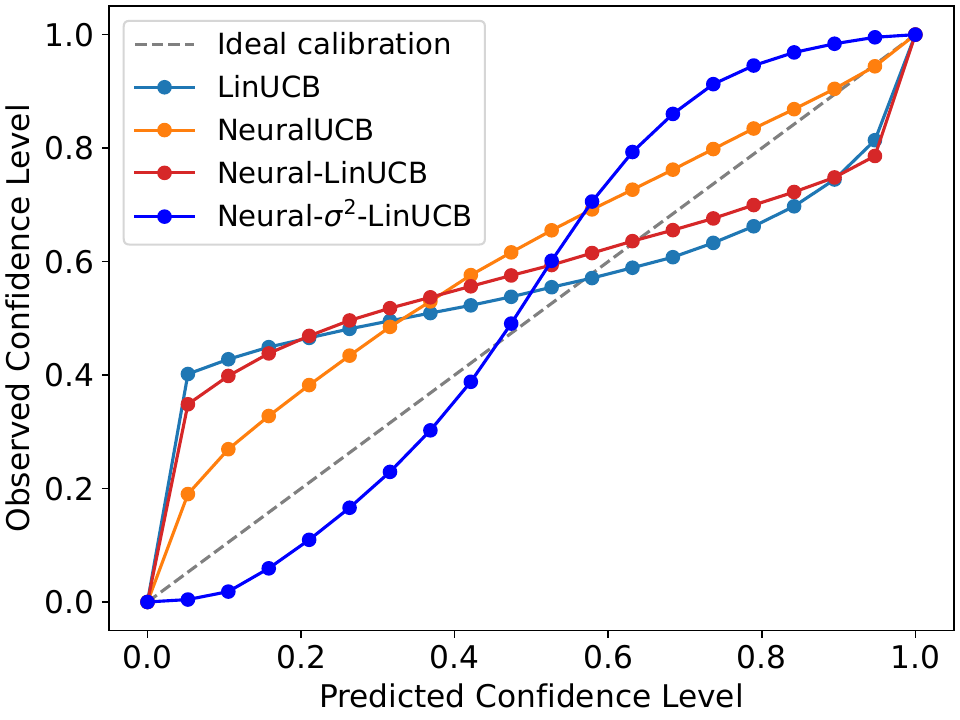}&
    \includegraphics[width=0.25\linewidth]{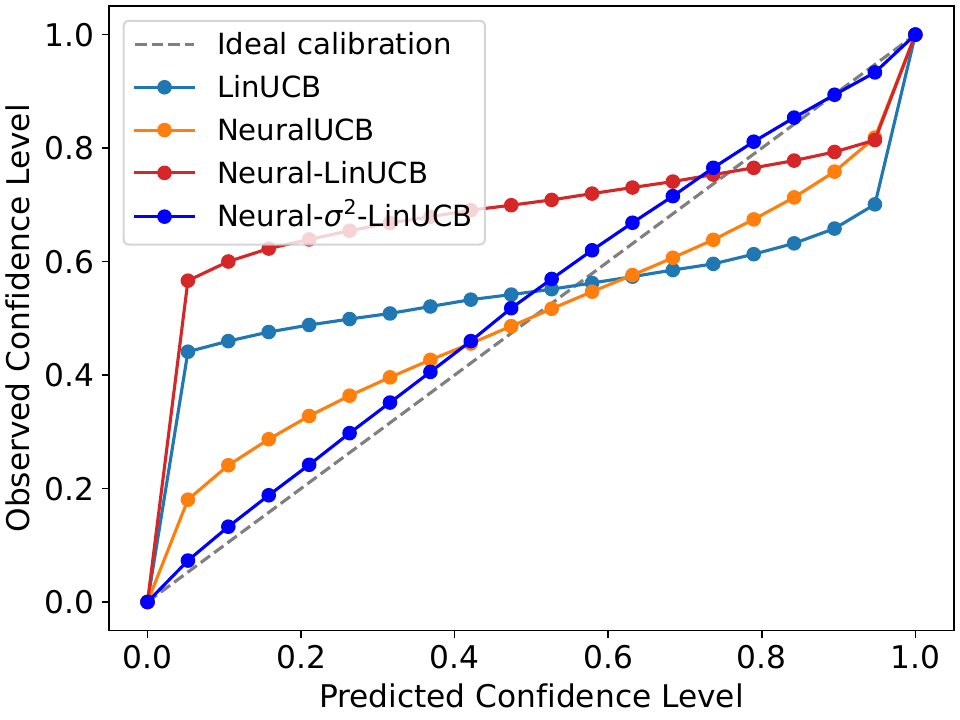}&
    \includegraphics[width=0.25\linewidth]{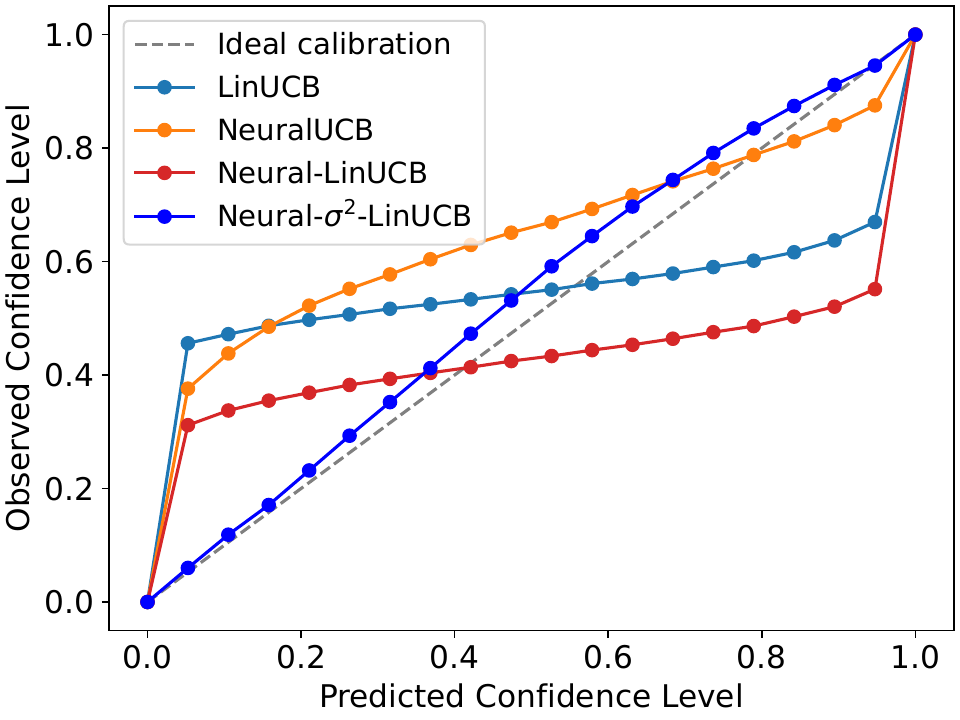}\\
        \scriptsize Round $t=0$ & \scriptsize Round $t=2000$  & \scriptsize Round $t=5000$ & \scriptsize Round $t=7500$
    \end{tabular}
    \caption{Visualization of calibration error in Equation~\ref{eq:calib2} with reliability diagram on $h_1(\mathbf{x}_{t,a})$ dataset (arm: $1$).}
    \label{fig:calib_details_1}
\end{figure*}

\begin{figure*}[ht!]
    \centering
    \setlength{\tabcolsep}{0.5pt}
    \hspace*{-0.05in}
    \begin{tabular}{cccc}
    \includegraphics[width=0.25\linewidth]{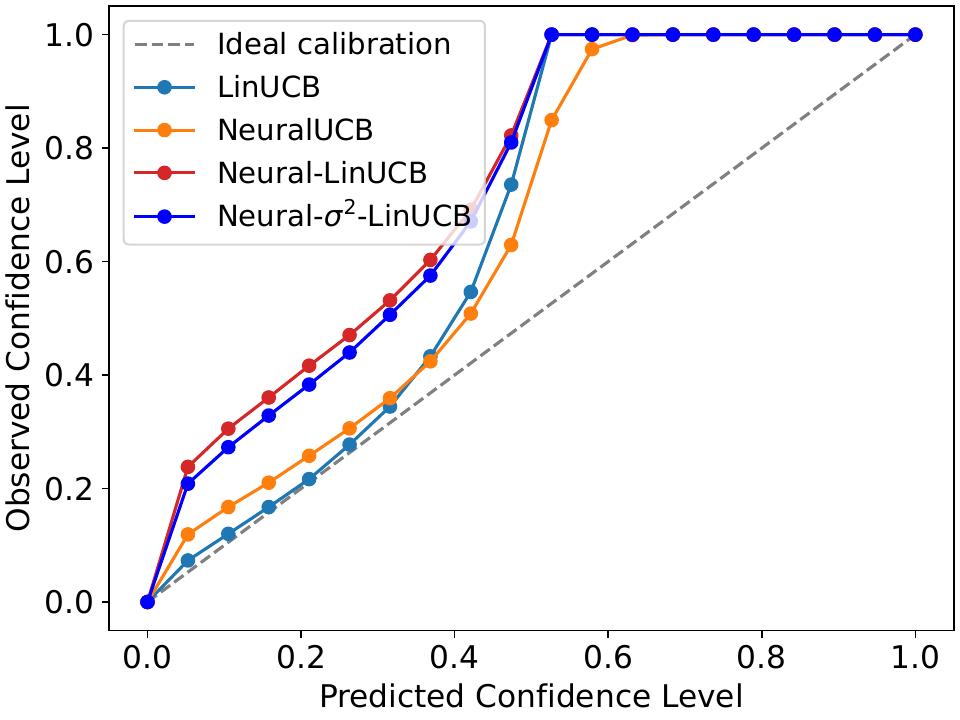}&
    \includegraphics[width=0.25\linewidth]{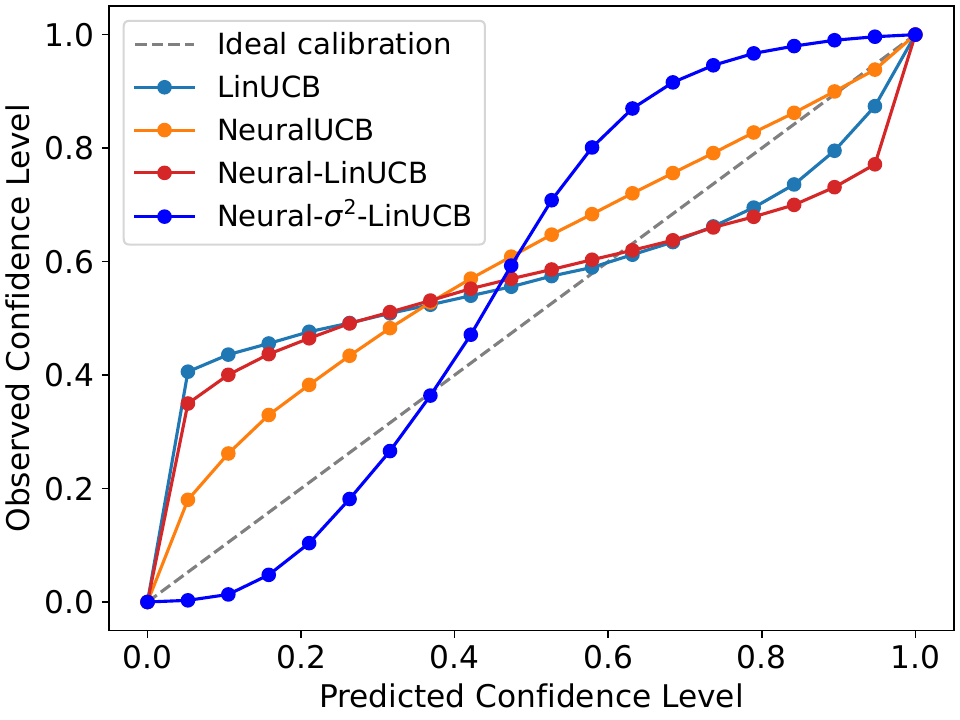}&
    \includegraphics[width=0.25\linewidth]{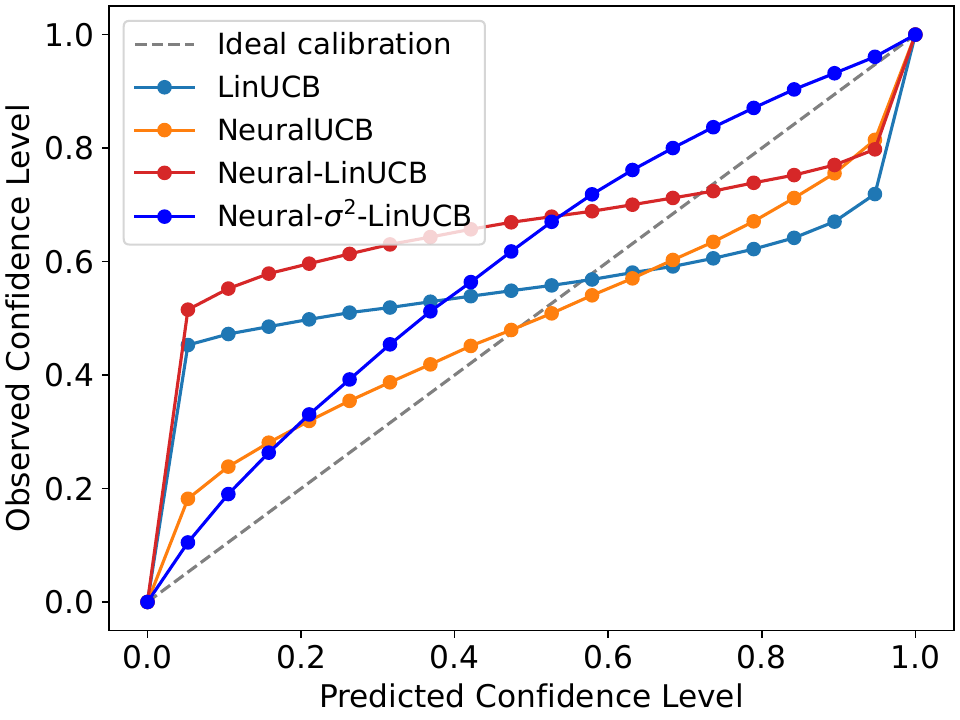}&
    \includegraphics[width=0.25\linewidth]{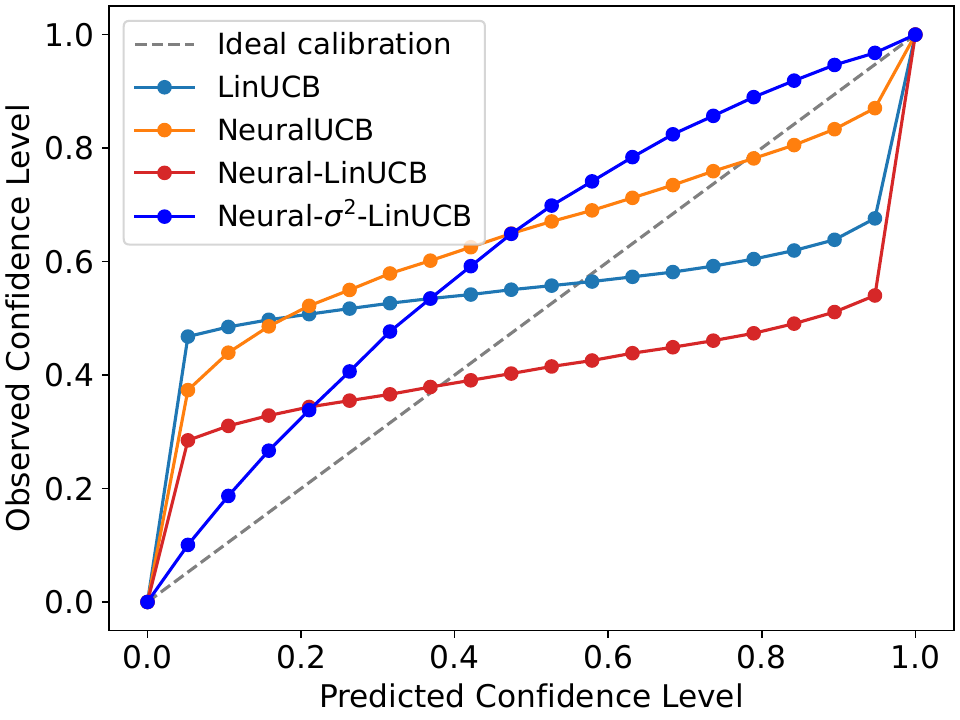}\\
        \scriptsize Round $t=0$ & \scriptsize Round $t=2000$  & \scriptsize Round $t=5000$ & \scriptsize Round $t=7500$
    \end{tabular}
    \caption{Visualization of calibration error in Equation~\ref{eq:calib2} with reliability diagram on $h_1(\mathbf{x}_{t,a})$ dataset (arm: $2$).}
    \label{fig:calib_details_2}
\end{figure*}

\begin{figure*}[ht!]
    \centering
    \setlength{\tabcolsep}{0.5pt}
    \hspace*{-0.05in}
    \begin{tabular}{cccc}
    \includegraphics[width=0.25\linewidth]{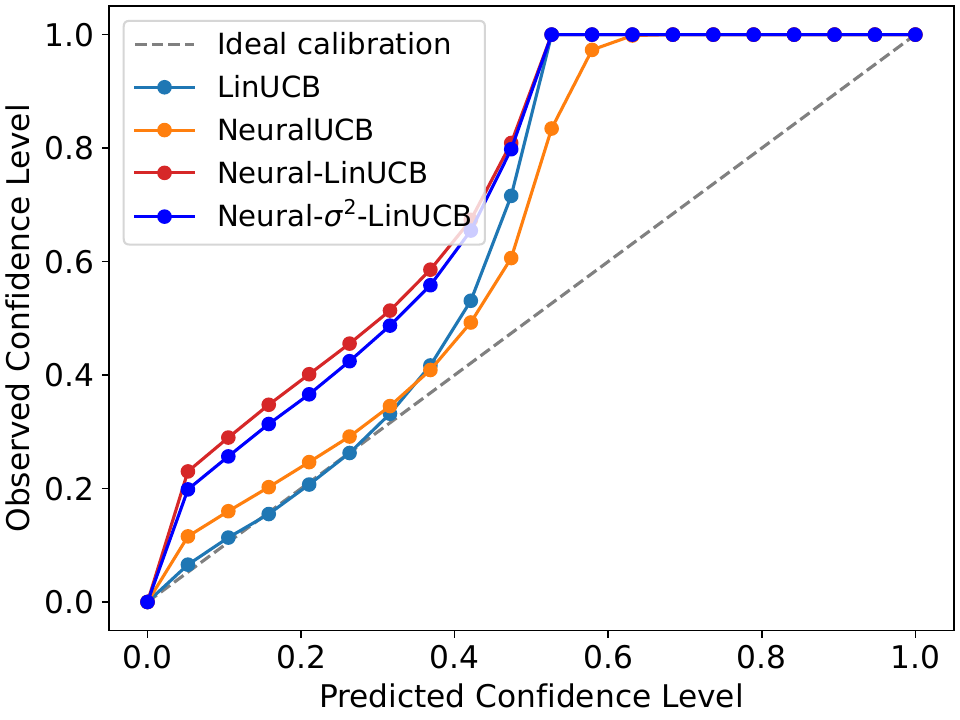}&
    \includegraphics[width=0.25\linewidth]{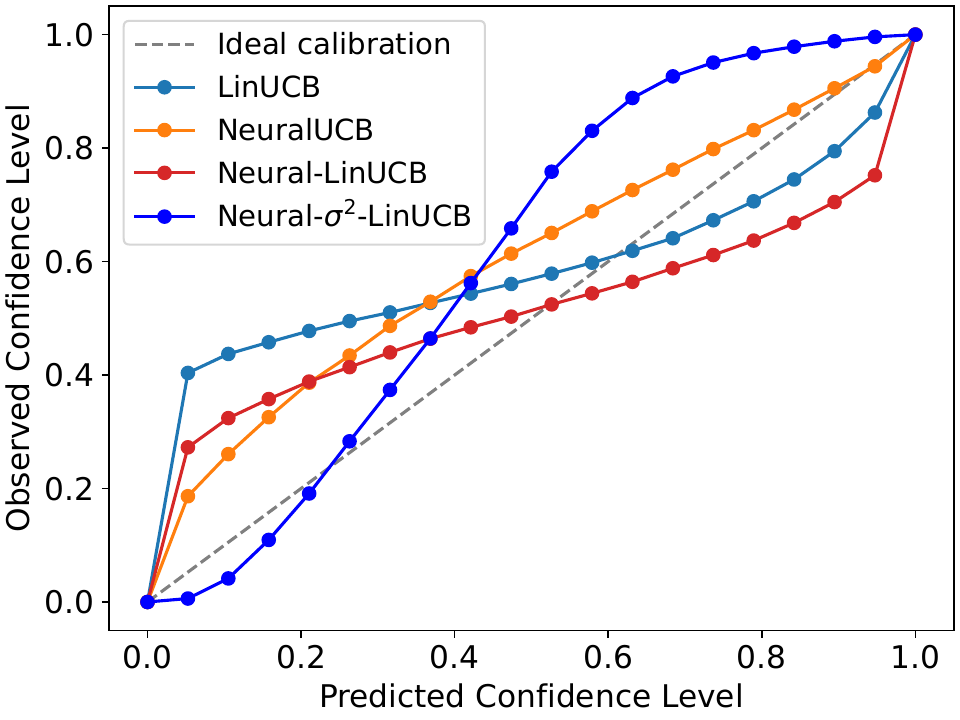}&
    \includegraphics[width=0.25\linewidth]{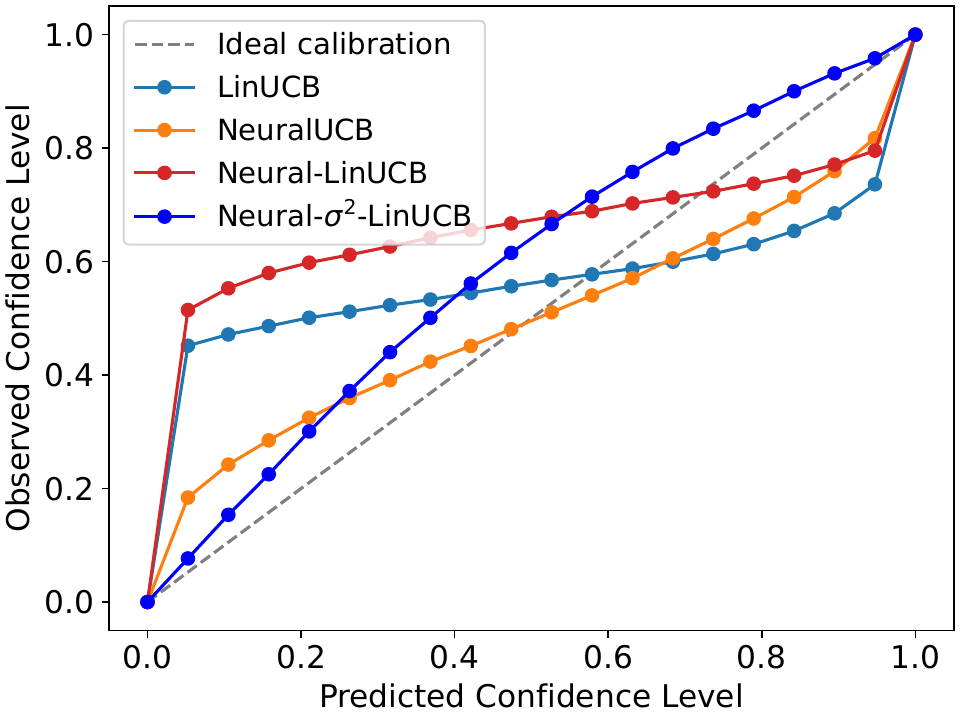}&
    \includegraphics[width=0.25\linewidth]{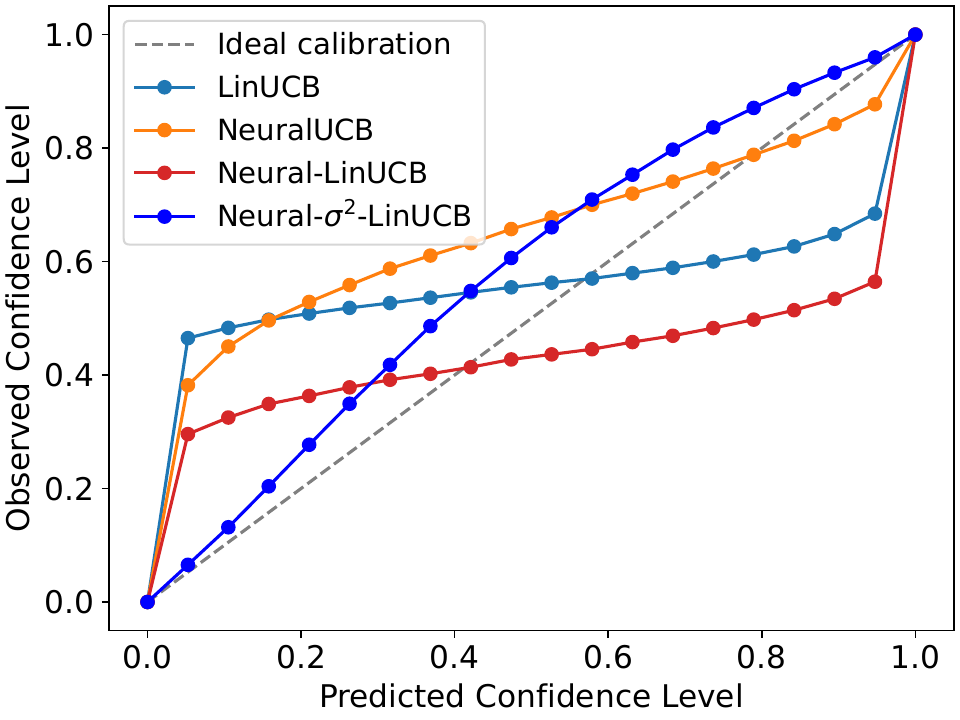}\\
        \scriptsize Round $t=0$ & \scriptsize Round $t=2000$  & \scriptsize Round $t=5000$ & \scriptsize Round $t=7500$
    \end{tabular}
    \caption{Visualization of calibration error in Equation~\ref{eq:calib2} with reliability diagram on $h_1(\mathbf{x}_{t,a})$ dataset (arm: $3$).}
    \label{fig:calib_details_3}
\end{figure*}

We visualize the calibration error in Equation~\ref{eq:calib2} by the reliability diagram across correspond to different arms, $a=\{0,1,2,3\}$ in Figure~\ref{fig:calib_details_0},~\ref{fig:calib_details_1},~\ref{fig:calib_details_2}, and~\ref{fig:calib_details_3} correspondingly. Overall, we can see that when $t=0$, all of the models are uncalibrated because of no learning data. But when $t$ grows, \acrshort{ours} are almost always more calibrated than other \acrshort{UCB} algorithms. This once again confirms the hypothesis that our \acrshort{ours} algorithm can improve the uncertainty quantification quality of \acrshort{UCB}.

\begin{figure*}[ht!]
    \centering
    \hspace*{-0.1in}
     \setlength{\tabcolsep}{-1.2pt}
    \begin{tabular}{ccc}
    \includegraphics[width=0.33\linewidth]{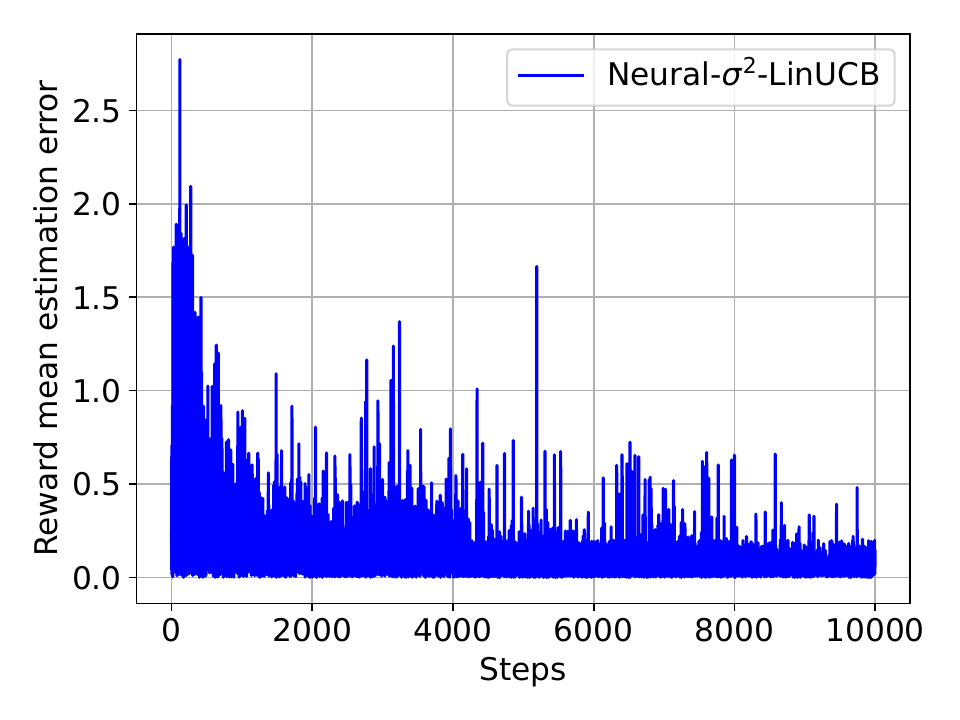}&
    \includegraphics[width=0.33\linewidth]{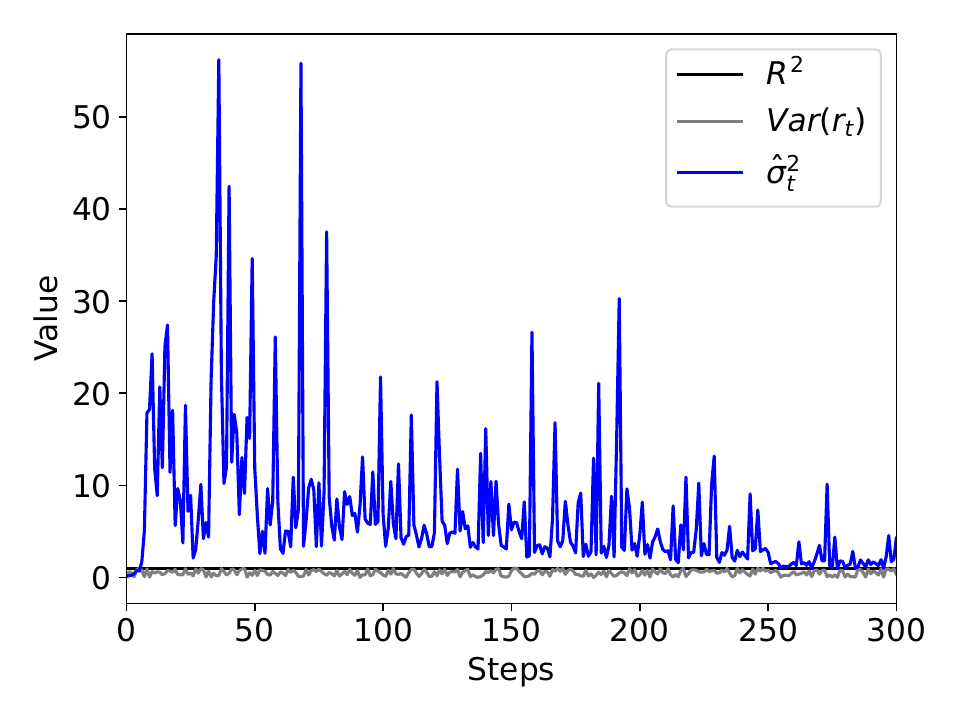}&
    \includegraphics[width=0.33\linewidth]{figures/main/ablation_study_var.pdf}\\
        \scriptsize (a) & \scriptsize (b) & \scriptsize (c)
    \end{tabular}
    \caption{(a) Reward mean estimation error $\left|h(\mathbf{x}_{t,a_t})-\theta_{t,a_t}^\top \phi(\mathbf{x}_{t,a_t};\mathbf{w})\right|$ across $t\in [T]$. (b) Reward variance $Var(r_t)$, our estimation for the variance upper bound $\sigma_t^2$, and the upper bound $R^2$ comparison at the first $300$ episodes; (c) Fig~\ref{fig:calib}~(b) in the main paper at the last $300$ episodes. \textbf{When the reward mean estimation quality improves in Figure~\ref{fig:ablation_study2}~(a), the quality of our estimation for $\sigma_t^2$ increase and more accurate in Figure~\ref{fig:ablation_study2}~(c) when compared to Figure~\ref{fig:ablation_study2}~(b).}}
    \label{fig:ablation_study2}
\end{figure*}

To further validate the estimation quality of $\hat{\sigma}_t^2$ in Equation~\ref{eq:est_sigma}. Firstly, recall that Theorem~\ref{thm:variance_bound} implies that the accurate estimation for the variance upper bound $\sigma_t^2$ is a necessary condition for good estimation quality for the reward mean $h(\mathbf{x}_{t,a_t})$. Figure~\ref{fig:calib}~(b) in the main paper confirms when we have a good reward estimation in the last $300$ episodes, then we can obtain an accurate estimation for the variance upper bound (by $\geq Var(r_t)$ and $\leq R^2$). We add Figure~\ref{fig:ablation_study2}~(b) to compare with Figure~\ref{fig:calib}~(b) (i.e., Figure~\ref{fig:ablation_study2}~(c)) in the first $300$ episodes, we can see that when the reward mean estimation has high estimation errors (see Figure~\ref{fig:ablation_study2}~(a)), the estimation for the variance upper bound $\sigma_t^2$ is inaccurate.

\subsection{Computational efficiency evaluations}\label{apd:exp_latency}
\begin{figure}[ht!]
    \centering
    \begin{tabular}{cc}
    \includegraphics[width=0.47\linewidth]{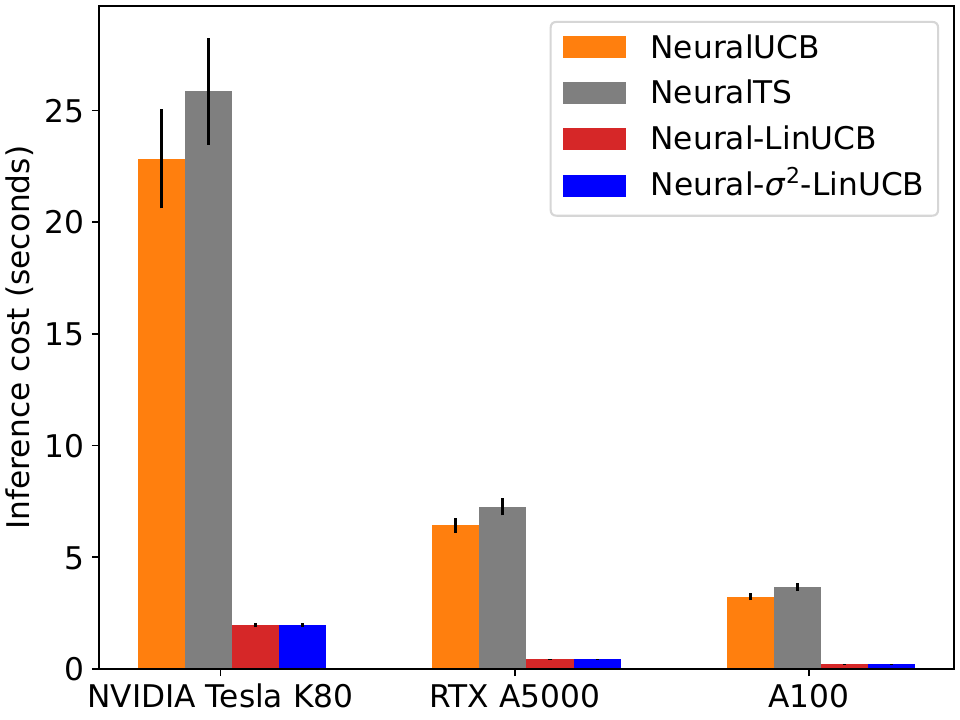}&
    \includegraphics[width=0.47\linewidth]{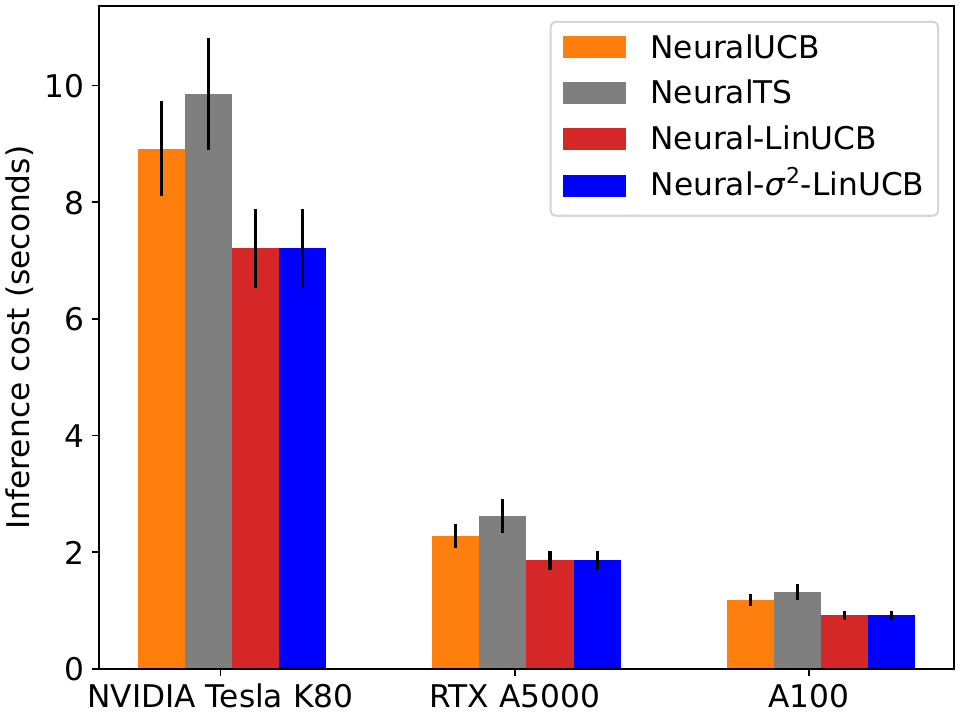}\\
        (a) & (b)
    \end{tabular}
    \caption{Computational cost comparison of neural contextual bandits algorithms on MNIST for running $100$ rounds across three modern GPU architectures, including in the arm selection step (a) and the \acrshort{DNN} update step (b).}
    \label{fig:latency}
\end{figure}
We extensively evaluate our model on three different settings, including: (1) a single GPU: NVIDIA Tesla~K80 accelerator-12GB GDDR5 VRAM with 8-CPUs: Intel(R) Xeon(R) Gold 6248R CPU @ 3.00GHz with 8GB RAM per each; (2) a single GPU: NVIDIA RTX~A5000-24564MiB with 8-CPUs: AMD Ryzen Threadripper 3960X 24-Core with 8GB RAM per each; and (3) a single GPU: NVIDIA A100-PCIE-40GB with 8 CPUs: Intel(R) Xeon(R) Gold 6248R CPU @ 3.00GHz with 8GB RAM per each. Figure~\ref{fig:latency} summarizes these results with the number of rounds to frequently update \acrshort{DNN} $H=10$ and the number of rounds $t=100$. 

Figure~\ref{fig:latency}~(a) shows the latency of the arm selection step in Line 5 to Line 8 of Algorithm~\ref{alg:ours}. We can see that by computing the \acrshort{UCB} value from a linear model on the last feature representation of \acrshort{DNN}, our \acrshort{ours} and \acrshort{NeuralLinUCB} are much more efficient than other baselines. Our better results are consistent across different CPU/GPU architectural settings. For instance, in the lower resource hardware like with NVIDIA Tesla~K80, our results are faster than around $20$ seconds. Regarding powerful hardware like NVIDIA A100, we are still faster than around $4$ seconds. These results are consistent with the result of~\citet{xu2022neural} and could be explained by the fact that \acrshort{NeuralUCB} and \acrshort{NeuralTS} need to perform \acrshort{UCB} and Thompson-sampling exploration on all the parameters of \acrshort{DNN}. As a result, the lower the computational hardware, the less computationally efficient than our algorithms.

Figure~\ref{fig:latency}~(b) shows the latency of the \acrshort{DNN} update step from Line 17 to Line 21 of Algorithm~\ref{alg:ours}. Similarly, we observe that the more powerful the hardware, the less time it takes to optimize the \acrshort{DNN} models. Regarding comparison with other baselines, by using the same technique to save computational cost in \acrshort{DNN} training from \acrshort{NeuralLinUCB}~\citep{xu2022neural}, \acrshort{ours} also enjoys a more computationally efficient than other neural contextual bandits baselines.

\subsubsection{Regret performance evaluations}\label{apd:exp_regret}
\begin{figure*}[ht!]
    \centering
    \setlength{\tabcolsep}{0.5pt}
    \begin{tabular}{ccc}
    \includegraphics[width=0.33\linewidth]{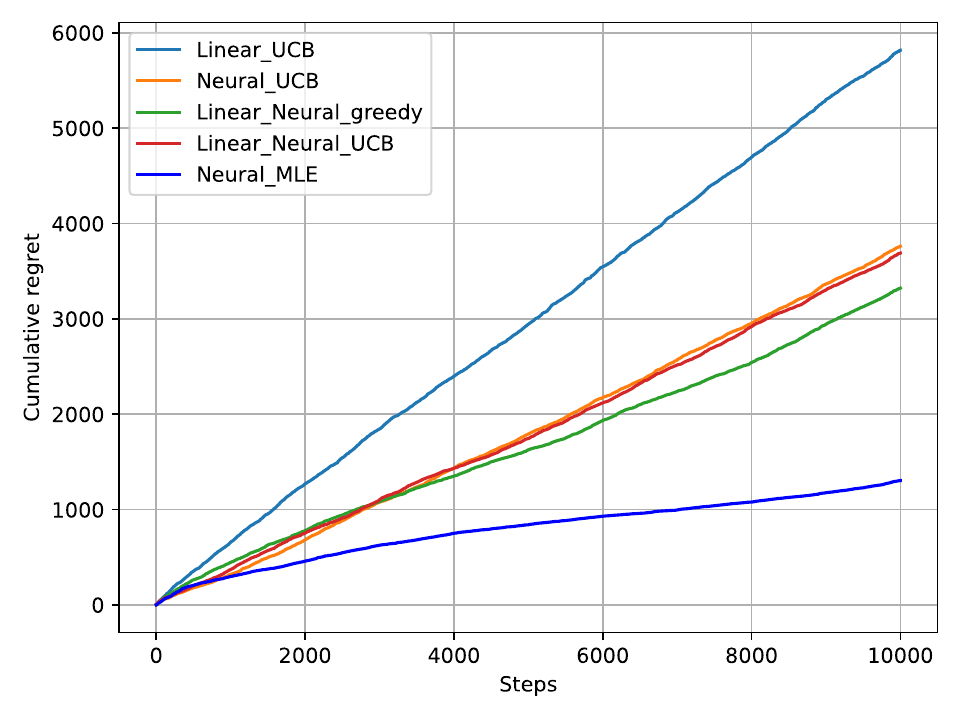}&
    \includegraphics[width=0.33\linewidth]{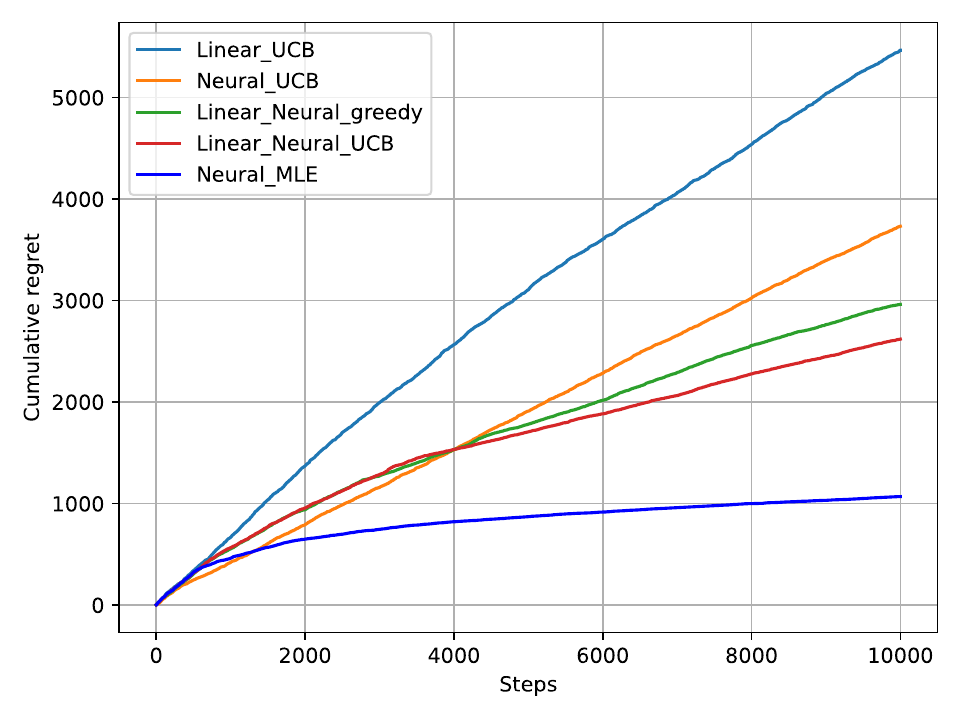}&
    \includegraphics[width=0.33\linewidth]{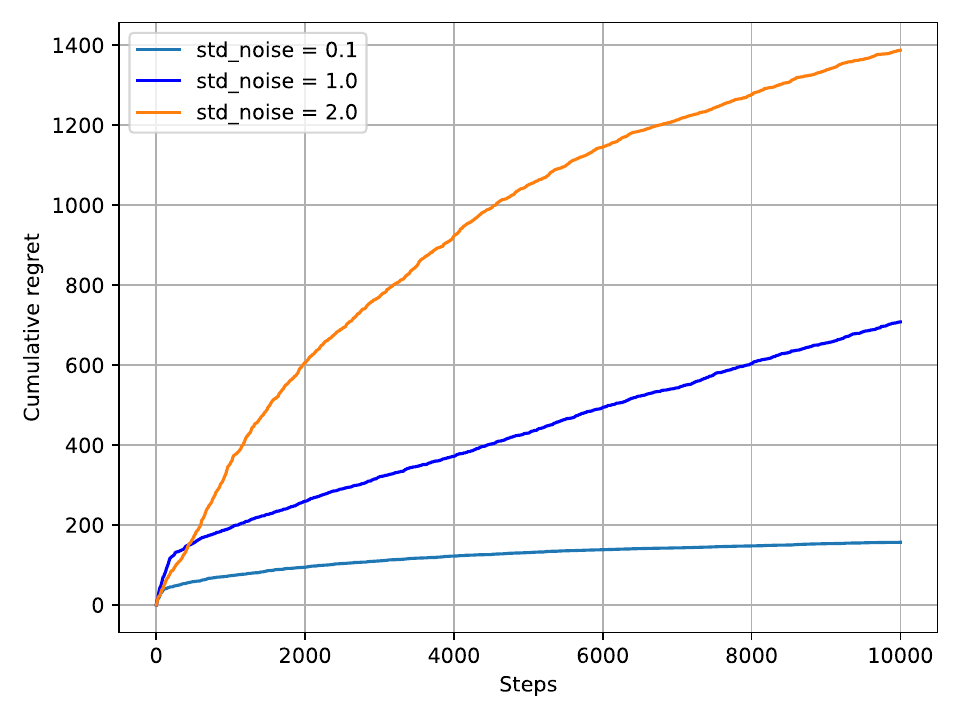}\\
        (a) & (b) & (c)
    \end{tabular}
    \caption{$h_1(\mathbf{x}) = 10(\mathbf{x}^\top \bm{\theta})^2$. (a) The noise $\xi_t$ monotonically increases from $1$ to $10$. (b) The noise $\xi_t$ monotonically decreases from $2$ to $0$. (c) Comparison across different $\text{std\_noise}$.}
    \label{fig:demo_in_de_crease_noise}
\end{figure*}

We additionally show our model behaviors across different types of stochasticity regarding the reward noise $\xi_t$ on $h_1(\mathbf{x}) = 10(\mathbf{x}^\top \bm{\theta})^2$ dataset in Figure~\ref{fig:demo_in_de_crease_noise}. Specifically, from $t=0$ to $t=T=10000$, we set $\xi_t$ increase monotonically from $1$ to $10$ in Figure~\ref{fig:demo_in_de_crease_noise}~(a), and decreases monotonically from from $2$ to $0$ in Figure~\ref{fig:demo_in_de_crease_noise}~(b). We observe that \acrshort{ours}'s results are robust by always having a significantly lower cumulative regret than other baselines. Furthermore, when the noise decreases and reaches very small values at the final steps, our cumulative regret becomes almost constant.

Similarly to the setting of~\citet{zhou2020neuralUCB}, we also consider the non-stochasticity for the reward noise variance at round $t$, i.e., $\xi_t \sim \mathcal{N}(0,{\text{std\_noise}}^2)$ in Figure~\ref{fig:demo_in_de_crease_noise}~(c), where $\text{std\_noise} = \{0.1,1.0,2.0\}$. It can be seen from this figure that when the $\text{std\_noise}$ decreases, our cumulative regret also decreases respectively. And for the $\text{std\_noise} = 0.1$, at the final steps, \acrshort{ours}'s cumulative regret also becomes almost constant. 

Regarding the effectiveness of highly noisy cases for the rewards to our model performance, we also observe that when the reward noise level is high (e.g., $\text{std\_noise} = 2.0$), our model often has a higher cumulative regret than the cases of lower reward noise levels (e.g., $\text{std\_noise} = \{0.1,1.0\}$). In addition, when the noise value increases across rounds, all methods have increased cumulative regret (e.g., Figure~\ref{fig:demo_in_de_crease_noise}~(a)), but our method is more robust by having a lower cumulative regret than other baselines. 

\begin{figure}[ht!]
\begin{minipage}{0.45\textwidth}
    \begin{center}
      \includegraphics[width=1.0\linewidth]{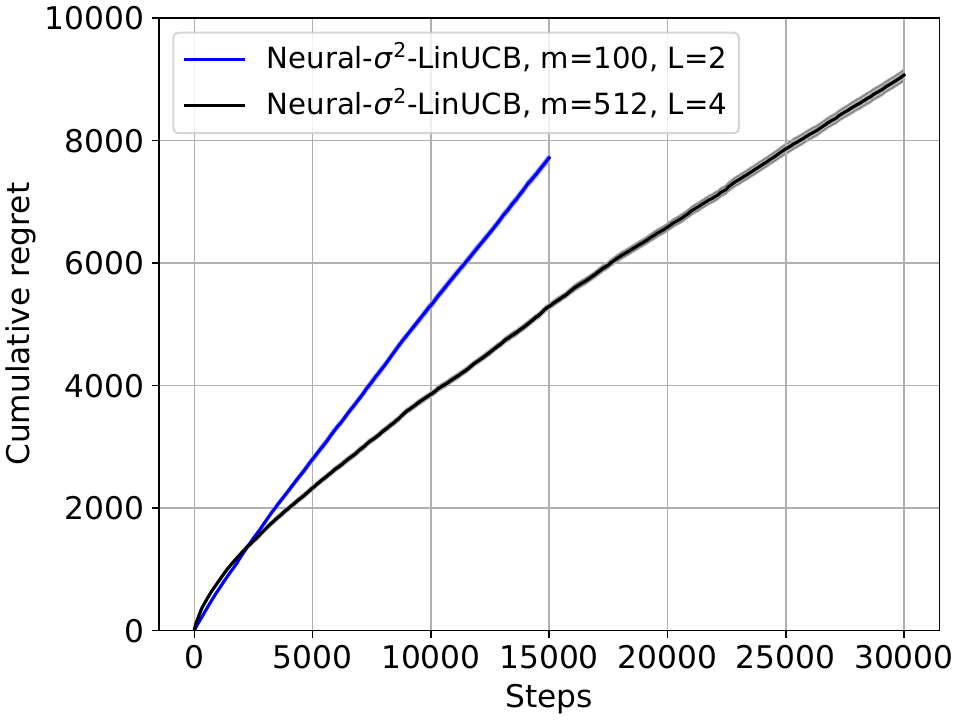}
    \end{center}
    \vspace{-0.1in}
    \caption{Blue line is the performance of model size $m=100$, $L=2$, $\mathbf{w}$ is updated every $H=10$ rounds starting from $t=10000$, and $T=15000$ of our method on CIFAR-10 in Figure~\ref{fig:real_world}~(d). The black line is with $m=512$, $L=4$, $\mathbf{w}$ is updated every $H=10$ rounds from $t=25000$, and $T=30000$. \textbf{This shows when the model capacity and the learning process increase, we can achieve a lower cumulative regret.}}
    \label{fig:model_size}
\end{minipage}
\hspace{0.5in}
\begin{minipage}{0.45\textwidth}
    \begin{center}
      \includegraphics[width=1.0\linewidth]{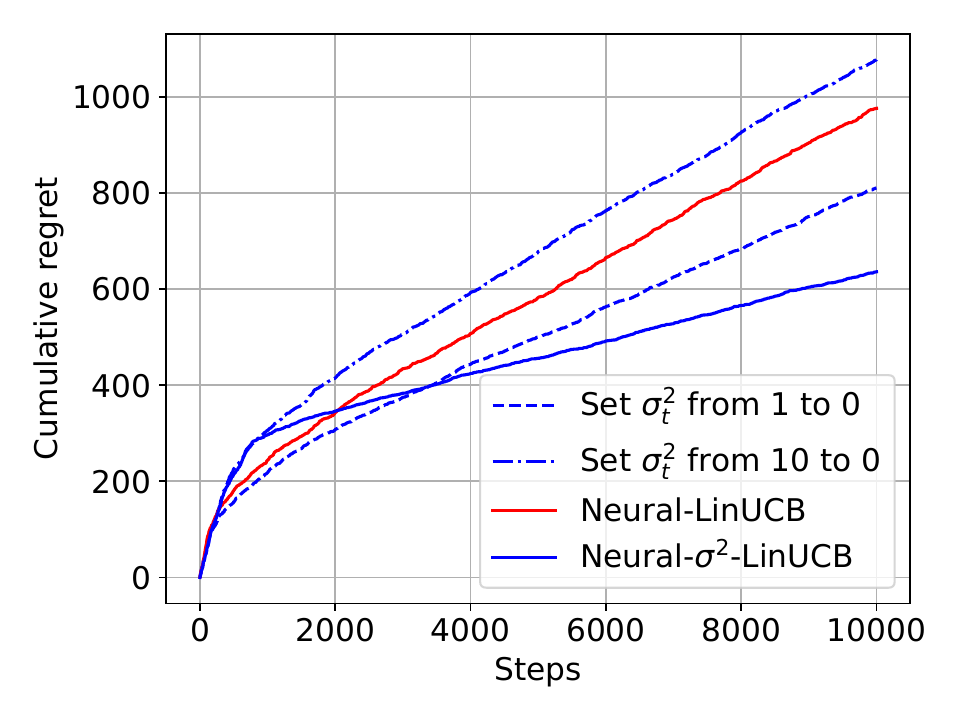}
    \end{center}
    \vspace{-0.1in}
    \caption{Simple baseline by setting $\sigma_t^2$ heuristically on $h_1(\mathbf{x}) = 10(\mathbf{x}^\top \bm{\theta})^2$ with $R=1$, including setting (1) $\sigma_t^2$ decrease from $1$ to $0$ and (2) $\sigma_t^2$ decrease from $10$ to $0$ across time horizon $T$. \textbf{Since this is a heuristic, so sometimes better (e.g., if $\sigma_t^2 \leq R^2$), sometimes worse (e.g., if $\sigma_t^2 > R^2$) than \acrfull{NeuralLinUCB}. But it is always worse than \acrshort{ours} that tries to estimate $\sigma_t^2$ because our estimation $\hat{\sigma}_t^2$ aims to satisfy Thm~\ref{thm:variance_bound}.}} 
    \label{fig:heuristic}
\end{minipage}
\end{figure}

In Figure~\ref{fig:model_size}, we show an ablation study by increasing the model size and a longer time horizon on CIFAR-10. We can see that when the model capacity increases, we can achieve a lower cumulative regret, confirming our claim in Remark~\ref{rem:est_error}.

In Figure~\ref{fig:heuristic}, we compare with a heuristic selection for $\sigma_t^2$ in practice. We can see that since this is a heuristic selection, $\sigma_t^2$ may not satisfy conditions in the Equation~\ref{eq:noise_gen}, leading to worse performances than our proposed estimation in Equation~\ref{eq:est_sigma}.

\begin{figure}[ht!]
\begin{minipage}{0.45\textwidth}
    \begin{center}
      \includegraphics[width=1.0\linewidth]{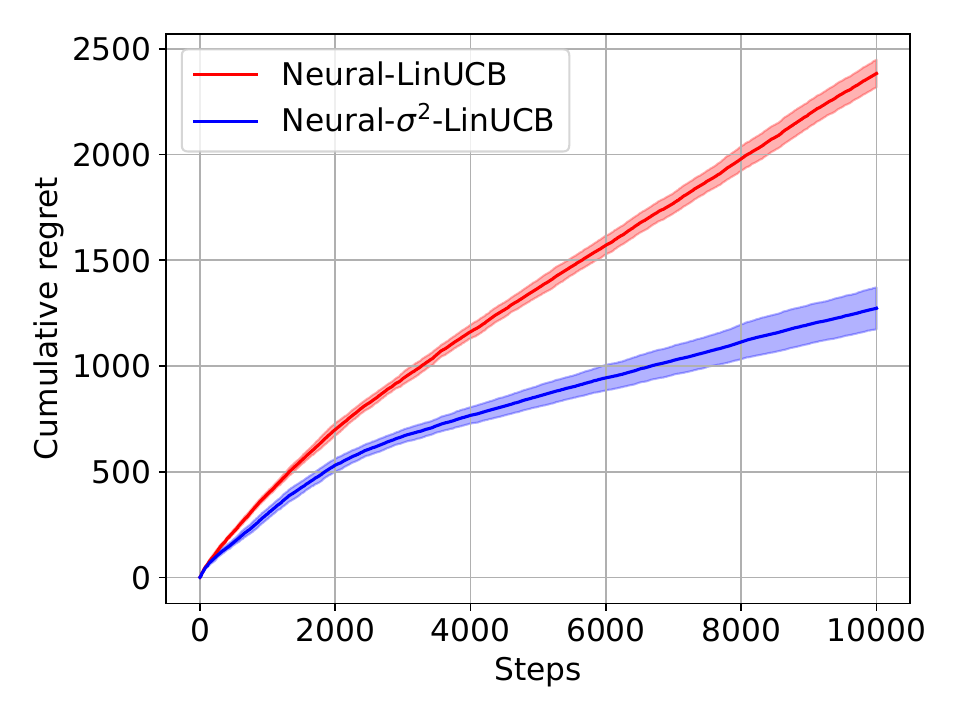}
    \end{center}
    \vspace{-0.1in}
    \caption{Cumulative regret results on the synthetic data $h_1(\mathbf{x}) = 10(\mathbf{x}^\top \bm{\theta})^2$ with the true $\alpha_t$ in our Theorem~\ref{theo:regret_upper_bound} with $d=20$, $H=100$, $K=4$, $\lambda=1$, $M=0.1$, and $\delta=0.1$. \textbf{This demonstrates that using the $\alpha_t$ suggested by the theory, our method is also better than \acrshort{NeuralLinUCB}, confirming our tighter regret bound in Theorem~\ref{theo:regret_upper_bound}}.} 
    \label{fig:true_alpha}
\end{minipage}
\hspace{0.05in}
\begin{minipage}{0.55\textwidth}
    \centering
    \includegraphics[width=1.0\linewidth]{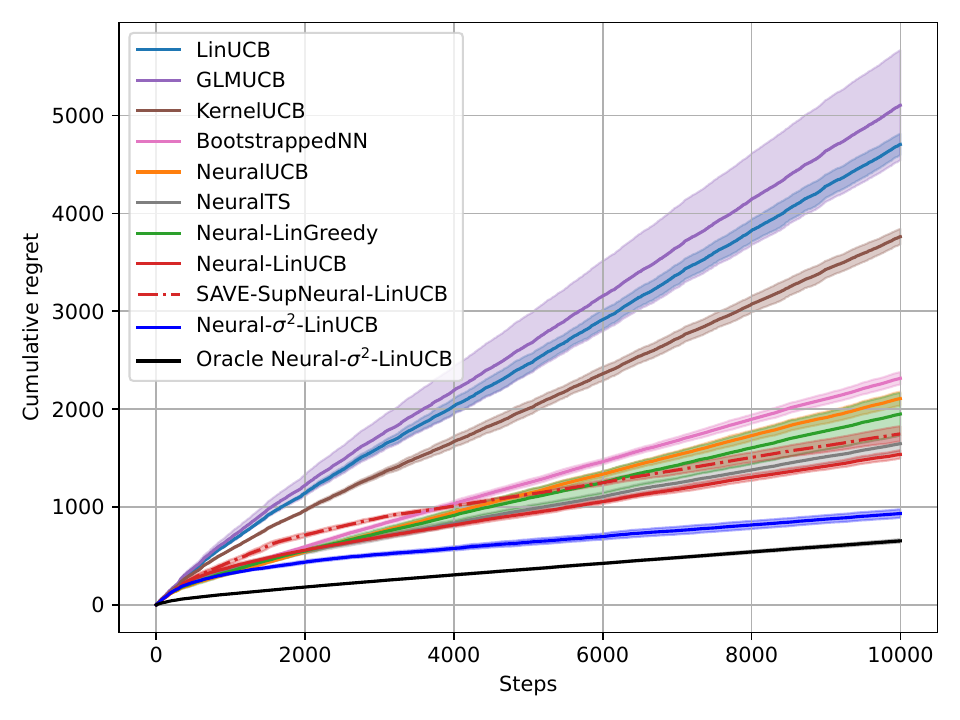}
    \vspace{-0.1in}
    \caption{Support figure for Figure~\ref{fig:demo}: Full cumulative regret results on $h_1(\mathbf{x}) = 10(\mathbf{x}^\top \bm{\theta})^2$ across 10 runs with different seeds.}
    \label{fig:full_demo}
\end{minipage}
\end{figure}

To explore the effect of the actual value used for exploration rate $\alpha_t$, we provide a result for setting the true value $\alpha_t$ in Theorem~\ref{theo:regret_upper_bound} and comparing with \acrshort{NeuralLinUCB} in Figure~\ref{fig:true_alpha} (we can not show \acrfull{NeuralUCB} results because computing $\gamma_t$ in~\citet{zhou2020neuralUCB} is very computationally expensive as the determinant of the gradient of the neural-net covariance matrix). Figure~\ref{fig:true_alpha} shows our algorithm is better than \acrshort{NeuralLinUCB}, once again confirming our theoretical and experimental results in the main paper.

Finally, we summarize our cumulative regret comparison with all other baselines in Figure~\ref{fig:full_demo}. It is also worth noticing that we also compare our results with an extension of \acrshort{SAVE} in neural bandits, i.e., SAVE-SupNeural-LinUCB by considering the context in their Algorithm~1~\citep{zhao2023variance} as the feature representation from \acrshort{DNN}. From Figure~\ref{fig:full_demo}, we can see that even assuming given knowledge of $T$, in neural contextual bandits, this approach still performs poorly in experiments because of the over-exploration~\citep{salgia2023provably}. 
\end{document}